\documentstyle[10pt,avm,psfig,lingmacros,tree-dvips,ulem,QobiTree,alltt]{report}
\avmfont{\sf}
\avmvalfont{\it}
\avmsortfont{\small\it}
\newcommand{\aF}{\sf }
\newcommand{\aV}{\it }
\newcommand{\aS}{\it }
\def\myenumsentence#1{\vspace{1em} \enumsentence{#1} \vspace{1em} }
\def\myeenumsentence#1{\vspace{1em} \eenumsentence{#1} \vspace{1em}}
\def\bibname{REFERENCES}

\begin{document}
\setcounter{page}{1}
\pagenumbering{roman}
\thispagestyle{empty}
\vspace*{1cm}

\begin{center}
{\Huge \bf
A SIGN-BASED PHRASE \\

STRUCTURE GRAMMAR \\

FOR TURKISH\\
}
\vspace{2cm}

{\Large 
by\\
\bf
Onur Tolga \c{S}ehito\u{g}lu
}
\vspace{2cm}

{\Large 
January, 1996\\
Middle East Technical University\\
ANKARA}

\vspace{3cm}

{\Large
\begin{minipage}{9cm}
\centering
In partial fullfilment of the requirements for the degree of \\
Master of Science\\
in\\
The Department of Computer Enginnering
\end{minipage}}
\end{center}

\newpage
\vspace*{.05\textheight}

\begin{center}
{\Huge \bf
Abstract\\}
\addcontentsline{toc}{chapter}{Abstract}
\vspace{4em}

{\Large \bf A Sign-Based Phrase Structure Grammar for Turkish}\\
\vspace{4em}

\c{S}ehito\u{g}lu, Onur Tolga\\
MS., Department of Computer Engineering\\
Supervisor: Assist. Prof. Dr. Cem Boz\c{s}ahin\\

January 1996, \pageref{sonsayfa} pages\\
\vspace{4em}

\end{center}

This study analyses Turkish syntax from an informational point of
view.  Sign based linguistic representation and principles of HPSG
(Head-driven Phrase Structure Grammar) theory are adapted to Turkish.
The basic informational elements are nested and inherently sorted
feature structures called signs.

In the implementation, logic programming tool ALE ---Attribute Logic
Engine--- which is primarily designed for implementing HPSG grammars is
used. A type and structure hierarchy of Turkish language is designed.
Syntactic phenomena such as subcategorization, relative clauses,
constituent order variation, adjuncts, nominal predicates and
complement-modifier relations in Turkish are analyzed. A parser is
designed and implemented in ALE.  
\vspace{2em}


\newpage
\vspace*{.05\textheight}

\begin{center}
{\Huge \bf
\"{O}z\\}
\addcontentsline{toc}{chapter}{\"{O}z}
\vspace{4em}

{\Large \bf T\"{u}rk\c{c}e \.{I}\c{c}in \.{I}m Temelli \"{O}bek Yap{\i}sal S\"{o}zdizimi}\\
\vspace{4em}

\c{S}ehito\u{g}lu, Onur Tolga\\
Y\"{u}ksek Lisans, Bilgisayar M\"{u}hendisli\u{g}i B\"{o}l\"{u}m\"{u}\\
Tez y\"{o}neticisi: Yrd. Do\c{c}. Dr. Cem Boz\c{s}ahin\\

Ocak 1996, \pageref{sonsayfa} sayfa\\
\vspace{4em}

\end{center}

Bu \c{c}al{\i}\c{s}mada, T\"{u}rk\c{c}e s\"{o}zdizimi bilgiye dayal{\i} bir bak{\i}\c{s} a\c{c}{\i}s{\i}yla
de\u{g}erlendirilmi\c{s}tir. \.{I}me dayal{\i} dilbilimsel g\"{o}sterim ve HPSG (
Ba\c{s}-s\"{u}r\"{u}ml\"{u} \"{O}bek Yap{\i}sal Dilbilim) kuram{\i} T\"{u}rk\c{c}e'ye uyarlanm{\i}\c{s}t{\i}r. HPSG, dildeki
nesnelerin bilgisel i\c{c}erikleriyle g\"{o}sterimine dayanan \c{c}a\u{g}da\c{s} bir s\"{o}zdizimi
ve anlambilim kuram{\i}d{\i}r. Temel bilgi \"{o}\u{g}esi im denilen i\c{c}i\c{c}e ve kal{\i}tsal 
t\"{u}rlendirilmi\c{s} \"{o}zellik yap{\i}lar{\i}d{\i}r.

Uygulamada mant{\i}k programlama dili olarak \"{o}zellikle HPSG uygulamalar{\i} i\c{c}in
tasarlanm{\i}\c{s} olan ALE kullan{\i}lm{\i}\c{s}t{\i}r. T\"{u}rk\c{c}e'deki dil \"{o}\u{g}elerinin bir t\"{u}r ve
yap{\i} tan{\i}m{\i} yap{\i}lm{\i}\c{s}t{\i}r. Altulamlama, yan c\"{u}mleler, \"{o}bek s{\i}ra de\u{g}i\c{s}imi,
t\"{u}mle\c{c}-niteleyen ili\c{s}kileri ve orta\c{c} yap{\i}lar{\i} ALE'de \c{c}al{\i}\c{s}an bir ayr{\i}\c{s}t{\i}r{\i}c{\i}
ile tasarlanm{\i}\c{s} ve uygulanm{\i}\c{s}t{\i}r.
\vspace{2em}

\noindent{\bf Anahtar Kelimeler:} s\"{o}zdizimi, T\"{u}rk\c{c}e
Dilbilgisi, ayr{\i}\c{s}t{\i}rma, \"{o}bek yap{\i}s{\i}\\

\newpage
\vspace*{.05\textheight}

\begin{center}
{\Huge \bf
Acknowledgments\\}
\end{center}
\vspace{4em}

I would like to thank NATO TU-LANGUAGE and T\"{U}B\.{I}TAK EEEAG-90 projects for 
providing development environment and research materials. 
Hardware and software resources of the laboratory 
established by NATO (LcsL) have been used in all stages of the preperation 
of the thesis.\\

I would like to thank Dr. Cem Boz\c{s}ahin and Elvan G\"{o}\c{c}men for their
contributions with corrections, discussions and especially for Turkish
syntax chapter.\\

Thanks are also due to all of the friends and family for their 
encouragement, support and friendship during the preperation of the
thesis.

\tableofcontents
\listoftables
\addcontentsline{toc}{chapter}{List of Tables}
\listoffigures
\addcontentsline{toc}{chapter}{List of Figures}
\newpage
\addcontentsline{toc}{chapter}{List of Abbreviations}
\twocolumn[\Huge \bf \vspace{2.4em} 
List of Abbreviations\\ \vspace{1em}]
\begin{description}
\item [{\em 1Sg, 2Sg, 3Sg}]{Agreement suffixes first, second and third
				person singular}
\item [{\em 1Pl, 2Pl, 3Pl}]{Agreement suffixes first, second and third
				person plural}
\item [{\em 1SP, 2SP, 3SP}]{Possessive suffixes first, second and third
				person singular}
\item [{\em 1PP, 2PP, 3PP}]{Possessive suffixes first, second and third
				person plural}
\item[{\em Abl}]{Ablative Case}
\item[{\em Acc}]{Accusative Case}
\item[{\em Dat}]{Dative Case}
\item[{\em Loc}]{Locative Case}
\item[{\em Ins}]{Instrumental/commitative Case}
\item[{\em Gen}]{Genitive Case}
\item[{\em Equ}]{Equitative Case}
\item[{\em Mun}]{Munitive Case}
\item[{\em Rlvz}]{Noun Relativizer}
\item[{\em Cop}]{Copula Suffix}
\item[{\em Aux}]{Auxilary Suffix} 
\item[{\em Pres}]{Present Tense (\tt -Ar)}
\item[{\em Prog}]{Progressive Tense (\tt -Hyor)}
\item[{\em Past}]{Past Tense (\tt -dH, -mH\c{s})}
\item[{\em Fut}]{Future Tense (\tt -(y)AcAk)}
\item[{\em Asp}]{Aspect markers ({\tt -dH, -mH\c{s}, -sA})} 
\item[{\em Pass}]{Passive Suffix} 
\item[{\em Caus}]{Causative Suffix} 
\item[{\em Neg}]{Negation suffix}
\item[{\em Ques}]{Question suffix}
\item[{\em Part}]{Complement Participle Suffix (\tt -DHk, -(y)AcAk)}
\item[{\em Inf}]{Infinitive Suffix (\tt -mAk)}
\item[{\em Ger}]{Gerundive Suffix (\tt -mA, -H\c{s})}
\item[{\em Rel}]{Relative Participle Suffix (\tt -An, -DHk, -(y)AcAk)}
\item[{\em Cond}]{Conditional Suffix (\tt -(y)sA)}
\item[{\em Adv}]{Adverbial Suffix (\tt -ken)}
\item[{\em Nec}]{Necessity Suffix (\tt -mAlH)}
\end{description}
\onecolumn

\newpage
\bibliographystyle{plain}
\setcounter{page}{1}
\pagenumbering{arabic}
\chapter{INTRODUCTION}
\label{ChIntro}

This study has two purposes: first, to study Turkish grammar in light of the
Head-driven Phrase Structure Grammar (HPSG) formalism, and second, to come up
with a computational model of the languages based on the HPSG principles.

Turkish grammar has been analyzed from the perspective of 
linguistic theories such as Transformational Grammar \cite{Meskill70},
Government-Binding, and Functional Grammar
\cite{Schaaik95}. Lewis \cite{Lewis67}, Underhill \cite{Underhill76},
Banguo\u{g}lu \cite{banguog}, and \c{S}im\c{s}ek \cite{Simsek} are also good sources
in the traditional descriptive style. However these studies do not shed
any light on how a computational model can be constructed from the
linguistic description.

Recent linguistic theories, such as HPSG \cite{PolSag87,PolSag94} and
Lexical Functional Grammar (LFG) \cite{Bres82}, differ from the earlier
ones in their rigorous definitions and incorporation of ideas from
computer science and artificial intelligence. These ideas range from
type-theory in programming languages to unification and knowledge
representation. Due to the formal representations, there are meta-tools
for constructing computational models from formal descriptions, such as
Attribute Logic Engine (ALE)\cite{ALE}, CUF \cite{CUF}, and Typed
Feature System (TFS)\cite{TFS}; Tomita's parser for LFG \cite{tomita}.

This work is one of the early attempts, together with LFG 
\cite{Zelal} and Categorial Grammar Models \cite{Hoffman,BozGoc95} to
study Turkish from the perspective of modern linguistic theories. Our
motivation was to design a parser based on the principled account of
Turkish syntax in the HPSG framework. It makes use of the ALE
formalism to model HPSG-style definitions.

HPSG makes universal claims about human languages. The main point is that,
although the grammars of languages differ in terms of phrase structure and
how grammatical functions are realized, certain principles always hold
accross the languages. An example of such a principle roughly states that
the `head' of a phrase plays the most prominent role in propagating the
syntactic and semantic properties of a phrase. Thus, an HPSG grammar for a
language is a collection of specifications for phrase structure,
realization of principles in the language, and the signature of the
language in terms of linguistic features. This division of linguistic
description is also reflected in the computational meta-level tools for
writing HPSG-style grammars. We hope that these kinds of
experiments point out the advantages and disadvantages of such frameworks for
underanalyzed language including Turkish.

We aimed to develop a competence grammar rather than a performance grammar
for Turkish. This requires postulating the Turkish realizations of HPSG
principles and their computational counterparts. We chose to provide a
breadth of coverage in terms of lexical types and phrases instead of a
comprehensive study with a large lexicon. Moreover, we have implemented
some of the morphosyntactic operations (eg. case marking, possessives,
relativization) in the lexicon.

The remainder of the thesis is organized as follows: Chapter~\ref{ChHPSG}
introduces the basic concepts of HPSG. Chapter~\ref{ChTurkish} is an outline
of Turkish syntax. Chapter~\ref{ChDes} describes HPSG model of Turkish and 
Chapter~\ref{ChImp} elaborates on the implementation.


\chapter{HEAD-DRIVEN PHRASE STRUCTURE GRAMMAR}
\label{ChHPSG}

HPSG (Head-driven Phrase Structure Grammar) was introduced by
Pollard and Sag\cite{PolSag87} as an information-based theory of syntax
and semantics . HPSG views a human language as a device used for
exchanging information, and tries to explain the relation between the
phonetic form of a word or a phrase, its grammatical structure, and its
informational content. In HPSG, a natural language is defined as
a system of correspondences between certain kind of utterances and
certain kinds of objects and situations in the world.

HPSG synthesizes most of the recent (principally nonderivational)
syntactic theories such as Categorial Grammar (CG), Generalized Phrase
Structure Grammar (GPSG), Arc Pair Grammar (APG), and Lexical
Functional Grammar (LFG); semantic theories like Situation Semantics,
and some basic concepts of computer science (data type theory,
knowledge representation, unification).

In HPSG, every linguistic component (words, phrases, rules, etc.) is
analyzed with a perspective of the information it provides to the
speaker of the language. This information may include not only the
syntactic features of the component, but also its grammatical
information, semantic content and its background semantics.

HPSG is a system based on {\em signs}. Any structural element (words and
phrases), and principles defining the language are
modeled by {\em sorted feature structures} (ie. feature structures with an
associated type or sort) and constraints and
operations defined on them. As being one of the most recent examples
of the family of the {\em unification based} grammar theories
\cite{Shieber86}, the most fundamental operation of HPSG is unification,
which combines a set of compatible feature structures, and returns a
minimum informative feature structure containing all information
present in the operands.  Phonetic, syntactic and semantic information
coded in the lexicon and information coming from other resources like
lexical rules, universal and language specific principles of
well-formedness, are combined by unification.

Similar to the majority of the contemporary linguistic theories, HPSG
defines a language by a finite set of recursively applicable rules
which yields the judgment of grammaticality. Basically, principles are
divided into two categories: 1) Universally applicable basic set of
constraints such as {\em head feature principle} and {\em
subcategorization principle}; and types of phrases available in any
human language. 2) Language specific principles of phrases lexicon
itself and further articulation and specification of the principles of
the universal grammar such as constituent order\cite{PolSag94}.

One of the distinctive aspects of HPSG is that 
it not only models the language
syntactically, but also concerns itself with the interactions
between all kinds of information of  a linguistic component. Both
syntactic and semantic information content of a sign is 
considered. Situation Semantics and Relational Theory of
Meaning are chosen for semantic modeling.

\section{Feature Structures}

HPSG, like other unification-based formalisms, uses recursively
embedded feature-value pairs representing linguistic objects. Feature
structures have different names in each theory: f-structures in LFG,
feature bundles, feature matrices or categories in GPSG, etc.  Feature
structures are informational objects that consist of
feature (attribute)-value pairs. Usually  feature structures are
represented by {\em attribute-value matrices} (AVM's). For example:

\myenumsentence{
\begin{avm}
\[ PHON & ``kedi'' \; {\em \%   cat}\\
   CAT & noun \\
   AGR & \[ PERSON & {\aS third} \\
	    NUMBER & {\aS singular}
	 \]
\]
\end{avm}
}

In (\ex{0}), features {\aF PHON}, {\aF CAT} and {\aF AGR} are defined
where value of {\aF PHON} is {\aS ``kedi''}, {\aF CAT} is {\aS noun} and
value of {\aF AGR} is another feature structure with features {\aF
PERSON} and {\aF NUMBER} which have values {\aS third} and {\aS
singular} respectively. As an alternative, feature structures can be
represented in graph notation where nodes are the intermediate feature
structures, vertices are attributes, and values are sink nodes.

\newsavebox{\tmp}
\savebox{\tmp}{\psfig{file=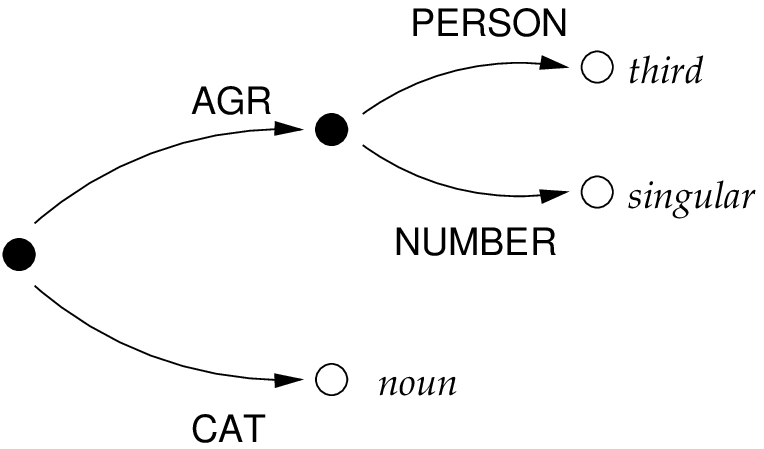,width=6cm}}

\enumsentence{
\hspace{1em}\par
\usebox{\tmp}
}

As a fundamental property, feature structures can be recursively
embedded.  Value of an attribute can be an atomic value or another
feature structure. To represent values embedded in feature structures,
``path of attributes'' notation is used as a shorthand. A path is an
ordered sequence of attributes separated by `$\vert$' to reach the
value. In the example (\ex{0}), {\aF AGR$\vert$PERSON} has the value {\aV
third}, {\aF AGR$\vert$NUMBER} is {\aV singular}.

A relation defined on feature structures is the {\em subsumption} relation.
When a feature structure $A$ is subsumed by another feature structure $B$, 
$A$ is equally or more informative than $B$. In other words, it contains
all of the information provided by $B$ and possibly more. It is often
said $A$ {\em extends} $B$ or $B$ {\em subsumes} $A$ and written as $A
\preceq B$. It indicates that any object described by $B$ can be
appropriately described by $A$. When a feature structure has no
information, it {\em subsumes} every feature
structure. It can describe any object whatsoever. This structure is the
root element of the subsumption ordering (called {\em Top}), shown as
$\top$:\\
$\top = [ \hspace{0.5cm} ]$\\ 

Subsumption relation defines a partial ordering between information
structures. It has the properties of reflexivity ($\forall A, A \preceq A$),
transitivity ($A \preceq B$ and $B \preceq C$ then $A \preceq C$), and
antisymmetricity ($A \preceq B$ and $B \preceq A$ then $A = B$). For
subsumption relation to hold between two feature structures, they
should have compatible types and compatible values in the corresponding
attributes. In the example below, feature structures have incompatible values
so, $A \not\preceq B$ and $B \not\preceq A$.
\myenumsentence{
\begin{avm}
$A$ = \[ PERSON & third 
      \], \hspace{0.5cm} $B$ = \[ PERSON & first \\
      				  NUMBER & singular \]
\end{avm}
}

Another important property of feature structures is {\em structure sharing}.
Two attribute paths in a feature structure may describe the same object. This
implies the {\em token identity} which should not be confused with the {\em
structural identity} where only type structure and feature values are equal.
Usually structure shared object are denoted by {\em tags} (boxed numbers).

\myeenumsentence{ \item
\begin{avm}
\[ HD-DTR & \[ CAT & {\aS verb} \\
	       AGR & \@1 \[ PERSON & {\aS first} \\
			    NUMBER & {\aS plur} \] 
	    \] \\
   SUBJ-DTR & \[ CAT & {\aS noun} \\
	         AGR & \@1 
	      \]
\]
\end{avm}
\item
\begin{avm}
\[ HD-DTR & \[ CAT & {\aS verb} \\
	       AGR & \[ PERSON & {\aS first} \\
		        NUMBER & {\aS plur} \] 
	    \] \\
   SUBJ-DTR & \[ CAT & {\aS noun} \\
	         AGR & \[ PERSON & {\aS first} \\
		           NUMBER & {\aS plur} \] 
	      \]
\]
\end{avm}
}

In (\ex{0}a), {\aF HD-DTR$\vert$AGR} attribute shares the same object
with the {\aF SUBJ-DTR$\vert$AGR} attribute. Although the paths in
question have the same value, the source of the values may be different
(i.e., not token identical) contain the same agreement in (\ex{0}b).
Intuitively, it is clear that the structure shared version is more
informative than the other; it is subsumed by the other.
Similarly, the effect of structure sharing is reflected in the formal
definition of the subsumption:
\myeenumsentence{ \item[]
if $A$ and $B$ are atomic, then $A \preceq B$ iff $A = B$.
\item[]
else, $A \preceq B$ holds iff,
\begin{enumerate}
\item[(i)] for every path in $B$, same path exists in $A$ and its value is
subsumed by the value in $B$.
\item[(ii)] for every structure sharing path in $B$, same path is structure
sharing in $A$.
\end{enumerate}
}

Perhaps the most important operation on feature structures is the {\em
unification}, which constructs a base to a group of linguistic
theories.  Unification operation builds a new feature structure which
contains all but not more of the information contained in its operand
feature structures. For feature structures to be unified, they should
have compatible types. Result of unification is the least informative
(the most general) feature structure which extends all of the operands.
Unification is denoted by the symbol $\wedge$ and if $C = A \wedge B$ then, 
$C \preceq A$ and $C \preceq B$.

\myenumsentence{
\begin{avm}
\avml
     \[ CAT & {\aS noun} \\
        AGR & \[ PERSON & {\aS third} \]
     \] $\bigwedge$ 
     \[ CAT & {\aS noun} \\
        AGR & \[ NUMBER & {\aS sing}\]
     \] $=$ \\
     \hspace*{4cm}  
     \[ CAT & {\aS noun} \\
        AGR & \[ PERSON & {\aS third} \\
                 NUMBER & {\aS sing}\]
     \]
\avmr
\end{avm}
}

When operands of the unification have incompatible types or values, the
resulting feature structure does not exist and unification fails. This is
indicated by the symbol `$\bot$' ({\em bottom}) which represents inconsistent
information. As $\top$ is the maximal element in the subsumption ordering,
$\bot$ is the minimal element that is subsumed by all feature
structures.

Also, unary negation operator $\neg$ and disjunction operator $\vee$ are
defined. In negation, $\neg a$ means any value other than $a$.
Similarly, disjunction $a \vee b$ means the attribute may be equal to
$a$ or $b$. Attributes may be {\em list}-or {\em set}-valued. Lists are
denoted by comma separated values enclosed in angle brackets, $\langle
a, b,...  \rangle$. Sets are enclosed in curly braces $\{ a, b,... \}$.
List valued attributes are unified by unifying corresponding elements
by order.  Unification of set values is a more complex operation. For
detailed information and formal definition about feature
structures, consult Rounds and Kasper\cite{RouKas86}.

The most significant formal property of HPSG feature structures is that they
are {\bf sorted}. Every feature structure has a type (sort), and a subtype
relation is defined between these sorts. All defined sort symbols are 
partially ordered by the subsumption relation. 

\begin{figure}[ht]
\centerline{\psfig{file=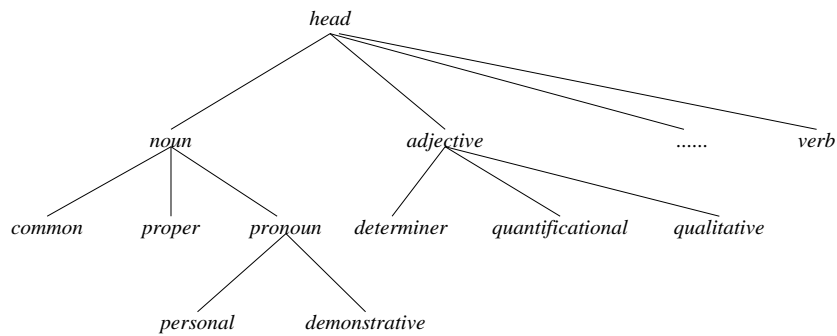,width=11cm}}
\caption{Subtype hierarchy for the type defined for {\aF HEAD} feature}
\label{head_tree}
\end{figure}

As the second formal property, HPSG feature structures should be {\em
totally well-typed}. For each sort, a set of appropriate features and
types is defined, and this set is inherited by the subsorts of the
sort. For example, if the {\aF CASE} feature of sort {\aS case} is defined
for the sort {\aS noun} in Figure~\ref{head_tree}, then it is appropriate
for the sorts {\aS common, proper, pronoun, personal} and {\aS
demonstrative} sorts. Any other feature which is not introduced in the
sort is not allowed in the feature structure.

Third, HPSG feature structures should be {\em sort-resolved} to satisfy the
criteria of completeness as models of the linguistic entities. {\em
Sort-resolved} means: for every attribute defined, a sort should be assigned.
This sort should be the most specific in the sort ordering (A leaf node in
the subtype hierarchy). For example {\aF HEAD} feature can be assigned
{\aV proper} or {\aV common} but not {\aV head} or {\aV noun} which subsume
other types in the ordering and actual sort (value) is not clear.

\section{Sign Structure}

Linguistic entities have the general sort {\aS sign}. Information in
all intermediate phrases, lexical entries, sentences and even
multisentence discourses are described by a corresponding {\aS sign}.
The {\aS sign} sort has two subtypes: {\aS word} and {\aS phrase}. {\aS
word} describes lexical entries, and {\aS phrase} describes phrasal
constructs. {\aS phrase} has an additional feature {\aF DTRS}
(daughters) to represent the phrase structure.

\begin{figure}[htb]
\centerline{\psfig{file=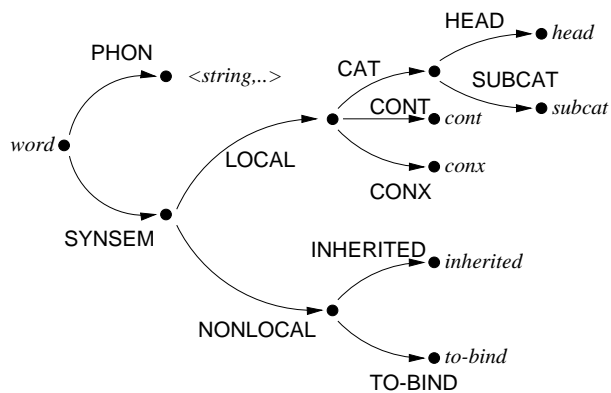,width=8cm}}
\caption{Basic Structure of a Lexical Sign (word).}
\label{sign_graph}
\end{figure}

A basic graph briefly describing the structure of a {\aS sign} can be
given as in Figure ~\ref{sign_graph}. All signs should have at least
two attributes:  {\aF PHON} and {\aF SYNSEM}. {\aF PHON} attribute is a
feature representation of the phonetic content of the {\aS phrase} or {\aS
word}. Usually, it has a list of strings describing phonological and
phonetic structure of the {\aS sign}. {\aF SYNSEM} attribute contains
both syntactic and semantic information of the {\aS sign}. Using {\aF
SYNSEM} instead of two distinct features {\aF SYNTAX} and {\aF
SEMANTICS} allows packing all information required for
subcategorization into one attribute.

{\aF SYNSEM} value is another structured object which has two attributes
{\aF LOCAL} and {\aF NONLOCAL}. {\aF NONLOCAL} represents the information
which is not bound to the phrase described by the {\aS sign}. This information
is used to handle unbounded dependency constructs like filler-gap
dependencies, relative clauses, etc. {\aF LOCAL} feature describes the local
information which consists of the attributes  CATEGORY {\aF (CAT)}, 
CONTENT {\aF (CONT)} and CONTEXT {\aF (CONX)}.

{\aF CAT} value includes both the syntactic category of the {\aS sign} and
the grammatical arguments it requires. {\aF CONT} value is the context
independent semantic interpretation and semantic contribution of the
{\aS sign}. 

{\aF CONX} value contains context-dependent linguistic information such as
indexicality, presupposition and conventional implication. The semantic 
features are not the object of this study so we will not go into
details of {\aF CONT} and {\aF CONX} features.

{\aF CAT} attribute consists of two attributes {\aF HEAD} and {\aF
SUBCAT}.  {\aF HEAD} feature is roughly lexical category (part of
speech) of the {\aS sign}. It describes the information to be passed to
phrasal projections of the {\aS sign}. Contents of {\aF HEAD} feature
varies according to the category of the {\aS sign}. It typically
contains basic features related with the category of the sign e.g.
case, agreement, verb form, prepositional form, etc.

{\aF SUBCAT} (subcategorization) describes the valence of the sign which specifies the group
of signs that sign in question requires to become {\em saturated}.
A saturated sign means all the subcategorization requirements are met.
Group of signs in {\aF SUBCAT} feature is described by a list of {\aS synsem} 
values. {\aS synsem} values are used for identifying the subcategorized
objects, so that {\aS sign} can select not only the syntactic category
of the complement, but also semantic role and even nonlocal
attributes.

The order of the {\aS synsem} values in the {\aF SUBCAT} list does
not correspond to the surface order of the phrase. However, it may
define an obliqueness order which can be used to describe the constituent
order. For example, in English, linear precedence rules defining surface
order is declared by this obliqueness order, as the least oblique
element linearly precedes others for non-verbal heads. When the head 
is a verb, the first oblique element is the subject element which
precedes the head. For languages having free constituent order, like
Turkish, {\aF SUBCAT} attribute may have different structure, e.g., 
unordered list.

\section{Phrases and Syntactic Structure}

HPSG is a constraint-based theory and constraints are defined by
partial descriptions that model linguistic utterances. Descriptions are
declarative, order independent and reversible. Judgment of whether a
phrase is well-formed or not is done by a set of universal principles
and language-specific rules. Universal principles are general
constraints on universally available phrase types. The most basic
principles in HPSG are {\bf Head Feature Principle} and {\bf
Subcategorization Principle}. Language specific phenomenon like {\bf\em
Linear Precedence} (constituent order) is described by a set of
language specific constraints and some kind of specialization of
universal principles.

As mentioned in the preceding section, a {\aS sign} has two subtypes: {\aS
word} and {\aS phrase}. {\aS phrase} has an additional feature DAUGHTERS
{\aF (DTRS)} in which phrase structure is represented. {\aF DTRS} feature has a
value of constituent-structure {\aS (cons-struc)} representing the immediate
constituents of the phrase. {\aS cons-struct} may have several subsorts each
has characterized by different daughter attribute. The most general sort of
{\aS comp-struc} is headed-structure {\aS (head-struct)}.

\myenumsentence{
\begin{avm}
\[ {\aS head-struc} \\
   HEAD-DTR & {\aS sign} \\
   COMP-DTRS & \< {\aS sign, ....} \>
\]
\end{avm}
}

Each {\aS head-struc} has one {\aF HEAD-DTR} attribute and another attribute
which is a list of signs which are the sisters of the {\aF HEAD-DTR}. For
example tree and {\aF DTRS} representation of the sentence ``Ahmet
k{\i}rm{\i}z{\i} kitab{\i} ald{\i}.'' (Ahmet took the red book.) is:

\savebox{\tmp}{
{\scriptsize 
\begin{avm}
\<
{\rm N[Ahmet]},\[ {\aS sign} \\
	          SYNSEM \; {\rm NP} \\
	          DTRS \; \[ {\aS head-struc} \\
		  	    HEAD-DTR & {\rm N[kitab{\i}]} \\
		            ADJ-DTRS & \< {\rm Adj[k{\i}rm{\i}z{\i}]} \>
		         \]
	       \]
\>
\end{avm}
}
}

\myeenumsentence{
\item
\begin{tabular}[t]{l}
 \\
\leaf{Ahmet}
\branch{1}{NP}
	\leaf{k{\i}rm{\i}z{\i}}
	\branch{1}{Adj}
	\leaf{kitab{\i}}
	\branch{1}{N}
\branch{2}{NP}
\leaf{ald{\i}}
\branch{1}{V}
\branch{3}{S}
\tree
\end{tabular}
\item
{\scriptsize
\begin{avm}
\[ {\aS phrase}\\
    SYNSEM \; {\rm S[fin]} \\
    DTRS \; \[ {\aS head-struc} \\
	      HEAD-DTR & {\rm V[ald{\i}]} \\
	      COMP-DTRS & \usebox{\tmp}
	   \]
\]
\end{avm}
}
}

The {\aF HEAD} value of a phrase is centrally important since it
defines the syntactic properties of the mother phrase. For example, the
lexical head of a sentence is of the sign {\aS verb}. {\aS verb}
combines with its complement sisters and forms a {\aV Verb Phrase (VP)}
which takes its syntactic properties from its head daughter (verb).
Similarly, verb phrase combines with the subject complement forming a
sentence. In other phrase types like Noun Phrase (NP), Prepositional
Phrase (PP), Adjective Phrase (AP), {\aF HEAD} feature is projected
---propagated--- along the upper phrases until phrase becomes
saturated. The key idea behind this projection is the {\bf
X-bar} theory\cite{Jackend77}. HPSG's {\bf Head Feature Principle}
describes this syntactic phenomena which is adopted from the {\bf Head
Feature Convention} of GPSG\cite{Gazdar85}.

\myenumsentence{
{\scriptsize
\begin{avm}
\avml  \avmspan{\hspace*{2cm} \node{a}{\[ LOCAL\|CAT\|
	\[ HEAD & \@3 \\
	   SUBCAT & \< \>
	\]
      \]}} \\
      & \\
      & \\
\node{b}{\@1 {\rm NP[nom]}}& \node{c}{\@2 {\rm NP[acc]}} \; \node{e}{
	\[ {\aF LOCAL\|CAT\|}
	\[ {\aF HEAD} & \@3 {\rm verb[fin]} \\
	   {\aF SUBCAT} & \< \@1 {\rm NP[nom]}, \@2 {\rm NP[acc] }
	   \>
	\]
	\]}\\
	\\
	\\
	\\
\node{g}{{\rm Ahmet}} & \node{h}{\rm {kalemi}} \hspace*{4cm} 
					\node{i}{\rm {ald{\i}.}} 
\avmr
\nodeconnect{a}{b}
\nodeconnect{a}{c}
\nodeconnect{a}{e}
\nodeconnect{b}{g}
\nodeconnect{c}{h}
\nodeconnect{e}{i}
\end{avm}
}
}

{\bf Head Feature Principle}(HFP) is defined as follows:
\myenumsentence{
The {\aF HEAD} value of a headed phrase is structure-shared with the {\aF HEAD}
value of the head daughter.\\
Formally:\\
\begin{avm}
\[ {\aS phrase} \\
   SYNSEM\|LOC\|CAT\|HEAD \; \@1 \\
   DTRS\|HEAD-DTR\|SYNSEM\|LOC\|CAT\|HEAD \; \@1
\]
\end{avm}
}

The other principle which together with HFP describes the basic Immediate
Dominance (ID) scheme of HPSG is {\bf Subcategorization Principle}. 
Subcategorization checks the requirements of the phrasal head to be saturated
and allows heads to select its complement sisters by structure sharing the
{\aF SYNSEM} values of the sisters with that  in the {\aF SUBCAT} list.
{\bf Subcategorization Principle} is defined as follows:
\myenumsentence{
In a headed phrase, {\aF SUBCAT} value of the head daughter of the
phrase is the concatenation of the {\aF SYNSEM} values of the complement
daughters.\\
Formally:\\
\begin{avm}
\[{\aS phrase}\\
  SYNSEM\|LOC\|CAT\|SUBCAT \@1 \\
  DTRS \; \[ HEAD-DTR\|SYNSEM\|LOC\|CAT\|SUBCAT \; \@2 $\oplus$ .... 
							$\oplus$ \@n $\oplus$
							\@1 \\
	    COMP-DTRS \; \< SYNSEM \@2, ...., SYNSEM \@n \>
	 \]
\]
\end{avm}\\
Where $\oplus$ is defined to be list concatenation operation.
}

The {\bf Subcategorization Principle} allows all constraints on the
arguments of a phrase to be controlled by an argument. Any kind of
argument restriction, complement structure like sentential complements,
unbounded dependencies and other constraints can be directly controlled
and coded into lexicon. In other words, HPSG crucially relies on the
complex descriptions in the lexicon. To deal with the redundancy caused
by the complexity of the lexical entries, lexical rules and multiple
inheritance hierarchy describing relation between lexical entries can
be expressed~\cite{PolSag87}.

Phrase structure rules defining tree structure of phrases are described
by immediate dominance (ID) and linear precedence (LP)
constraints. There is a general trend in contemporary syntactic
theories towards the lexicalization of grammar and elimination of
construction-specific rules in favor of schematic immediate dominance
templates. These schemata may vary for language-specific phrase types
and constituent relations. Examples of typical phrase structures are
head-complement, specifier-head, and adjunct-head, conjunct-daughters.

Linear precedence constraints are mostly defined as language-specific rules
and constraints on the surface constituent order of the phrases. In
English, LP rules are defined on the obliqueness hierarchy of the {\aF
SUBCAT} list. Subject is the least oblique argument of a
verb. The direct object and the indirect object come next in the
obliqueness order.  Also, least oblique constituents precede the
others. LP rules for English can be defined as follows:

\myeenumsentence{
\item[1.]
Any lexical head sign precedes other signs:\\
\begin{avm}
\[ HEAD-DTR \@1 word \] $\Longrightarrow$ \[ \@1 $\leq$  \[ \hspace{0.3cm} \]
					  \]
\end{avm}
\item[2.]
Subject complement precedes the Head daughter:\\
\begin{avm}
\[ SUBJ-DTR \; \@1 \] $\Longrightarrow$ \[ \@1 $\leq$ \[ {\aS phrase} \] \]
\end{avm}
\item[3.]
Least oblique elements linearly precede the others:\\
\begin{avm}
\[ COMP-DTRS \< ...,\@1, ....,\@2,...\> \] $\Longrightarrow$ \[ \@1 $<$ \@2 \]
\end{avm}\\
where $\leq$ means {\em immediately precede} and $<$ means {\em precede}.
}

\section{Lexical Organization}
\label{SeLex}

Lexicalization and the use of meta-rules controlled by a set of
universal principles results in a few number of simple grammar
rules. However, the associated information structures become more
complex.  Lexical signs in lexicaly-oriented theories should very rich in
information content so it is not always possible to enter and maintain a
lexicon without any organization.

In HPSG ---as the other lexicalized formalisms--- it is necessary to
organize lexicon such that lexical entries should be represented as
compact as possible. Two main devices, lexical type hierarchy and
lexical rules, are the basic solutions to redundancy problem in the
lexicon.

The main idea behind the lexical type hierarchy is the repetition of
information in the lexical items of same category class. Only a small
part of a lexical entry carries exceptional information from the other
entries having the same category. For example, all nouns have {\aS
noun} as the value of the {\aF HEAD} feature and empty {\aF SUBCAT}
value. All common nouns have {\em third person} in their agreement. So
the idea is to create a hierarchy of types each of that is assigned a
set of attribute-value pairs which are inherited along the hieararchy.
Lexical entries can be defined by means of these types plus the
special features.

Lexical hierarchies solve some portion of the redundancy. However, some
specific features of lexical entries may be related to each other by
recurrent patterns. These patterns include some derivational and
inflectional phenomena in the language like passivization of verbs,
case marking, verb inflections, nominalization etc. The solution that
has been used by most of the unification-based formalisms is to define
functions mapping one class of words to another, called {\em lexical
rules}.

Lexical rules are generally expressed as procedures converting an input
form to an output form. So that all inflections and derivations of a
word can be generated from a base form by application of lexical rules
several times. In the example (\ex{1}a), a simplified lexical rule for
passivization of verbs for English is given where $f_{PSP}$ is the
function mapping verb base to its past-participle form. Also, it is
possible to generate different readings and syntactic behaviours of the
same word.  Lexical rule in~(\ex{1}b) duplicates lexical entry of verbs
for non-referential objects in preverbal position in Turkish.

\myeenumsentence{
\item[a.] {\scriptsize \begin{avm}
\avml
\[ {\aS word} \\
     PHON \; \@1\\
     SYNSEM\|LOCAL \; \[ CAT \; \[ HEAD \; \[ {\aS verb}\\
					      TENSE & base\\
					      PASSIVE & $-$\\
					   \] \\
				   SUBCAT \; \< ... [ \; ]$_{\@2}$,
						 [ \; ]$_{\@3}$
						 ...\>\\
				   SUBJ  \; \@2
				\] \\
			CONT \; \@4
		      \]
\] \; $\longmapsto$ \\ \hspace{3cm}
\[ {\aS word} \\
   PHON \; $f_{PSP}(\@1)$ \\
   SYNSEM\|LOCAL \; \[ CAT \; \[ HEAD \; \[ {\aS verb}\\
					    PASSIVE & $+$\\
					 \] \\
				 SUBCAT \; \< PP[BY]$_{\@2}$, ...,
					 [ \; ]$_{\@3}$
						 ...\>\\
				 SUBJ  \; \@3
			      \] \\
		      CONT \; \@4
		    \]
\]
\avmr
\end{avm}
}
\item[b.] {\scriptsize \begin{avm}
\avml
\[ {\aS word} \\
     PHON \; \@1\\
     SYNSEM\|LOCAL\|CAT \; \[ HEAD \; {\aS verb}\\
                              SUBCAT \; \{ NP$_{acc}$[    ],
					\@2 (....)
                                                 \}
                           \] 
\]  \; $\longmapsto$ \\ \hspace{3cm}
\[ {\aS word} \\
   PHON \; \@1 \\
   SYNSEM\|LOCAL\|CAT \; \[ HEAD \; {\aS verb}\\
                            SUBCAT \; \< \{ \@2 \} , NP$_{nom}$[     ] \>
			 \]
\]
\avmr
\end{avm}
}
}


\chapter{TURKISH SYNTAX OVERVIEW}
\label{ChTurkish}

This chapter is adopted from~\cite{SyntaxTR}. Turkish is an
agglutinative language where words are formed by affixation of
derivational and inflectional morphemes to root words. So most of the
syntactic properties of a word such as case, agreement, relativization
of nouns, tense, modality, aspect of verbs, and even passivization,
negation, causatives, reflexives and some auxiliaries are marked by
suffixes.

\myeenumsentence{
\item \begin{tabular}[t]{lllll}
ev&-imiz&-de&-ki&-nin \\
house&-{\em 1PP}&-{\em Loc}&-{\em Rlvz}&-{\em Gen} \\
\multicolumn{5}{l}{`of the one that is in our house'}
\end{tabular}
\item \begin{tabular}[t]{lllllll}
bak&-t{\i}r&-a&-m{\i}&-yor&-mu\c{s}&-sun \\
look&-{\em Caus}&-{\em Able}&-{\em Neg}&-{\em Prog}&-{\em Asp}&-{\em 2Sg} \\
\multicolumn{7}{l}{`you were not able to make look (reported past)'}
\end{tabular}
}

As a result, Turkish words ---especially heads of phrases--- have complex and
rich syntactic forms and carry much information. 

As another distinct property, Turkish is head-final. Specifiers and
modifiers always precede the specified or modified. Similarly complements
and arguments precede the head in their usual formation. However when
head is a verb or predicative noun, complements and objects may follow the 
head.

\myeenumsentence{ 
\item \shortex{4}{Benim & kap{\i}daki & k{\i}rm{\i}z{\i} & \uline{arabam}}
		 {ben-{\em Gen} & door-{\em Loc-Rlvz} & red & car-{\em 1SP}} 
		 {`my red car at the door'}
\item \shortex{5}{Hicabi & kitab{\i} & \c{c}ok & \c{c}abuk & \uline{okudu}.}
		 {Hicabi & book-{\em Acc} & very & quick & read-{\em Past-3Sg}}
		 {`Hicabi read the book very quickly.'}
\item \shortex{3}{Kitaplar{\i} & \uline{verdim} & Ahmet'e.}
		 {book-{\em Plu} & give-{\em Pass-1Sg} & Ahmet-{\em Dat}}
		 {`I gave Ahmet the books.'}
}

Also in Turkish, constituents have free order. The most usual sentence order is
S-O-V. However they can scramble causing different readings and
interpretations. Sentence-initial position marks the {\em topic},
pre-verbal constituent is the {\em emphasis} and post-verbal position
is for the {\em background} or {\em afterthought} information~\cite{Erguvan}.

\myeenumsentence{ \item 
\shortex{4}{Onur & kalemi & \c{c}ocu\u{g}a & verdi.}
	   {Onur & pencil-{\em Acc} & child-{\em Dat} & 
			give-{\em Past}-{\em 3Sg}}
	   {`Onur gave the child the book.'}
\item Onur \c{c}ocu\u{g}a kalemi verdi. \\
      `Onur gave \uline{the pencil} to the child.'
\item  \c{C}ocu\u{g}a kalemi Onur verdi.\\
       `It is Onur who gave the child the book.'
\item Kalemi Onur verdi \c{c}ocu\u{g}a. \\
      (c)
\item Onur verdi kalemi \c{c}ocu\u{g}a. \\
      `Onur did give the child the pencil.'
}

When the object is non-referential (ie. no case marked or specified), it 
should immediately precede the verb.

\myeenumsentence{ \item
\shortex{4}{Adam & bah\c{c}ede & \c{s}iir & yaz{\i}yordu}
	   {man & garden-{\em Loc} & poem & write-{\em Prog-Asp-3Sg}}
	  {`The man was writing poem in the garden'}
\item (*) \c{S}iir bah\c{c}ede adam yaz{\i}yordu.\\
      `The poem was writing the man in the garden'
\item * Adam \c{s}iir bah\c{c}ede yaz{\i}yordu.
}

Similarly, adverbs and sentential complements may scramble freely
(\ex{1}a--c).  Also order variation of constituents is valid for the
embedded sentences such as relative clauses, infinitive and gerundive
forms, and sentential complements. Relative clauses are strictly
head-final; no constituent belonging to relative clause can follow
the head verb (\ex{1}d--f).

\myeenumsentence{ \item 
\shortex{6}{Ger\c{c}ekten & onun & s{\i}nav{\i}& kazanmas{\i}n{\i} & herkes & istemi\c{s}ti.}
	   {really & he-{\em Gen} & exam-{\em Acc} & 
			pass-{\em Inf-3Sg-Acc} & everyone &
	   			want-{\em Past-Asp}}
	   {`Everybody realy wanted him to pass the exam.'}
\item  Onun s{\i}nav{\i} kazanmas{\i}n{\i} herkes ger\c{c}ekten istemi\c{s}ti.
\item Herkes onun s{\i}nav{\i} kazanmas{\i}n{\i} ger\c{c}ekten istemi\c{s}ti.
\item \shortex{5}{Bakkaldan & d\"{u}n & ald{\i}\u{g}{\i}m & kalem & k{\i}r{\i}ld{\i}}
		 {store-{\em Abl} & yesterday & buy-{\em Rel-1Sg} & pencil &
		 break-{\em Pass-Past-3Sg}}
		 {`The pencil that I bought yesterday from the store was
		 broken'}
\item D\"{u}n bakkaldan ald{\i}\u{g}{\i}m kalem k{\i}r{\i}ld{\i}.
\item * D\"{u}n ald{\i}\u{g}{\i}m bakkaldan kalem k{\i}r{\i}ld{\i}.
}

\section{Noun Phrase}

Phrases with nominal heads are noun phrases. The head noun is the final
constituent of the phrase and determines the syntactic role of the whole
phrase. Noun phrases may act as a subject, object or complement of a sentence
or modifier or specifier of another noun group. A noun ---so a noun phrase---
can have the cases listed in Table~\ref{nouncases}.\footnote{We use
{\tt A} to stand for {\tt a} or {\tt e}, {\tt H} to stand for {\tt {\i},
i, u or \"{u}}, and {\tt D} to stand for {\tt d} or {\tt t}.}

\begin{table}[h]
\caption{Cases for Turkish nouns}
\label{nouncases}

\begin{center}
\begin{tabular}{|l|l|l|}\hline
case & suffix & examples \\ \hline
nominative  &  & adam, kedi\\
accusative &{\tt -(y/n)H} & adam{\i}, kediyi\\
dative/allative &{\tt -(y/n)A} & adama, kediye\\
locative &{\tt -(n)DA} & adamda, kedide\\
ablative &{\tt -(n)DAn} & adamdan, kediden\\
genitive &{\tt -(n)Hn} & adam{\i}n, kedinin\\
comitative/instrumental &{\tt -(y)lA} & adamla, kediyle\\ \hline
\end{tabular}
\end{center}

\end{table}

Also three suffixes {\tt -cA}, {\tt -lH} and {\tt -sHz} (equative, munitive
and privative respectively) are considered as cases by Banguo\u{g}lu
\cite{banguog}.

Nominative case is used for marking subjects (\ex{1}a),
indefinite/nonreferential objects (\ex{1}b). Also noun with the nominative
case can be a classifier for another noun~(\ex{1}c). 

\myeenumsentence{
\item \shortex{3}{{\em K\"{o}pek} & kediyi & kovalad{\i}.}
                    {dog & cat-{\em Acc} & chase-{\em Past}-{\em 3Sg}}
                    {`The dog chased the cat.'}
\item \shortex{3}{Adam & {\em ku\c{s}} & avlad{\i}.}
                    {man & bird & hunt-{\em Past}-{\em 3Sg}}
                    {`The man hunted a bird.'}
\item \shortex{5}{G\"{u}zel & bir & {\em k\"{o}pek} &  evi & yapt{\i}k.}
                    {nice & a & dog &  house & make-{\em Past-1Pl}}
                    {`We made a nice dog house.'}
}

The accusative case is used for marking definite objects. It is
obligatory with pronouns and proper nouns in object position.

\myeenumsentence{
\item[a.]\shortex{3}{\c{C}ocuk & kitab{\i} & okumam{\i}\c{s}.}
                   {child & book-{\em Acc} & read-{\em Neg-Past-3Sg}}
                   {`The child hasn't read the book.'}
\item[b.]\shortex{3}{K\"{o}pek & Ay\c{s}e'yi & {\i}s{\i}rd{\i}.}
                    {dog & Ay\c{s}e-{\em Acc} & bite-{\em Past-3Sg}}
                    {`The dog bit Ay\c{s}e.'}
\item[c.]\shortex{3}{Herkes & onu & su\c{c}luyor.}
                    {everyone & he/she-{\em Acc} & blame-{\em Prog-3Sg}}
                    {`Everyone blames him/her.'}
}

Noun phrases with dative/allative case ({\tt -(y/n)A} suffix) have
three roles: they behave as prepositional phrases indicating target or
aim (\ex{1}a--b), mark the indirect object (\ex{1}c), and they are
subcategorized as the oblique object in some verbs (\ex{1}d).

\myeenumsentence{
\item \shortex{3}{\c{C}ocuklar{\i} & Ankara'ya & g\"{o}nderdik.}
                    {child-{\em Plu}-{\em Acc} & Ankara-{\em Dat} & send-{\em
		    Pass-2Pl}}
                    {'(We) sent the children to Ankara.'}
\item \shortex{3}{\c{C}i\c{c}ekleri & sana & ald{\i}m.}
                    {flower-{\em Plu}-{\em Acc} & you-{\em Dat} & buy-{\em
		    Past-1Sg}}
                    {'(I) bought the flowers for you.'}
\item \shortex{4}{Mehmet & ekme\u{g}i & adama & verdi.}
                    {Mehmet & bread-{\em Acc} & man-{\em Dat} & give-{\em
		    Past-3Sg}}
                    {`Mehmet gave the man the bread.'}
\item \shortex{3}{Kad{\i}n & bah\c{c}eye & bakt{\i}.}
                    {woman & garden-{\em Dat} & look-{\em Past-3Sg}}
                    {`The woman looked at the garden.'}
}

Noun phrases with locative case ({\tt -DA} suffix)  is used to
express the location of an action or object (\ex{1}).

\myenumsentence{ 
\shortex{3}{Kitab{\i}n & masada & duruyor.}
	   {book-{\em 2SP} & table-{\em Loc} & stand-{\em Prog}-{\em 3Sg}}
	   {`Your book lays on the table'}
}

The ablative case ({\tt -DAn} suffix) indicates the source of an action
or object as the English preposition ``from'' (\ex{1}a--b). Also can be
subcategorized as direct object by a group of verbs (\ex{1}c) .

\myeenumsentence{
\item \shortex{3}{\.{I}stanbul'dan & yeni & gelmi\c{s}.}
		 {\.{I}stanbul-{\em Abl} & just & come-{\em Past}-{\em 3Sg}}
		 {`He has just come from \.{I}stanbul'}
\item \shortex{5}{Genelde & bu & \"{u}z\"{u}mlerden & \c{s}arap & yap{\i}l{\i}yor}
		 {usually & these & grape-{\em Plu}-{\em Abl} & wine & 
					make-{\em Pass-Prog-3Sg}}
		 {`Usually wine has been done from these grapes'}
\item \shortex{4}{Ahmet&kedilerden&nefret eder.}
		 {Ahmet&cat-{\em Plu-Abl}&hate-{\em-Pres-3Sg}}
		 {`Ahmet hates cats.'}
}

The genitive case is used to mark the possessor in the
possesive-possessor relation. Noun with the genitive case behaves as a
specifier of possessed noun which is marked with the possessive suffix.
Person and number information of the noun should agree with this possesive
suffix.

\myeenumsentence{
\item \shortex{4}{Araban{\i}n & anahtar{\i}n{\i} & unuttum.}
		 {car-{\em Gen} & key-{\em 3SP}-{\em Acc} & forget-{\em Past}-{\em 1Sg}}
		 {`I forgot the key of the car.'}
\item \shortex{3}{Senin & kalemini & kulland{\i}m.}
		 {you-{\em Gen} & pencil-{\em 2SP}-{\em Acc} & use-{\em Past}-{\em 1Sg}}
		 {`I have used your pencil.'}
\item \shortex{4}{\.{I}lker'in & arabas{\i}n{\i}n & motoru & bozuk}
		 {\.{I}lker-{\em Gen} & car-{\em 3SP}-{\em Gen} & engine-{\em 3SP} & broken}
		 {`The engine of the \.{I}lker's car is broken.'}
}

The {\tt -(y)lA} suffix is the combined form of the postposition
{\bf ile} with the noun. It marks the commutative (\ex{1}a) and instrumental
(\ex{1}b) relationships.

\myeenumsentence{
\item \shortex{3}{Kitab{\i} & Ahmet'le & g\"{o}rd\"{u}k.}
		 {book-{\em Acc} & Ahmet-{\em Ins} & see-{\em Past-1Pl}}
		 {`Ahmet and I saw the book together.'}
\item \shortex{4}{Ku\c{s}lar{\i} & d\"{u}rb\"{u}nle & seyrediyoruz.}
		 {bird-{\em Plu}-{\em Acc} & binocular-{\em Ins} & 
						watch-{\em Prog-1Pl}}
		 {`We could see the birds with telescope.'}
}

{\tt -cA} suffix is used for marking subject of a passive sentence.
Postposition {\bf taraf{\i}ndan} is more commonly used compared to
equative case.

\myeenumsentence{
\item \shortex{3}{Kampanya & vatanda\c{s}larca & destekleniyor.}
		 {campaign & citizen-{\em Plu-Equ} & support-{\em Pass}-{\em Prog}-{\em 3Sg}}
		 {`The campaign is supported by citizens.'}
}

{\tt -lH} and {\tt -sHz} (munitative and privative) suffixes have similar
meaning with the prepositional phrases formed by `with' and `without' in
English. Noun phrases with these suffixes behave as adjective.
However, {\tt -lH} suffix saves some of the properties of the noun it
is attached to. Noun may be still the head of a noun group and can be
modified~(\ex{1}).

\myeenumsentence{
\item \shortex{3}{k{\i}rm{\i}z{\i} & kanatl{\i} & b\"{o}cek}
		 {red & wing-{\em Mun} & insect}
		 {`the insect with red wings'}
\item \shortex{3}{\"{u}\c{c} & tekerlekli & bisiklet}
		 {three & wheel-{\em Mun} & bicycle}
		 {`The bicycle with three wheels'}
}

Another inflection that a noun group may have is the {\em relativizer}
({\tt -ki} suffix). This suffix is attached to some temporal adverbs and
nouns with locative case. A relativized noun becomes a specifier for
another noun group~(\ex{1}).

\myenumsentence{
\shortex{2}{bah\c{c}edeki & \c{c}i\c{c}ekler}
	  {garden-{\em Loc-Rlvz} & flower-{\em Plu}}
	  {`The flowers in the garden'}
}

{\tt -ki} suffix following a genitive noun group behaves as a pronoun
meaning `one that belongs to' and different from the relativizer {\tt -ki}.

\myenumsentence{
\shortex{3}{Ay\c{s}e'ninkiler & yar{\i}n & gelecek}
	   {Ay\c{s}e-{\em Gen-Pro-Plu} & tomorrow & come-{\em Fut-3Sg}}
	   {`Ones that Ay\c{s}e owns will come tomorrow.'}
}

A noun group consists of an optional group of specifier and modifiers and a
head noun. Head noun can be a common noun, a pronoun or a proper noun.
Order and grammatical combinations of specifiers and modifiers change
according to the type of specifiers and modifiers. Order and valid
combinations of specifiers and modifiers are pragmatically controlled. Some
specifiers/modifiers put some restrictions on the 
specifier/modifier types that can further specify/modify the noun.

\newcommand{\exfont}{\normalsize}

General structure of a noun group can be viewed as a sequence of segments, 
head noun being the last one. These segments are listed in Table 
\ref{SEGMENTS}.\\

\begin{table}
\begin{tabular}{lll} 
Segments & Alternatives & Examples \\ \hline
Specifier & Quantifier & her, baz{\i}, biraz, kimi, herbir, bir\c{c}ok \\
	& Article & bir \\
	& Demonstrative Adjective & bu, \c{s}u, o, di\u{g}er, ilk, sonuncu \\
	& Genitive noun & bah\c{c}enin\\
	& Classifier noun & {\em mutfak} dolab{\i} \\
Modifier & Quantitative Adjective & d\"{o}rt, yar{\i}m, iki\c{s}er, \"{u}\c{c}l\"{u} \\
	& Qualitative Adjective & g\"{u}zel, zor \\
	& Relativized noun & evdeki, ak\c{s}amki \\
	& Relative clause & postadan \c{c}{\i}kan, yolda g\"{o}rd\"{u}\u{g}\"{u}m \\
	& Unit noun & bardak, salk{\i}m, tane \\
Head	& Common noun & ev, kitap \\
	& Proper noun & Deniz, Ankara \\
	& Pronoun & ben, sen, onlar \\ \hline
\end{tabular}
\caption{Segments of a noun group.}
\label{SEGMENTS}
\end{table}

Specifier and modifier segments are optional:

\myeenumsentence{
{\exfont
\item[a.]\shortex{2}{bah\c{c}enin & kap{\i}s{\i}}
		    {garden-{\em Gen} & gate-{\em 3SP}}
		    {`the gate of the garden'}
\item[b.]\shortex{2}{\c{s}u & k{\i}z}
		    {that & girl}
		    {`that girl'}
\item[c.] Ankara
}
}

The order of the specifier and modifier segments may vary.

\myeenumsentence{
{\exfont
\item[]
\begin{tabular}[t]{llllp{0.5cm}llll}
a. & Her & k{\i}rm{\i}z{\i} & \c{c}i\c{c}ek & & b. & k{\i}rm{\i}z{\i} & her & \c{c}i\c{c}ek \\
   & every & red & flower  & &    & red & every & flower \\
   & \multicolumn{3}{l}{`every red flower'} & &    &
			\multicolumn{3}{l}{`every flower that is red'}\\
c. & g\"{u}zel & bir & k{\i}z     & & d. & bir & g\"{u}zel & k{\i}z  \\
   & beautiful & a & girl  & &    & a   & beautiful & girl \\
   & \multicolumn{3}{l}{` a beautiful girl'} & &    &
                        \multicolumn{3}{l}{`one beautiful girl'}
\end{tabular}
}
}

Each segment of the noun group are elaborated below.

\subsection{Specifier segment}
Specifiers pick out noun(s) out of a set of possible nouns. In Turkish,
specifier segment position is filled by a specifier that can be a {\em
quantifier} (\ex{1}a), an {\em article} (\ex{1}b), a {\em demonstrative
adjective} (\ex{1}c), a {\em genitive noun} (\ex{1}d) or a {\em a classifier
noun} (\ex{1}e).

\myeenumsentence{
{\exfont
\item[a.]\shortex{5}{Yazd{\i}klar{\i}m{\i}z & {\em baz{\i}} & insanlar{\i} & rahats{\i}z &
							     edecek.}
           {write-{\em Part}-{\em 1PP} & some & people-{\em Acc} & 
						disturbed & make-{\em Fut}}
                  {`What we have written will disturb some people.'}
\item[b.]\shortex{4}{Yolda & {\em bir} & kalem & buldum.}
		{road-{\em Loc} & a & pencil & find-{\em Past-1Sg}}
		{`I found a pencil on the street.'}
\item[c.]\shortex{3}{{\em \.{I}lk} & s{\i}nav{\i}m{\i} & ge\c{c}tim.}
		    { first & exam-{\em 1SP} & pass-{\em Past-1Sg}}
                    {`I passed my first exam.'}
\item[d.]\shortex{3}{{\em yazar{\i}n} & her & kitab{\i}}
                    { author-{\em Gen} & every & book-{\em 3SP}}
                    {`every book of the author'}
\item[e.]\shortex{5}{Onur'un & buldu\u{g}u & iki & {\em caz }& pla\u{g}{\i}}
		{Onur-{\em Gen} & find-{\em Part}-{\em 3Sg} & two & jazz & 
					record-{\em 3SP}}
		{`two jazz records that Onur found'}
}
}

The valid combinations and and order of specifiers are pragmatically
controlled. A noun group may have only one quantifier (\ex{1}a--b). Also 
quantifiers cannot be used with demonstrative adjectives and article
(\ex{1}c,d).

\myeenumsentence{
{\exfont
\item[a.]\shortex{3}{*her & \c{c}o\u{g}u & kitap}
			{every & most & book}
			{}
\item[b.]\shortex{3}{*bir\c{c}ok & kimi & \"{o}\u{g}renciler}
			{many & some & student-{\em Plu}}
			{}
\item[c.]\begin{tabular}[t]{l@{}lcc}
	*&\uline{her} & \uline{bu} & \uline{kitap} \\
	 &{\tiny Quant.} & {\tiny Dem. Adj.} & - \\
	 &every & this & book
\end{tabular}
\item[d.]\begin{tabular}[t]{l@{ }lcc}
	*&\uline{kimi}&\uline{bir}&\uline{insan}\\
	 &{\tiny Quant.} & {\tiny Art.} & - \\
	 &some & a  & person       
\end{tabular}
}
}

The use of the article with demonstrative adjectives and quantifiers depends on
some selectional restrictions.

\myeenumsentence{
{\exfont
\item[a.]\begin{tabular}[t]{l@{}lll}
	*&ilk & bir & kitap\\
         &first & a & book
\end{tabular}
\item[b.]\begin{tabular}[t]{lll}
	bir & ilk & kitap\\
        a & first & book
\end{tabular}
\item[c.]\shortex{3}{bir & \"{u}\c{c}\"{u}nc\"{u} & kitap}
			{a & third & book }
			{`yet a third book'}
\item[d.]\begin{tabular}[t]{l@{}lll}
	*&sonuncu & bir & kitap\\
         &last & a & book
\end{tabular}
\item[e.]\shortex{3}{bir & \c{s}u & kitap}
			{ a & that & book}
			{`only that book'}
\item[f.]\shortex{3}{\c{s}u & bir & kitap}
			{ that & one & book}
			{`that single book'}
\item[g.]\shortex{3}{di\u{g}er & bir & kitap}
			{other & a & book}
			{`another book'}
}
}

There are some points to be underlined here, about the different meanings of 
``bir'' and about some exceptions:\\
The sequence depicted in (\ex{0}b) has a limited usage referring to the 
\uline{first} book of an author. In (\ex{0}c), ``bir'' is an adverb 
meaning ``yet'' or ``another''. In (\ex{0}e), ``bir'' is an adverb with a 
meaning ``only''. In (\ex{0}f), ``bir''  is not an article but a cardinal 
number (a modifier). In (\ex{0}g), ``di\u{g}er'' acts as an adverb.\\

Concerning the demonstrative adjectives, there are two subgroups:
\begin{itemize}
\item[i)] bu, \c{s}u, o
\item[ii)] ilk, sonuncu, ordinal numbers, di\u{g}er
\end{itemize}

Only one element from each group can be used within a noun group. 
The elements of the first group can sometimes be used in front of the elements
of the second group for emphasizing the demonstration. \\

\myeenumsentence{
{\exfont
\item[a.]\shortex{5}{Bu & ikinci & kitab{\i} & pek & 										be\u{g}enmedim}
			 {this & second & book-{\em Acc} & much 							& like-{\em Neg}-{\em Past-1Sg}} 
			 {`I didn't like the second book much'}
\item[b.]\shortex{4}{\c{S}u & di\u{g}er & valiz & benimki}
			  {that & other & suitcase & mine}
			  {`The other suitcase is mine'}
\item[c.]\shortex{3}{*di\u{g}er & sonuncu & k{\i}z}
				 {other & last & girl}{}
}
}

Nouns or noun groups with genitive marking also function 
as specifiers within a noun group. Genitive nouns can be used in combination 
with other specifiers (\ex{1}a--b). The main restriction is that all
specifiers and modifiers modifying the possessive marked noun should
follow the genitive noun.  Otherwise they specify/modify the genitive
noun (\ex{1}c):\\

\myeenumsentence{
{\exfont
\item[a.]\shortex{3}{yazar{\i}n & bir & kitab{\i}}
			 {author-{\em Gen} & a & book-{\em 3SP}}
			 {`a book of the author'}
\item[b.]\shortex{3}{kitab{\i}n & bu & sayfas{\i}}
			  {book-{\em Gen} & this & page-{\em 3SP}}
			  {`this page of the book'}
\item[c.]\shortex{3}{ilk & seminerin & konu\c{s}mac{\i}s{\i}}
			  {first & seminar-{\em Gen} & speaker-{\em 3SP}}
			  {`the speaker of the first seminar'}
}
}

Genitive nouns can rather be interpreted as complements of possessive marked
nouns since possessive nouns require a genitive noun which is subject of
the owner relation in the possessive group.

Classifier nouns resemble genitive nouns in that they require a 
possessive-marked noun group modified by the classifier noun. However, 
classifier nouns take no genitive suffix.\footnote{These noun groups are 
called {\bf izafet} by Lewis \cite{Lewis67}} The difference between a 
genitive noun and a classifier noun is that 
the former provides a definite reading where the latter provides an 
indefinite  or nonreferential one.

\myeenumsentence{
{\exfont
\item[a.]\shortex{2}{duvar & boyas{\i}}
			  {wall & paint-{\em 3SP}}
			  {`wall paint'}
\item[b.]\shortex{2}{duvar{\i}n & boyas{\i}}
			  {wall-{\em Gen} & paint-{\em 3SP}}
			  {`the paint of the wall'}
}
}

Classifier noun groups can act as specifiers of other classifier nouns:

\myeenumsentence{
{\exfont
\item[]\shortex{4}{kredi & kart{\i} & faiz & y\"{u}zdesi}
			  {credit & card-{\em 3SP} & interest & 
					percentage-{\em 3SP}}
			  {`credit card interest rate'}
}
}

A classifier noun is the immediate predecessor of the head noun. Hence, other 
specifiers and modifiers precede it.

\myeenumsentence{
\item[a.]\shortex{3}{her & \c{c}ocuk & arabas{\i}}
			  {every & child & car-{\em 3SP}}
			  {`every stroller'}
\item[b.]\shortex{3}{o & dere & yata\u{g}{\i}}
			  {that & stream & bed-{\em 3SP}}
			  {`that river bed'}
\item[c.]*\c{c}ocuk her arabas{\i}
\item[d.]*dere o yata\u{g}{\i}
\item[e.]*ev bir kap{\i}s{\i}
\item[f.]*duvar evin boyas{\i}
\item[g.]\shortex{3}{evin & duvar & boyas{\i}}
			  {home & wall & paint-{\em 3SP}}
			  {`wall paint of the house'}
}

\subsection{Modifier segment}

Modifiers provide information about the properties of the entity or its
relations with other entities. A modifier is either an adjective group, or 
a noun group. More than one modifier may exist within a noun
group.

\myeenumsentence{
{\exfont
\item[]\shortex{3}{g\"{u}zel & mavi & ete\u{g}in}
			{beautiful & blue & skirt-{\em 2SP}}
			{`your beautiful blue skirt'}
}
}

As a general rule, ``whatever precedes modifies'' in Turkish. Hence, if a 
modifier itself is a noun group or a clause containing a noun, any 
preceding modifier modifies the first of the succeeding nouns.\\
For example, in the phrase below, the modifier ``ya\c{s}l{\i}'' modifies ``adam''
rather than the head noun ``kad{\i}n''.

\myeenumsentence{
{\exfont
\item[]\shortex{4}{ya\c{s}l{\i} & adam{\i}n & konu\c{s}tu\u{g}u & kad{\i}n}
           {old & man-{\em Gen} & talk-{\em Part}-{\em 3SP} & woman}
           {`the woman to which the old man talked/talks'}
}
}

Certain restrictions apply to the combinations of modifiers.
When a noun is modified by both a qualitative and a quantitative adjective,
order of the adjectives may vary but the quantitative adjective usually 
precedes the qualitative one.

\myeenumsentence{
{\exfont
\item[a.]\shortex{3}{\"{u}\c{c} & k{\i}rm{\i}z{\i} & kalem}
			{three & red & pencil}
			{`three red pencils'}
\item[b.]\shortex{3}{hassas & ikili & ili\c{s}kiler}
			{sensitive & dual & relationship-{\em Plu}}
			{`sensitive dual relationships'}
\item[c.]\shortex{3}{iki\c{s}er & kal{\i}n & battaniye}
			{by-two & thick & blanket}
			{`two thick blankets for each'}
\item[d.]\shortex{3}{rahat & \"{u}\c{c}l\"{u} & kanepe}
			{comfortable & triple & sofa}
			{`a comfortable triple sofa'}	
\item[e.]\shortex{3}{yar{\i}m & \c{c}\"{u}r\"{u}k & elma}
			{half & rotten & apple}
			{`a half rotten apple'}
\item[f.]\shortex{3}{\c{c}\"{u}r\"{u}k & yar{\i}m & elma}
		{rotten & half & apple}
		{`a half rotten apple'}
}
}

When used as modifiers, unit nouns are preceded by a cardinal number (\ex{1}a),
a fractional number (\ex{1}b), or a distributive adjective (\ex{1}c):

\myeenumsentence{
{\exfont
\item[a.]\shortex{3}{iki & bardak & s\"{u}t}
                        {two & glass & milk}
                        {`two glasses of milk'}
\item[b.]\shortex{3}{yar{\i}m & somun & ekmek}
			{half & loaf & bread}
			{`half loaf of bread'}
\item[c.]\shortex{3}{birer & dilim & pasta}
			{by-one & slice & cake}
			{`a slice of cake (for each)'}
}
}
When the unit noun denotes a container, the word {\em dolusu} 
(``full''-{\em 3SP})
may optionally be inserted between the unit noun and the head.

\myeenumsentence{
{\exfont
\item[]\shortex{4}{\"{u}\c{c} & ka\c{s}{\i}k & dolusu & \c{s}eker}
                  {three & spoon & full-{\em 3SP} & sugar}
                  {`three spoonful of sugar'}
}
}

The other group of modifier is the relativized nouns which are inflected by
relativizer suffix {\tt -ki}.
If the head noun is modified by a relativized noun, all other modifiers and
specifiers of the head come after the relativized noun (\ex{1}a). Otherwise,
any modifier preceding a relativized noun modifies the relativized noun rather
than the head (\ex{1}b). 

\myeenumsentence{
{\exfont
\item[a.]\shortex{4}{\c{c}antamdaki & \"{u}\c{c} & k\"{u}\c{c}\"{u}k & anahtar}
			{handbag-{\em 1SP}-{\em Loc-Rlvz} & three & small & key}
			{`three small keys in my handbag'}
\item[b.]\shortex{4}{k\"{u}\c{c}\"{u}k & \c{c}antamdaki & \"{u}\c{c} & anahtar}
			{small & handbag-{\em 1SP}-{\em Loc-Rlvz} & three &  
						key}
			{`three keys in my small handbag'}
}
}

Noun groups can also be modified by relative clauses. In Turkish, the noun on 
which the relativization is performed is placed at the final position of the 
relative clause.

\myeenumsentence{
{\exfont
\item[a.]\shortex{3}{A\u{g}abeyim & Ankara'da & \c{c}al{\i}\c{s}{\i}yor.}
		{brother-{\em 1SP} & Ankara-{\em Loc} & work-{\em Prog-3Sg}}
		{`My elder-brother works in Ankara.'}
\item[b.]\shortex{3}{Ankara'da & \c{c}al{\i}\c{s}an & a\u{g}abeyim }
		{Ankara-{\em Loc} & work-{\em Rel} & brother-{\em 1SP} }
		{`My elder-brother who works in Ankara'}
}
}

As seen above, the main verb of the relative clause is used in participle 
form. The example depicts the suffix {\tt -en} (phonological realization of 
{\tt -(y)An} after morphophonemic processes) which is used in producing 
subject participle (in present). Other subject suffixes are given below:\\

\begin{tabular}{cc}
{\bf Suffix} & {\bf Tense} \\ \hline
{\tt -mH\c{s}} (olan) & past \\
{\tt -(y)AcAk} (olan) & future\\
{\tt -Hyor} (olan) & progressive
\end{tabular}\\

\vspace*{0.5cm}

The word {\em olan} (``being'') can optionally be used with past, future or
progressive participles, but not with present participle.

\myeenumsentence{
{\exfont
\item[a.]\shortex{4}{Ankara'da & \c{c}al{\i}\c{s}m{\i}\c{s} & olan & a\u{g}abeyim }
		{Ankara-{\em Loc} & work-{\em Part} & be-{\em Rel} & 
				elder-brother-{\em 1SP}}
		{`my elder brother who have worked in Ankara'}
\item[b.]\shortex{4}{Ankara'da & \c{c}al{\i}\c{s}acak & olan & a\u{g}abeyim }
		{Ankara-{\em Loc} & work-{\em Part} & be-{\em Rel} & 
				elder-brother-{\em 1SP} }
		{`my elder brother who will work in Ankara'}
\item[c.]* Ankara'da \c{c}al{\i}\c{s}an olan a\u{g}abeyim
}
}

{\em Olan} can also be used in forming participle form of the copula.

\myeenumsentence{
{\exfont
\item[a.]\shortex{3}{Arkada\c{s}{\i}m{\i}n & annesi & hasta.}
		{friend-{\em 1SP}-{\em Gen} & mother-{\em 3SP} & ill}
		{`My friend's mother is ill.'}
\item[]\shortex{4}{annesi & hasta & olan & arkada\c{s}{\i}m}
		{mother-{\em 3SP} & ill & be-{\em Rel} & friend-{\em 1SP}}
		{`my friend whose mother is ill'}
\item[b.]\shortex{3}{Evin & pencereleri & geni\c{s}.}
		{house-{\em Gen} & window-{\em Plu-Acc} & wide}
		{`Windows of the house are wide.'}
\item[]\shortex{5}{pencereleri & geni\c{s} & olan & bir & ev}
		{window-{\em Plu}-{\em 3SP} & large & be-{\em Rel} & a & 									house}
		{`a house which has large windows'}
}
}

Apart from the subject participle form, the verb of a relative clause may take 
{\em complement participle} form, which is obtained by attachment of either 
{\tt -DHk} or {\tt -yAcAk} suffixes. {\tt -DHk} suffix, as itself, produces 
adjectives from verbs, although it is not productive:

\myenumsentence{
{\exfont
\item[]\shortex{3}{bilmek & $\rightarrow$ & bildik}
		{`to know' & & `known'}
		{}
\item[]\shortex{3}{umulmamak & $\rightarrow$ & umulmad{\i}k}
		{`to be not expected' & & `unexpected'}
		{}
}
}

When used in complement participles, {\tt -DHk} is always followed by a 
possessive (marks the agreement in this case) and participle suffix
group becomes {\tt DH\u{g}}-{\em Agr}.  The tense of this participle can
be past or present, as examples (\ex{1}) and (\ex{2}) depicts,
respectively. Actual tense is usually determiner from the discourse.

\myeenumsentence{
{\exfont
\item[]\shortex{4}{Kitab{\i} & k{\i}za & geri & verdim.}
		{book-{\em Acc} & girl-{\em Dat} & back & give-{\em Past-1Sg}}
		{`I gave back the book to the girl.'}
\item[a.]\shortex{4}{k{\i}za & geri & verdi\u{g}im & kitap}
		{girl-{\em Dat} & back & give-{\em Rel-1Sg} & book-{\em Acc}}
		{`The book that I gave back to the girl.'}
\item[b.]\shortex{4}{kitab{\i} & geri & verdi\u{g}im & k{\i}z}
		{book-{\em Acc} & back & give-{\em Rel-1Sg} & girl}
		{`The girl to whom I gave back the book.'}
} 
}

If there is an overt subject noun in the clause, it is marked with the genitive 
suffix:

\myeenumsentence{
{\exfont
\item[]\shortex{4}{\"{O}\u{g}renci & s{\i}n{\i}fta & \c{s}ark{\i} & s\"{o}yl\"{u}yor.}
		{student & class-{\em Loc} & song & sing-{\em Prog-3Sg}}
		{`The student is singing a song in the classroom.'}
\item[a.]\shortex{4}{\"{o}\u{g}rencinin & s{\i}n{\i}fta & s\"{o}yledi\u{g}i & \c{s}ark{\i} }
		{student-{\em Gen} & class-{\em Loc} & sing-{\em Rel}-{\em 3Sg} & 
					song}
		{`the song that the student is singing in the classroom'}
\item[b.]\shortex{4}{\"{o}\u{g}rencinin & \c{s}ark{\i}y{\i}  & s\"{o}yledi\u{g}i & s{\i}n{\i}f}
		{student-{\em Gen} & song-{\em Acc} & sing-{\em Rel}-{\em 3Sg} & 
					class}
		{`the classroom in which the student is singing the song'}
}
}

Complement participles in future tense are formed by attaching {\tt -(y)AcAk} 
suffix to verb stem. Just like {\tt -DHk} suffix, {\tt -(y)AcAk} combines 
with a possesive suffix to produce {\tt -(y)AcA\u{g}}-{\em Agr} as the future 
complement participle.\\

\myeenumsentence{
{\exfont
\item[a.]\shortex{3}{\"{o}\u{g}rencinin & s\"{o}yleyece\u{g}i & \c{s}ark{\i} }
		{student-{\em Gen} & sing-{\em Rel}-{\em 3Sg} & song}
		{`the song that the student will sing'}
\item[b.]\shortex{4}{kitab{\i} & geri & verece\u{g}im & k{\i}z}
		{book-{\em Acc} & back & give-{\em Rel-1Sg} & girl}
		{`the girl to whom I will give back the book'}
}
}

Relative clauses can be embedded as adnominals:

\myeenumsentence{
{\exfont
\item[]\shortex{7}{k\"{o}yde & ya\c{s}ayan & k{\i}z{\i}n & yeti\c{s}tirdi\u{g}i & ine\u{g}in &
				\"{o}ld\"{u}\u{g}\"{u} & yer}
		{village-{\em Loc} & live-{\em Rel} & girl-{\em Gen}
			 & breed-{\em Rel-3Sg} & 
			cow-{\em Gen} & die-{\em Rel-3Sg} & 
					place}
		{`the place at which the cow that was breeded by the girl
 			who lives in the village died '}
}
}

\subsection{The head}

The last segment of the noun group is the head, and this position is filled 
either by a common noun (\ex{1}a), a proper noun (\ex{1}b) or a pronoun
(\ex{1}c).

\myeenumsentence{
\item[a.]\shortex{3}{k\"{u}\c{c}\"{u}k & bir & elma}
			{small & a & apple}
			{`a small apple'}
\item[b.]\shortex{2}{g\"{u}zel & Ay\c{s}e}
		{beautiful & Ay\c{s}e}
		{`beautiful Ay\c{s}e'}
\item[c.]\shortex{4}{unutkanl{\i}\u{g}{\i}yla & bilinen & sen}
			{forgetful-{\em 3Sg-Ins} &
					know-{\em Pass-Rel} & you}
			{`you who are known as forgetful'}
}

\subsubsection{Pronouns}

When the head is a pronoun, no determiner or modifier segments are allowed:

\myeenumsentence{
{\exfont
\item[]*\shortex{2}{baz{\i} & sen}
           {some & you}
           { }
\item[]*\shortex{2}{sar{\i}\c{s}{\i}n & ben}
		{blond & I}
		{}
}
}

\subsubsection{Proper Nouns}
	
When it is used as the head, a proper noun imposes certain restrictions on the 
selection of the preceding segments. For example, particular determiners can be used in front of a proper noun, while others are not applicable.\\

\myeenumsentence{
{\exfont
\item[a.]\shortex{4}{{\em Bu} & {\em \.{I}stanbul} & nas{\i}l & d\"{u}zelir?}
                    {this & \.{I}stanbul & how & get-better-{\em Pres-3Sg}}
                    {`How could this \.{I}stanbul get better?'}
\item[b.]\shortex{4}{Nerede & kald{\i} & {\em \c{s}u} & {\em Hasan?}}
                    {where & left & that & Hasan}
                    {`Where on the earth is Hasan?'}
\item[c.]\shortex{4}{Trakya'da & {\em birka\c{c}} & {\em Ye\c{s}ilk\"{o}y'e} & rastlad{\i}m }
		{ Thrace'-{\em Loc} & several & Ye\c{s}ilk\"{o}y'-{\em Dat} & 
				come-across-{\em Past-1Sg} }
		{ `I came across more than one Ye\c{s}ilk\"{o}y in Thrace'}
\item[d.]\shortex{4}{Ailemizdeki & {\em di\u{g}er/ikinci} & {\em Mehmet} & 
								dedemdir.}
		    {family-{\em 1PP-Loc-Rlvz} & other/second & Mehmet & 
							grandfather-{\em
							1SP-Cop}}
		    {`The other/second Mehmet in our family is my
							grandfather'}.
}
}

\section{Postposition group}
By postposition group, we mean a group of elements whose head is a 
proposition. Postposition group consists of a head and an optional 
complement noun group. The former always occupies the final position.\\

\subsection{Postpositions}

Postpositions form a closed class of words. They can be viewed in subgroups,
with respect to the case of the complement they subcategorize for.
(\cite{Lewis67}, pp. 85-89) Various types of postpositions exist which
subcategorize for: infinitives or nouns with nominative case (\ex{1}a,b),
nouns with accusative case (\ex{1}c), dative case (\ex{1}d) and ablative
case (\ex{1}e).

\myeenumsentence{
\item[a.]\shortex{2}{gelmek & \"{u}zere}
                { come-{\em Inf}& for }
		{`for the purpose of coming'}
\item[b.]\shortex{2}{sokak & boyunca}
                { street& along}
                {`along the street'}
\item[c.]\shortex{2}{S{\i}nav{\i} & m\"{u}teakiben}
		{exam-{\em Acc} & following}
		{`after the exam'}
\item[d.]\shortex{2}{\c{s}imdiye & dek}
		{now-{\em Dat} & until}
		{`until now'}
\item[e.]\shortex{3}{D\"{u}nden & beri}
		{yesterday-{\em Abl} & since}
		{`since yesterday'}
}

\subsection{Postposition Attachment}

Attachment of a sequence of postpositions is determined without ambiguity
by morphosyntactic cues (e.g., relative suffixes and case marks) and 
positional cues (head-final structure). However, if a sentence involves
relative clauses and postpositions, ambiguities may arise (\ex{1}a).\footnote{
In writing, it this may be disambiguated by seperating the sentential
complement with a coma (before {\bf yazd{\i}\u{g}{\i}m{\i}z} in \ex{1}a).}
In ``I read the newspaper on the couch'', if {\em on the couch} 
were an adnominal, it would be relativized in Turkish~(cf.,\ex{1}b-c). 
Chained postposition groups are not ambiguous because the predecessor 
modifies the successor.

\myeenumsentence{
\label{ppattach}
\item[a.]\shortex{6}{Bu & bilgilere & g\"{o}re & yazd{\i}\u{g}{\i}m{\i}z & rapor
  				& de\u{g}i\c{s}meyecek.}
                {this & data-{\em Plu-Dat} & according & 
					write-{\em Rel-1Pl} & report
                                & change-{\em Neg-Fut-3Sg}}
                      {\shortstack{`The report that we wrote according to these
			data will not change.' \\
			`According to these data, the report that we wrote
						will not change'}}
\item[b.]\shortex{3}{Kanepedeki & gazeteyi & okudum.}
                    {couch-{\em Loc-Rlvz} & newspaper-{\em Acc} & read-{\em
		    Past-1Sg}}
                    {`I read [ the newspaper on the couch].'}
\item[c.]\shortex{3}{Kanepede & gazeteyi & okudum.}
                    {couch-{\em Loc} & newspaper-{\em Acc} & read-{\em
		    Past-1Sg}}
                    {`I read the newspaper [ on the couch].'}
}

\section{Adjective group}
Adjective group is a sequence of words last of which is an adjective.
Adjective groups are typically formed by comparative and superlative
adjectives.\\

\subsection{Comparative adjectives}
The head of a comparative adjective group is a qualitative adjective. Three
comparatives can precede the head: ``daha'', ``az'' and ``\c{c}ok'' meaning
``more'', ``less'' and ``very'', respectively.\\

\myeenumsentence{
{\exfont
\item[a.]\shortex{6}{Elvan & daha & b\"{u}y\"{u}k & bir & eve & ta\c{s}{\i}nd{\i}.}
		{Elvan & more & big & a & house-{\em Dat} & 
					move-in-{\em Past-3Sg}}
                {`Elvan moved in to a bigger house.'}
\item[b.]\shortex{4}{Az & \c{s}ekerli & kahve & i\c{c}erdi.}
		{less & sweet & coffee & drink-{\em Pres-Asp}}
		{`(S)he used to drink coffee with a little sugar.'}
\item[c.]\shortex{4}{\c{C}ok & h{\i}zl{\i} & arabalardan & ho\c{s}lanm{\i}yorum.}
		{very & fast & car-{\em Plu}-{\em Abl} & 
					like-{\em Neg-Prog-1Sg}}
		{`I don't like very fast cars.'}
\item[d.]\shortex{7}{Annem & benden & \c{c}ok & daha & iyi & yemek & yapar.}
		{mother-{\em 1SP} & I-{\em Abl} & very & more & good & dish & 
						make-{\em Pres-3Sg}}
		{`My mother cooks much better than I do.'}
}
}

\subsection{Superlative adjectives}
The head adjective is qualitative for this group, too. Superlative form
is obtained by preceding the head with ``en'' (``most'').\\

\myeenumsentence{
{\exfont
\item[]\shortex{5}{S{\i}n{\i}f{\i}n & en & \c{c}al{\i}\c{s}kan & \"{o}\u{g}rencisi & Ali'ydi.}
		{class-{\em Gen} & most & hardworking & student-{\em 3SP} & 
					Ali-{\em Cop}}
		{`Ali was the most hardworking student of the class.'}
}
}

\section{Adverb group}
An adverb group is a segment which has an adverb as its head.
Modifiers of an adverbial head may be adverb or adjective groups,
including the comparative {\em daha} and the superlative {\em en}.
Adverbial heads may be classified as manner ({\em alelacele}), temporal
({\em sonra, \"{o}nce}), position ({\em a\c{s}a\u{g}{\i}, beri, ileri}),
repetition ({\em gene, yeniden, tekrar}), sentential ({\em besbelli, 
asla, ku\c{s}kusuz}), frequency ({\em seyrek, s{\i}k}), possibility
({\em herhalde, belki}), definiteness ({\em katiyen, muhakkak}), and
question ({\em nas{\i}l, hani}) adverbs.
Basic types of adverb groups are described below.\\

\subsection{Reduplications}
Nouns and adjectives can be reduplicated to form an adverb group.\\

\myeenumsentence{
{\exfont
\item[a.]\shortex{4}{Yeme\u{g}imizi & \c{c}abuk &\c{c}abuk & yedik.}
		{meal-{\em 1PP-Acc} & quick &quick & eat-{\em Past-1Pl}}
		{`We ate our meal quickly.'}
\item[b.]\shortex{4}{Ak\c{s}am & ak\c{s}am & can{\i}m{\i}z{\i} & s{\i}kt{\i}.}
		{evening & evening & soul-{\em 1PP-Acc} & 
					bother-{\em Past-3Sg}}
		{`It bothered us at this time of the evening.'}
\item[c.]\shortex{7}{Ge\c{c}en & yaz & bu & sahilleri & koy & koy & dola\c{s}t{\i}k.}
		{last & summer & this & shore-{\em Plu}-{\em Acc} & cove & 
					cove & go-around-{\em Past-1Pl}}
		{`We visited each and every cove of this shore last summer.'}
}
}

Some of the reduplicated adverbs are onomatopoeic words:\\

\myeenumsentence{
{\exfont
\item[]\shortex{6}{\c{S}{\i}r{\i}l & \c{s}{\i}r{\i}l & akan & derenin & sesini & dinledim.}
		{\multicolumn{2}{l}{`splashing'} & flow-{\em Part} &
                                       stream-{\em Gen} & sound-{\em 3SP-Acc}}
		{`I listened to the sound of the stream that flows gently.'}
}
}

Distributive adjectives, when used as adverbs, are reduplicated:\\

\myeenumsentence{
{\exfont
\item[]\shortex{4}{Merdivenleri & \"{u}\c{c}er & \"{u}\c{c}er & \c{c}{\i}kt{\i}k.}
		{stairs-{\em Plu}-{\em Acc} & three-{\em Dist} & 
				three-{\em Dist}& go-up-{\em Past-1Pl}}
		{`We went upstairs three steps by three steps.'}
}
}

Adverbs or adjectives can be intensified by phonological reduplication
to produce adverbs as well:\\
\hspace*{0.5cm}

\begin{tabular}{llp{5mm}ll}
{\bf \c{c}abuk} & quick & & {\bf \c{c}ar\c{c}abuk} & very quickly\\
{\bf h{\i}zl{\i}} & fast & & {\bf h{\i}ph{\i}zl{\i}} & very fast\\
\end{tabular}

\subsection{Case-marked place adverbs}
Adverbs of place act as the head of an adverb group either by themselves
or by taking a case suffix.\\
\hspace*{0.5cm}

\begin{tabular}{llp{0.5cm}ll}
{\bf i\c{c}eri} & inside & & {\bf d{\i}\c{s}ar{\i}} & outside \\
{\bf yukar{\i}} & up & & {\bf a\c{s}a\u{g}{\i}} & down \\
{\bf ileri} & forward & & {\bf geri} & backward \\
{\bf \"{o}te} & yonder & & {\bf beri} & hitter \\
{\bf \"{o}n} & front & & {\bf arka} & behind \\
{\bf kar\c{s}{\i}} & opposite & & &
\end{tabular}

\myeenumsentence{
{\exfont
\item[a.]\shortex{3}{Evden & d{\i}\c{s}ar{\i} & \c{c}{\i}kmad{\i}m.}
		{house-{\em Abl} & outside & go-out-{\em Neg}-{\em Past-1Sg}}
		{`I didn't go out from the house.'}
\item[b.]\shortex{3}{Yolun & ilerisi & g\"{o}r\"{u}lm\"{u}yor.}
		{road-{\em Gen} & forward-POSS & see-{\em Pass}-{\em Neg}}
		{`The forward part of the road is not visible.'}
\item[c.]\shortex{4}{Nehirden & \"{o}teye & nas{\i}l & ge\c{c}ilir?}
		{river-{\em Abl} & yonder-{\em Dat} & how &  
				pass-{\em Pass-Pres-3Sg}}
		{`How can one go beyond the river?'}
}
}

\subsection{Temporal adverb groups}

``sonra'' (``after'') and ``\"{o}nce'' (``before'') succeed noun groups
denoting a time period or a point in time, and form adverb groups.\\

\myeenumsentence{
{\exfont
\item[a.]\shortex{5}{D\"{o}rt & g\"{u}n & sonra & yola & \c{c}{\i}kaca\u{g}{\i}z.}
		{four & day & after & road & go-out-{\em Fut-1Pl}}
		{`We'll set out on a journey in four days.'}
\item[b.]\shortex{54}{Umar{\i}m & Per\c{s}embeden & \"{o}nce & burada & olmaz.}
		{hope-{\em Pres-1Sg} & Thursday-{\em Abl} & before & 
					here & be-{\em Neg}}
		{`I hope he/she won't be here before Thursday.'}
}
}

Another type of adverb group denoting time is the one that uses special
temporal nouns in head position. These temporal nouns are some time units
({\bf g\"{u}n}:``day'', {\bf hafta}:``week'', {\bf ay}:``month'',
{\bf mevsim}:``season'', {\bf y{\i}l}:``year'', 
{\bf y\"{u}zy{\i}l}:``century'', {\bf d\"{o}nem}: ``semester, age'', 
{\bf \c{c}a\u{g}}:``era, epoch''), days of week, months and year.
In such adverb goups, however, the set of words that may modify the head is
rather limited: {\bf \"{o}nceki} (``previous'', ``before''), {\bf ertesi} 
(``following'', ``after''), {\bf ge\c{c}en} (``last''), {\bf gelecek} (``next''), 
{\bf bu} (``this''), {\bf o} (``that'').\\

\myeenumsentence{
{\exfont
\item[a.]\shortex{6}{Ertesi & g\"{u}n & eski & bir & arkada\c{s}{\i}ma & rastlad{\i}m.}
		{following & day & old & a & friend-{\em 1SP-Dat} &
                                     come-across-{\em Past-1Sg}}
		{`The following day, I came across with an old friend of mine.'}
\item[b.]\shortex{4}{Gelecek & yaz & Paris'e & gidece\u{g}im.}
		{next & summer & Paris-{\em Dat} & go-{\em Fut-1Sg}}
		{`I will go to Paris next summer.'}
}
}

\subsection{Verb groups with adverbial use}
Verb stems may function as adverbs with the addition of certain suffixes. 
These suffixes are discussed below.\\

{\tt -(y)A} suffix denotes a repeated action that takes place at the same
time with the main verb. Verb groups in this gerundive form consist of
two gerunds (either of the same verb or different verbs).

\myeenumsentence{
{\exfont
\item[a.]\shortex{4}{A\u{g}ac{\i} & buda{\em ya} & buda{\em ya} & bi\c{c}imlendirdi.}
{tree-{\em Acc} & prune & prune & shape-{\em Past-3Sg}}
{`He/she shaped the tree pruning.'}
\item[b.]\shortex{4}{\c{C}ocuk & d\"{u}\c{s}{\em e} & kalk{\em a} & b\"{u}y\"{u}r.}
{child & fall & rise & grow-{\em Pres}}
{`A child grows falling and rising.'}
}
}

{\tt -(y)ArAk} suffix denotes a continuous action or a point action which
takes place either at the same time with the main verb or just before
it.

\myeenumsentence{
{\exfont
\item[a.]\shortex{2}{\"{O}p\"{u}\c{s}{\em erek} & ayr{\i}ld{\i}lar.}
{kiss-{\em Recp} & leave-{\em Past-3Pl}}
{`They kissed each other as they said goodbye.'}
\item[b.]\shortex{3}{Ko\c{s}{\em arak} & kar\c{s}{\i}ya & ge\c{c}tik.}
{run & opposite-{\em Dat} & pass-{\em Past-1Pl}}
{`We crossed the street running.'}
}
}

{\tt -(y)Hp} suffix is attached to the first of consecutive verb stem
pairs and provides a connection (e.g., temporal sequence) between these stems.

\myeenumsentence{
{\exfont
\item[a.]\shortex{4}{\c{S}emsiyemi & i\c{s}yerinde & unut{\em up} & gelmi\c{s}im.}
{umbrella-{\em 1SP-Acc} & office-{\em Loc} & forget & come-{\em Past-1Sg}}
{`I came, having forgotten my umbrella at the office'}
\item[b.]\shortex{2}{Otur{\em up} & konu\c{s}al{\i}m}
{sit-down & talk-{\em -Wish-1Pl}}
{`Let's sit down and talk.'}
}
}

{\tt -(y)HncA} suffix marks its stem as the temporal predecessor of the
main verb.

\myeenumsentence{
\item[a.]\shortex{4}{Eve & var{\em {\i}nca} & seni & arar{\i}m.}
{house-{\em Dat} & arrive & you-{\em Acc} & call-{\em Pres-1Sg}}
{`I'll call you when I arrive home.'}
\item[b.]\shortex{4}{Haberleri & dinle{\em yince} & yolculu\u{g}umu & erteledim.}
{news-{\em Acc} & listen & travel-{\em 1SP-Acc} & postpone-{\em Past-1Sg}}
{`I postponed my travel when I listened to the news.'}
}

-{\tt DHk\c{c}A} suffix is a composite one which combines participle suffix
{\tt -DHk} with {\tt \c{c}A}. This composite suffix has the meaning ``so long
as'' or ``the more''.

\myeenumsentence{
{\exfont
\item[a.]\shortex{3}{\c{C}al{\i}\c{s}ma{\em d{\i}k\c{c}a} & ba\c{s}ar{\i}l{\i} & olamazs{\i}n.}
{study-{\em Neg} & successful & be-{\em Neg}-{\em Pres-2Sg}}
{`So long as you don't study, you cannot be successful.'}
\item[b.]\shortex{4}{Ankara'ya & gel{\em dik\c{c}e} & bize & u\u{g}rar.}
{Ankara-{\em Dat} & come & we-{\em Dat} & visit-{\em Pres-3Sg}}
{`Every time he/she comes to Ankara, he/she visits us.'}
\item[c.]\shortex{3}{\.{I}p & atla{\em d{\i}k\c{c}a} & susuyorum.}
{rope & skip & be-thirsty-{\em Prog-1Sg}}
{`The more I skip, the more I get thirsty.'}
}
}

The suffix sequence {\tt -(H)r}$\cdot\cdot\cdot${\tt mAz} 
attach to the same verb stem to
produce a verb group that can be used like an adverb. This construction
has a meaning similar to ``as soon as''.

\myeenumsentence{
{\exfont
\item[a.]\shortex{4}{\.{I}bibikler & \"{o}t{\em er} & \"{o}t{\em mez} & oraday{\i}m.}
{hoopoe-{\em Plu} & sing & sing-{\em Neg} & there-{\em Loc-Cop(1Sg)}}
{`I will be there as soon as the hoopoes sing.'}
\item[b.]\shortex{5}{Otob\"{u}sten & in{\em er} & in{\em mez} & onu & g\"{o}rd\"{u}m.}
{bus-{\em Abl} & get-off & get-off-{\em Neg} & he/she/it & see-{\em Past-1Sg}}
{`I saw her/him/it as soon as I got off the bus.'}
}
}

{\tt -(y)ken} suffix is the last one that is to be discussed in this
section. It can be translated to English as ``as''. This suffix
differs from the previous ones as it attaches not to a verb stem, but
usually to third person singular inflection of the verb in aorist.
It may also attach to narrative past, present and future tense forms for
third person singular. The suffix does not harmonize with the vowels of
the verb stem.

\myeenumsentence{
\item[a.]\shortex{6}{\c{C}ay{\i}m{\i} & i\c{c}er{\em ken} & gazete & ba\c{s}l{\i}klar{\i}na &
					g\"{o}z & atar{\i}m}
{tea-{\em 1SP-Acc} & drink & newspaper & 
			headline-{\em 3PP-Dat} &
			eye & throw-{\em Pres-1Sg}}
{`I glance through newspaper headlines as I drink my tea'}
\item[b.]\shortex{5}{D\"{u}\c{s}\"{u}mde & d\"{o}v\"{u}\c{s}mektey{\em ken} & yan{\i}mda & yatan{\i} &
			tekmelemi\c{s}im.}
{dream-{\em 1Sg-Loc} & fight & 
	side-{\em 1Sg-Loc} & lie-{\em Rel-Acc} &
	kick-{\em Past-1Sg}}
{`I had kicked the one lying next to me as I was fighting in my dream.'}
\item[c.]\shortex{6}{Buraya & kadar & gelmi\c{s}{\em ken} & geri & d\"{o}nmek & olmaz}
{here-{\em Dat} & upto & come & back & turn-{\em Inf} & 
					be-{\em Neg}-{\em Pres}}
{`It's impossible to go back now that we came up to here.'}
}

\section{Verb group}
\subsection{Predicate types}
Predicates
in Turkish can be verbal~(\ref{pred}a), nominal with an attached
auxiliary suffix~(\ref{pred}b), nominal with a copula~(\ref{pred}c--d), 
or existential(\ref{pred}e--f).

\myeenumsentence{\label{pred}
\item[a.]\shortex{4}{Adam & topa & sert & vurdu.}
                    {man  & ball-{\em Dat} & hard & hit-{\em Past-3Sg}}
                    {'The man hit the ball hard.'}
\item[b.]\shortex{3}{Kitab{\i}n & arabadayd{\i}.}
                    {book-{\em 2SP} & car-{\em Loc-Aux}}
                    {'(Your) book was in the car.'}
\item[c.]\shortex{4}{Benimki & en & h{\i}zl{\i} & arabad{\i}r.}
                    {I-{\em Gen-Pro} & most & fast & car-{\em Cop}}
                    {'Mine(my car) is the fastest car.'}
\item[d.]\shortex{3}{G\"{o}ky\"{u}z\"{u} & hep & mavidir.}
                    {sky & always & blue-{\em Cop}}
                    {'The sky is always blue.'}
\item[e.]\shortex{4}{Ay\c{s}e'nin & iki & \c{c}ocu\u{g}u & var.}
                    {Ay\c{s}e-{\em Gen} & two & child-{\em Acc} & exist}
                    {'Ay\c{s}e has two children (there exist
                two children of Ay\c{s}e).'}
\item[f.]\shortex{3}{Sokakta & kimse & yok.}
                    {street-{\em Loc} & nobody & not-exist}
                    {'There is (there exists) no one on the street.'}
}

\subsection{Subcategorization}
Every verb---except the intransitives---subcategorize for a noun group
or a set of noun groups. These noun groups may be in accusative
(\ex{1}a), dative (\ex{1}b), locative (\ex{1}c), ablative (\ex{1}d) or
instrumental/commitative case (\ex{1}e).

\myeenumsentence{
{\exfont
\item[a.]\shortex{3}{Raporu & hen\"{u}z & bitirmedik.}
{report-{\em Acc} & yet & finish-{\em Neg-Past-1Pl}}
{`We haven't finished the report yet.'}
\item[b.]\shortex{3}{Yar{\i}n & sinemaya & gidelim.}
{tomorrow & cinema-{\em Dat} & go-{\em Wish-1Pl}}
{`Let's go to the cinema tomorrow.'}
\item[c.]\shortex{4}{Buzdolab{\i}nda & hi\c{c}bir & \c{s}ey & kalmam{\i}\c{s}t{\i}.}
{refrigerator-{\em Loc} & no-at-all & thing & remain-{\em Neg-Past-Asp-3Sg}}
{`Nothing was left at all in the refrigerator.'}
\item[d.]\shortex{4}{Atakule'den & d\"{o}nerken & Evrim'i & g\"{o}rd\"{u}m.}
{Atakule-{\em Abl} & return-{\em Adv} & Evrim-{\em Acc} & see-{\em Past-1Sg}}
{`I saw Evrim as I was coming back from Atakule.'}
\item[e.]\shortex{3}{\c{C}ocuklar & oyuncaklar{\i}yla & oynuyorlar.}
{child-{\em Plu} & toy-{\em 3PP-Ins} & play-{\em Prog-3Pl}}
{`The children are playing with their toys.'}
\shortex{4}{Ufuk & bir & arkada\c{s}{\i}yla & \c{c}al{\i}\c{s}acak.}
{Ufuk & a & friend-{\em 3SP-Ins} & work-{\em Fut-3Sg}}
{`Ufuk will work with a friend of his.'}
}
}

The number of required noun groups depend on the valency of the verb.\\

{\bf Transitive verb:}
\hspace{2cm}\parbox[t]{5cm}{{\exfont \shortex{2}{Kitap & okuyordu.}
{book & read-{\em Prog-Asp-3Sg}}
{`He/she was reading a book.'}
}}

{\bf Ditransitive verb:}
\hspace{1cm}\parbox[t]{5cm}{{\exfont \shortex{3}{Mehmet'e & gitar{\i}m{\i} & verdim.}
{Mehmet-{\em Dat} & guitar-{\em 1SP-Acc} & give-{\em Past-1Sg}}
{`I gave my guitar to Mehmet.'}
}}

More noun groups may be provided to increase the amount of information
provided; they act as complements.
\myeenumsentence{
{\exfont
\item[]\shortex{5}{Sand{\i}klar{\i} & \.{I}zmir'den & Samsun'a & gemiyle & yollad{\i}k}
{chest-{\em Plu}-{\em Acc} & \.{I}zmir-{\em Abl} & Samsun-{\em Dat} & ship-{\em Ins}
& send-{\em Past-2Pl}}
{`We sent the chests from \.{I}zmir to Samsun by ship.'}
}
}

Some verbs subcategorize for a complement clause:\\
{\bf s\"{o}ylemek}:``to say'', {\bf s\"{o}z vermek}:``to promise'', {\bf iddia
etmek}:``to claim'', {\bf inanmak}:``to believe'', {\bf zannetmek}:``to
assume'', {\bf tahmin etmek}:``to guess'', {\bf sanmak}:``to suppose'',
{\bf ispat etmek}:``to prove'', {\bf inkar etmek}:``to deny'', {\bf yemin
etmek}:``to swear'', {\bf d\"{u}\c{s}\"{u}nmek}: ``to think'', {\bf emin olmak}:``to be
sure'', {\bf ku\c{s}kulanmak}:``to suspect'' etc.\\

\myeenumsentence{
{\exfont
\item[a.]\shortex{3}{Dosyay{\i} & bulaca\u{g}{\i}na & s\"{o}z vermi\c{s}tin.}
{file-{\em Acc} & find-{\em Part-2Sg-Dat} & promise-{\em Past-Asp-2Sg}}
{`You had promised that you would find the file.'}
\item[b.]\shortex{3}{Randevumuzu & unuttu\u{g}umu & iddia ediyor.}
{Appointment-{\em 1PP-Acc} & forget-{\em Part-1Sg-Acc} & claim-{\em Prog-3Sg}}
{`She/he claims that I forgot our appointment.'}
}
}

Some verbs ({\bf ye\u{g}lemek}:``to prefer'', {\bf kabul etmek}:``to accept'',
{\bf \c{c}al{\i}\c{s}mak}:``to try'', {\bf \c{c}abalamak}:``to struggle'',
{\bf al{\i}\c{s}mak}:``to get accustomed to'', {\bf \"{o}zenmek}:``to desire'',
{\bf karar vermek}:``to decide'', {\bf niyetlenmek}:``to intend'',
{\bf bahsetmek}:``to mention'', {\bf vaz\-ge\c{c}\-mek}:``to give up'',
{\bf anlamak}:``to understand'' etc.) subcategorize for an infinitive
form of the verb.

\myeenumsentence{
{\exfont
\item[a.]\shortex{4}{Bu & i\c{s}i & bitirmeye & s\"{o}z verdik.}
{this & job-{\em Acc} & finish-{\em Inf-Dat} & promise-{\em Past-1Pl}}
{`We promised to finish this job.'}
\item[b.]\shortex{4}{Bug\"{u}n & al{\i}\c{s}veri\c{s} & yapmaktan & vazge\c{c}tik.}
{today & shopping & make-{\em Inf-Abl} & give-up-{\em Past-1Pl}}
{`We gave up (the idea of) shopping today.'}
}
}

\subsection{Auxiliary verbs}
In Turkish, some verbs are composed of a noun and an auxiliary verb. The
auxiliary verbs used in such constructions are {\bf etmek}:``to do'' and
{\bf yapmak}:``to make'', the former being more frequent.\\
\myeenumsentence{
{\exfont
\item[a.]
\begin{tabular}[t]{lcl}
alay & $\rightarrow$&alay etmek \\
`mockery' & & `to mock'
\end{tabular}
\item[b.]
\begin{tabular}[t]{lcl}
kabul & $\rightarrow$ &kabul etmek \\
`acceptance' & & `to accept'
\end{tabular}
\item[c.]
\begin{tabular}[t]{lcl}
al{\i}\c{s}veri\c{s} & $\rightarrow$ & al{\i}\c{s}veri\c{s} yapmak\\
`shopping' & & `to shop/to do shopping'
\end{tabular}
}
}

There is another auxiliary which attaches to nouns to form nominal
predicates: {\bf olmak} (``to be''). This auxiliary differs from {\bf
etmek} and {\bf yapmak} in two respects. First, it does not appear as a
separate word, but rather a morpheme when the sentence is in past or
present tense. Second, its inflection does not resemble to that of a verb
but the copula.

\myeenumsentence{
{\exfont
\item[a.]\shortex{4}{Babam & ge\c{c}en & ay & yurtd{\i}\c{s}{\i}ndayd{\i}.}
{father-{\em 1SP} & last & month & abroad-{\em Loc-Aux}}
{`My father was abroad last month.'}
\item[b.]\shortex{4}{\"{U}\c{c} & g\"{u}nd\"{u}r & \c{c}ok & uykusuzdum.}
{three & day & very & sleepless-{\em Aux}}
{`I have been very sleepless for three days.'}
\item[c.]\shortex{3}{Kitaplar & masan{\i}n & \"{u}st\"{u}ndeymi\c{s}.}
{book-{\em Plu} & table-{\em Gen} & top-{\em 1SP-Loc-Aux}}
{`The books were on the table.'}
}
}

This auxiliary is not present for third person form if the sentence
is in the present tense.

\myeenumsentence{
{\exfont
\item[]\shortex{2}{Bardaklar & rafta.}
{glass-{\em Plu} & shelf-{\em Loc-(Cop)}}
{`The glasses are on the shelf.'}
}
}

For future tense as well as conditional and necessitative forms, {\bf
olmak} succeeds the noun as a separate word.

\myeenumsentence{
{\exfont
\item[a.]\shortex{5}{Babam & ge\c{c}en & ay & yurtd{\i}\c{s}{\i}nda & olmasayd{\i}...}
{father-{\em 1SP} & last & month & abroad-{\em Loc} & be-{\em Neg-Cond-Asp}}
{`If my father weren't abroad last month...'}
\item[b.]\shortex{4}{Kitaplar & masan{\i}n & \"{u}zerinde & olacak.}
{book-{\em Plu} & table-{\em Gen} & top-{\em 1SP-Loc} & be-{\em Fut-3Sg}}
{`The books will be on the table.'}
}
}

Another point to be emphasized about this auxiliary is that it has a
different negative form than the other verbs when the sentence is in
past or present tense. Negativization is performed by introducing the
word {\bf de\u{g}il} (``not'') just after the nominal. The tense marker, if
exists, attaches to {\bf de\u{g}il}.

\myeenumsentence{
{\exfont
\item[]\shortex{3}{Cem & evde & de\u{g}ildi.}
{Cem & house-{\em Loc} & not-{\em Aux-3Sg}}
{`Cem wasn't at home.'}
}
}

An ambiguity may arise with negative questions of predicates. This
ambiguity is resolved by stress in speech and by a comma preceding
{\bf de\u{g}il} in writing.

\myeenumsentence{
{\exfont
\item[]\shortex{4}{Kedi & bah\c{c}ede & de\u{g}il & mi?}
{cat & garden-{\em Loc} & not & {\em Ques}}
{\shortstack{`The cat is in the garden, isn't it?'\\
`Isn't the cat in the garden?'}}
}
}

\subsection{Existential predicates}
Existential predicates are formed using {\bf var} (``existent'') and {\bf
yok} (``non-existent'').

\myeenumsentence{
{\exfont
\item[a.]\shortex{4}{Odada & d\"{o}rt & koltuk & vard{\i}.}
{room-{\em Loc} & four & armchair & exist-{\em Aux}}
{`There were four armchairs in the room.'}
\item[b.]\shortex{3}{Burada & kimse & yok.}
{here & anybody & non-existent}
{`There isn't anybody here.'}
\item[c.]\shortex{2}{Arabas{\i} & yokmu\c{s}.}
{car-POSS & non-existent}
{`She/he doesn't have a car.'}
}
}

{\bf var} and {\bf yok} cannot be used in future, conditional or
necessitative forms. For these cases, {\bf olmak} replaces {\bf var} and
{\bf yok}.

\myeenumsentence{
{\exfont
\item[a.]\shortex{4}{Odada & d\"{o}rt & koltuk & olmal{\i}}
{room-{\em Loc} & four & armchair & be-{\em Nec}}
{`There should be four armchairs in the room.'}
\item[b.]\shortex{3}{Burada & kimse & olmayacak.}
{here & nobody & be-{\em Fut-3Sg}}
{`There won't be anybody here.'}
}
}

\subsection{Infinitive form of the verbs}
Infinitive form of a verb is formed with suffix {\tt -mAk} attached to
the stem.

\myeenumsentence{
{\exfont
\item[]\shortex{5}{Fransa'ya & gitmek & \c{c}ok & para & ister.}
{France-{\em Dat} & go-{\em Inf} & much & money & require-{\em Pres}}
{`Going to France costs much'}
}
}

Infinitive suffix cannot be followed by genitive or possesive suffixes.
However case suffixes are allowed.

\myeenumsentence{
{\exfont
\item[a.]\shortex{2}{Ko\c{s}maktan & yoruldum.}
{run-{\em Inf-Abl} & tire-{\em Pass-Past-1Sg}}
{`I got tired of running.'}
\item[b.]\shortex{2}{K{\i}zmakta & hakl{\i}s{\i}n.}
{get-angry-{\em Inf-Loc} & right-{\em Cop(2Sg)}}
{`You are right to be angry.'}
}
}

\subsection{Gerundive forms of the verbs}
There are a couple of suffixes for producing gerundive forms of a verb:
{\tt -mA} and {\tt -(y)H\c{s}}. {\tt -mA} is used for referring to
the action or its result. Genitive and possesive suffixes can be attached 
to {\tt
-mA}.

\myeenumsentence{
{\exfont
\item[a.]\shortex{5}{Onunla & g\"{o}r\"{u}\c{s}menin & bana & faydas{\i} & olmaz.}
{he/she-{\em Gen-Ins} & meet-{\em Ger-Gen} & me & use & be-{\em Neg-Pres-3Sg}}
{`Meeting with him/her is of no use to me.'}
\item[b.]\shortex{2}{Okumas{\i} & d\"{u}zeliyor.}
{read-{\em Ger-3SP} & improve-{\em Prog-3Sg}}
{`His/her reading is improving.'}
}
}

{\tt -(y)H\c{s}} produces a gerundive which emphasizes the manner the action is
performed. This suffix can also be succeeded by genitive and possesive
suffixes.

\myeenumsentence{
{\exfont
\item[]\shortex{2}{G\"{u}l\"{u}\c{s}\"{u}n\"{u} & hat{\i}rl{\i}yorum.}
{smile-{\em Inf-3SP-Acc}& remember-{\em Prog-1Sg}}
{`I remember the way you/(s)he smile/s.'}
}
}

\subsection{Syntax of causative verbs}
In Turkish, verbs are causativized by attaching the causative suffixes
{\tt -DHr, -Hr, -t, -Ht} and {\tt -Ar} to the stem. Using these suffixes,
one can obtain a causative verb almost from any verb, including the
causatives themselves.

\myeenumsentence{
{\exfont
\item[a.]
\begin{tabular}[t]{l@{\ }c@{\ }l}
inanmak & $\rightarrow$ & inand{\i}rmak \\
`to believe' & & `to persuade'
\end{tabular}
\item[b.]
\begin{tabular}[t]{l@{\ }c@{\ }l}
do\u{g}mak& $\rightarrow$ & do\u{g}urmak\\
`to be born' & & `to give birth to'
\end{tabular}
\item[c.]
\begin{tabular}[t]{l@{\ }c@{\ }l}
oturmak& $\rightarrow$ & oturtmak\\
`to sit' & & `to seat'
\end{tabular}
\item[d.]
\begin{tabular}[t]{l@{\ }c@{\ }l}
korkmak& $\rightarrow$ & korkutmak\\
`to fear' & & `to frighten'
\end{tabular}
\item[e.]
\begin{tabular}[t]{l@{\ }c@{\ }l}
\c{c}{\i}kmak& $\rightarrow$ & \c{c}{\i}kartmak\\
`to go out/to go up' & & `to remove/to raise'
\end{tabular}
}
}

Appropriate combinations of causative suffixes allow production of
multiple causatives:

\myeenumsentence{
{\exfont
\item[]
\begin{tabular}[t]{l@{\ }c@{\ }l@{\ }c@{\ }l}
\"{o}lmek& $\rightarrow$ & \"{o}ld\"{u}rmek & $\rightarrow$ & \"{o}ld\"{u}rtmek\\
`to die' & & `to kill'& & `to have someone killed'
\end{tabular}
}
}

From syntactical point of view, causativization process has two important
results: increase in the valency of the verb, and the changes in 
the grammatical functions of the noun groups.\\

Any verb form of valency $n$ will require $n+1$ noun groups after 
causativization:\\

{\exfont
{\bf Intransitive verb:}\begin{tabular}[t]{l@{\ }c@{\ }l}
uyumak & $\rightarrow$ & uyutmak\\
`to sleep' & & ` to send to sleep'
\end{tabular}\\
{\bf Transitive verb:}\begin{tabular}[t]{l@{\ }c@{\ }l}
bir \c{s}eyi okumak & $\rightarrow$ & birine bir\c{s}eyi okutmak\\
`to read something' & & ` to make someone read something'
\end{tabular}\\
{\bf Ditransitive verb:}\\
\begin{tabular}[t]{l@{\ }c@{\ }l}
bir \c{s}eyi bir yere koymak & $\rightarrow$ & birine bir\c{s}eyi bir yere
koydurtmak\\
`to put something to somewhere' & & ` to make someone put something to
somewhere'
\end{tabular}
}

Another effect of causativization is that the noun groups of the
original clause change their grammatical functions:

{\bf Causativization of an Intransitive Verb:} The subject of the
intransitive verb becomes the direct object (accusative-marked noun
group) in the causative clause. A new noun group is introduced for subject
position of the causative clause.

\myeenumsentence{
\item[]
\begin{tabular}[t]{l@{\ }c@{\ }l}
Yi\u{g}itcan g\"{u}ld\"{u} & $\rightarrow$ & I\c{s}{\i}k Yi\u{g}itcan'{\i} g\"{u}ld\"{u}rd\"{u} \\
NOM   & & NOM  ACC  \\
`Yi\u{g}itcan laughed' & & `I\c{s}{\i}k caused Yi\u{g}itcan to laugh'
\end{tabular}
}

{\bf Causativization of a Transitive Verb:} The subject of the
transitive verb becomes the dative-marked indirect object
in the causative clause, whereas the direct object 
(e.g., {\em \c{s}iir} in \ex1)) preserves its
grammatical function in the causative clause. Also, a new noun group
is introduced for subject position.

\myeenumsentence{
{\exfont
\item[]
\begin{tabular}[t]{l@{\ }c@{\ }l}
Arzu \c{s}iir okudu & $\rightarrow$ & \"{O}\u{g}retmen Arzu'ya \c{s}iir okuttu\\
NOM  NOM  & & NOM  DAT   NOM \\
`Arzu read a poem' & & `The teacher made Arzu read a poem'
\end{tabular}
}}

The indirect object of the causative clause may sometimes be omitted:

\myeenumsentence{
{\exfont
\item[]\shortex{3}{\"{O}\u{g}retmen & \c{s}iir & okuttu.}
{teacher & poem & read-{\em Caus-Past-3Sg}}
{`The teacher caused a poem to be read.'}
}
}

If the main verb subcategorizes for a dative noun group, this noun group
remains unaltered in the causative clause. In such a case, the subject
of the main verb is marked as accusative and becomes the direct
object of the causative clause.

\myeenumsentence{
{\exfont
\item[]
\begin{tabular}[t]{l@{\ }c@{\ }l}
\c{C}ocuk okula ba\c{s}lad{\i} & $\rightarrow$ & \c{C}ocu\u{g}u okula ba\c{s}latt{\i}k\\
NOM  DAT & & ACC   NOM \\
`The child started school' & & `We made the child start school'
\end{tabular}
}
}

{\bf Causativization of a Ditransitive Verb:} The subject of the
ditransitive verb becomes the dative-marked indirect object
in the causative clause, whereas the accusative-marked direct object
and the dative-marked object of the main verb preserve their grammatical
functions in the causative clause. Subject position is again filled
by a new noun group.

\myeenumsentence{
{\exfont
\item[]
\begin{tabular}[t]{l@{\ }c@{\ }l}
Hakan kitab{\i} masaya koydu & $\rightarrow$ & Ali Hakan'a kitab{\i} masaya koydurdu\\
NOM  ACC  DAT & & NOM  DAT   ACC  DAT \\
`Hakan put the book on the table' & & `Ali made Hakan put the book on the table'
\end{tabular}
}
}

Just as for the tansitive verb, the subject of the main verb may
be omitted:

\myeenumsentence{
\item[]\shortex{4}{Ali & kitab{\i} & masaya & koydurdu.}
{Ali & book-{\em Acc} & table-{\em Dat} & put-{\em Caus-Past-3Sg}}
{}
}


\chapter{DESIGN}
\label{ChDes}

Every language is expected to have a different realization of the
language-independent principles. Some solutions proposed by Pollard and
Sag \cite{PolSag94} for English have to be modified to model Turkish.
The basic features that distinguish Turkish from English are:
importance of morphology in the specification of grammatical
functions, overt case and agreement marking, final head position, free
constituent order, pronoun drop, complement drops, and the nature of
unbounded dependency constraints.  Some of these items are resolved by
extending the sign structure and introducing Turkish-specific versions
of some principles.

\savebox{\tmp}{
{\scriptsize
\begin{avm}
\[ {\aS cat} \\
   CAT & \[ {\aS cat} \\
	    HEAD & {\aS head} \\
	    SUBJ & {\aS synsem} \\
	    SUBCAT & {\aS subcat-type}
	 \]\\
   CONT & {\aS sem-obj}
\]
\end{avm}
}
}

\section{Sign Structure}  

\begin{figure}
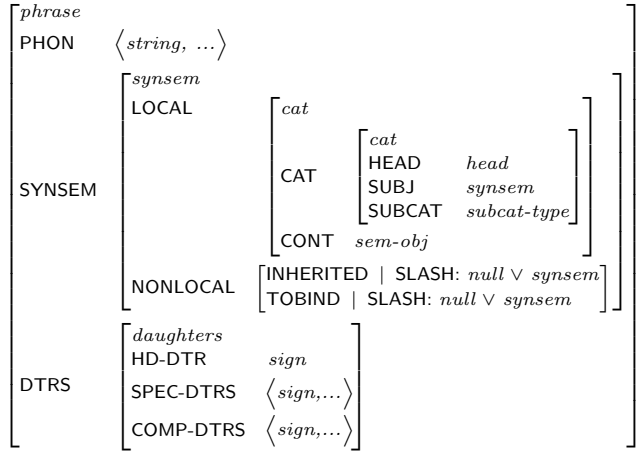

{\scriptsize
\begin{avm}
\[ {\aS phrase} \\
   PHON & \<{\aS string, ...}\> \\
   SYNSEM & \[ {\aS synsem} \\
	       LOCAL & \usebox{\tmp} \\
	       NONLOCAL & \[ INHERITED \| SLASH: {\aS null} $\vee$ 
						     {\aS synsem}\\
			     TOBIND \| SLASH: {\aS null} $\vee$ 
				  		     {\aS synsem}
			  \]
	    \] \\
    DTRS & \[ {\aS daughters} \\
	      HD-DTR & {\aS sign} \\
	      SPEC-DTRS & \< {\aS sign,...}\> \\
	      COMP-DTRS & \< {\aS sign,...}\> 
	   \]
\]

\end{avm}
}
\caption{Sample sign structure for Turkish} 
\label{Turkishsign}
\end{figure}

The template for phrases is given in Figure~\ref{Turkishsign}.
A structure similar to that of English is used for signs in Turkish,
with minor changes. In category ({\aS cat}) type, an additional feature
{\aF SUBJ} is added, which is coindexed with the subject complement if
it exists in the subcat list. It is used to distinguish subject
complement from other complements which may have the same case and
agreement. It refers to the subject complement directly.
This is necessary for implementing some syntactic phenomena like
subject raising relative clauses and other bindings to subject
complement. Example (\ex{1}) shows a simplified version of the {\aF CAT} 
feature of a lexical entry for the verb ``seviyor'' (love-{\em Prog-3Sg\/}).

\myenumsentence{
{\small
\begin{avm}
\[ {\aS word}\\
   PHON \; \< {\rm ``seviyor''} \> \; \; {\rm \% love-{\em Prog-3Sg}} \\
   SYNSEM \| CAT \; \[ {\aS cat}\\
		       HEAD & {\aS head} \\
		       SUBJ & \@1 \\
		       SUBCAT & \{ \@1 {\rm NP[nom], NP[acc]}\}
		   \]
\]
\end{avm}
}
}

One of the major differences in sign structure is the {\aF SUBCAT}
feature.  Since Turkish has free constituent order,\footnote{
For more information on order-freeness, see Chapter~\ref{ChTurkish} and
Section~\ref{SeComp}.
}
a type which is a nested combination of ordered and unordered
lists is used instead of lists indicating obliqueness of the
complements. Unordered lists are denoted with curly braces similar to
sets.

Only the {\aF SLASH} feature of type {\aS null $\vee$ local} is defined
as a nonlocal feature. Nonlocal features are used to process the
information coming from arbitrary daughters (not only head daughter)
which will be transmitted to upper phrases and bound to outer
structures.  Unbounded dependency and other binding constraints are
defined by nonlocal features. HPSG defines three basic nonlocal
features: {\aF SLASH, REL} and {\aF QUE} which are used to implement
filler-gap dependencies, relative clauses and questions respectively.
In this study, we only used the {\aF SLASH} feature.

Since adjuncts may exist in any position preceding the head and probably
between any subcategorized constituent, we have chosen  to combine adjunct
and complements of the phrase into one ``daughters'' attribute.
Daughters ({\aF DTRS}) attribute consists of adjunct daughters ({\aF
ADJ-DTRS}) and complement daughters ({\aF COMP-DTRS}) which are 
lists of {\aS sign}. Head daughter ({\aF HD-DTR}) is of type {\aS sign}.
If phrase has a subject complement, there is another feature subject
complement ({\aF SUBJ-COMP}) which is coindexed with the
subject in the {\aF COMP-DTRS} list.

\section{Major Categories and Head Features}

Since derivations involving category change is possible in Turkish,
(e.g.  relativizer {\tt -ki} turning a noun into a specifier), and
derived words preserve their syntactic behavior, head features must be
extendable. For example, finite verbs are not nominalized hence do not
carry case. However, sentential complements have inflections which make
them behave as nominals and take case:

\myenumsentence{
\shortex{4}{Eve & girdi\u{g}imizdeki & manzaray{\i} & g\"{o}rd\"{u}n\"{u}z.}
	   {house-{\em Dat} & enter-{\em Part-1Pl-Loc-Rlvz} & view-{\em Acc} & 
					see-{\em Past-2Pl}}
	   {`You saw the view of the house when we entered.'}
}

Basic major categories are: nouns, verbs, adjectives, adverbs and
conjunctives. Category information consists of {\aF HEAD, SUBJ} and
{\aF SUBCAT} features. Each category has its own set of appropriate
features in head attribute. {\aF HEAD} feature is of type {\aS head}
which has the following categories defined as subsorts.

\begin{figure}[thb]
\centerline{\psfig{file=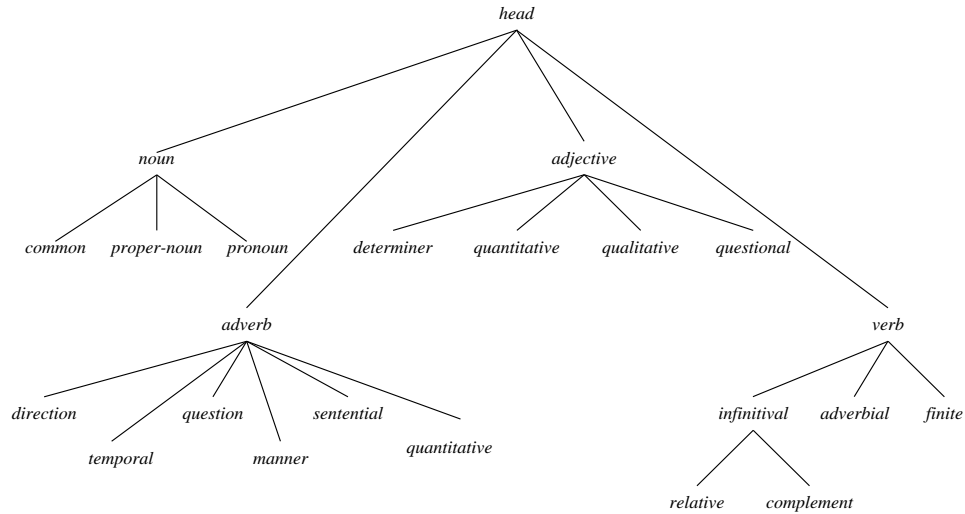,width=5in}}
\caption{Sort hierarchy of type {\aS head}}
\label{Headsort}
\end{figure}

Appropriate features for the head sort {\aS noun} are case, agreement,
relativized, nominal-index and possession (cf. \ex{1}). {\aF CASE} can be
nominative, objective, genitive, dative, ablative, genitive or
instrumental. In Turkish, there is no gender of objects so
agreement({\aF AGR}) consists of person and number information. Since
possession (if noun is used in the possessive group) is overtly marked,
it is included as a head feature of type either {\aS none} or
agreement. Also feature nominal index ({\aF N-IND}) is defined to
handle adjuncts modifying the derived noun (see Section~\ref{SeAdjunct}) and
coindexed with the semantic index of the root noun.

\myenumsentence{
{\scriptsize
\begin{avm}
\[ PHON & \< ``{\rm bah\c{c}elerimizin}'' \> \; \; {\rm \% ``of our garden''} \\
   SYNSEM & \[ CAT \| HEAD & \[ {\aS common-noun}\\
			      CASE \; {\aS genitive} \\
			      AGR \; \[ {\aS agr}\\
				       PERSON & third \\
				       NUMBER & plural
				    \]\\
			      N-IND \; \@1 \\
			      POSS \; \[ {\aS agr}\\
					PERSON & third \\
					NUMBER & plural
				     \]
			    \]\\
	     CONT \| INDEX & {\aS agr} \@1
	  \]
\]
\end{avm}
}
}

As can be seen in Figure~\ref{Headsort}, head sort {\aS verb} has three
subsorts, {\aS finite}, {\aS infinitival} and {\aS adverbial}. Finite
verbs are the heads of the finite sentences. Infinitivals are verbal
heads which modify or specify other phrases or subcategorized by other
heads as complements.  However, they still have the properties of verbs
and construct embedded sentences.  Relative clauses are verbs inflected
by suffixes {\tt -An, -dHk-{\em Poss}, -AcAk+{\em Poss}} and contain a
gap which is filled by a following noun phrase. Another group of verbal
heads are sentential complements which can be arguments of some verbs.
These are verbs inflected by suffixes {\tt -mA(k), -H\c{s}-{\em Poss},
-dHk-{\em Poss}}, and {\tt -AcAk-{\em Poss}} where the possessive suffix
marks agreement rather than possession. Similarly, sentential adverbs
modify the matrix verb.

In addition to the syntactic roles, there are structural differences in
these three sorts. Finite verbs take an secondary tense or aspect
marker which is one of {\aS none, past, dubitative} or {\aS
conditional}. Infinitivals carry case information so they have the {\aF
CASE} feature. The following attribute definitions are appropriate for
these groups:

\myeenumsentence{
\item[]
\begin{tabular}[t]{ll}
{\bf verb:}& \\ \cline{1-1}
{\aF AGR} & {\aS null $\vee$ agr} \\
{\aF NEG} & {\aS plus $\vee$ minus} \\
{\aF TENSE} & {\aS base $\vee$ present $\vee$ continuous $\vee$
		past $\vee$} \\ 
		& \hspace*{1cm} {\aS dubitative $\vee$ wish  } \\
& \\
{\bf finite:}& \\ \cline{1-1}
{\aF AGR} & {\aS agr} \\
{\aF AUX-TENSE} & {\aS past $\vee$ dubitative $\vee$ conditional}\\
& \\
{\bf infinitival:}& \\ \cline{1-1}
{\aF CASE} & {\aS case} 
\end{tabular}
}

\section{Complement Selection and Linear Precedence}
\label{SeComp}

Heads select their arguments using the {\aF SUBCAT} feature. The {\aF
SUBCAT} feature is a structured type consisting of arguments of sort
{\aS synsem}.  Therefore, a head can select any syntactic property of
its arguments like category, case, agreement, nonlocal features and
even semantic content. Any type of category is allowed including
sentential complements, adverbs, adjectives, etc.

As mentioned in the preceding section, scrambling of constituents is handled
by unordered lists. Linear precedence constraints of Turkish can be 
described generally as:

\myeenumsentence{
\item 
{\scriptsize
\begin{avm}
\[ HD-DTR & \@1 \[{\aF SYNSEM\|CAT\|HEAD} \; $\neg$ {\aS verb}\] \\
   COMP-DTRS & \< ...., \@2 {\aS sign} ,....\>
\] \; $\Longrightarrow$ \; \[ \@2 \] \; $<$ \; \[ \@1 \] 
\end{avm}
}
\item
{\scriptsize
\begin{avm}
\[ HD-DTR & \@1 \\
   ADJ-DTRS & \< ...., \@2 {\aS sign} ,....\>
\] \; $\Longrightarrow$ \;
\[ \@2 \] \; $<$ \; \[ \@1 \]
\end{avm}
}
}

(\ex{0}a) describes the constraint ``complement daughters 
should precede the head daughter when head daughter is not a verb'' and 
(\ex{0}b) describes the constraint ``adjunct daughters should precede the 
head daughter''.

To handle the cases such as `nonreferential object should immediately
precede the verb', we use a special sort for {\aS subcat} feature. This
sort has a nested mixed structure of {\aS list} and {\aS
set}.\footnote{ Set is used to indicate the property of order-freeness
where all permutations of the members of the set is possible at the
surface.} Subcat type can be either a list or a set, and recursively,
element of a set can be either a {\aS synsem-arg} value or a {\aS
list}. Element of a list can be either a {\aS synsem-arg} value or a
{\aS set} (\ex{1}b). {\aS synsem-arg} sort is used to enable optional
arguments (\ex{1}a). If {\aF OPT} attribute of an argument is {\aS
plus} than it is optional and can be omitted. Optional arguments are
generally denoted with enclosing parentheses.

\myeenumsentence{\label{SubType}
\item
{\small
\begin{avm}
\[ \avmspan{{\aS synsem-arg}}\\
   {\aF OPT} & {\aS plus $\vee$ minus} \\
   {\aF ARG} & {\aS synsem}
\]
\end{avm}
}
\item
{\small
\begin{avm}
\avml & \node{a}{{\aS subcat-type}} \\
      & \\
      \node{b}{
     \[ \avmspan{{\aS list-subcat}} \\
	{\aF HD} & synsem-arg $\vee$ set-subcat \\
	{\aF TL} & e-list $\vee$ list-subcat
     \]} & & \node{c}{
     \[ \avmspan{{\aS set-subcat}} \\
	{\aF EL} & synsem-arg $\vee$ list-subcat \\
	{\aF ELS} &  e-list $\vee$ set-subcat
     \]}
\avmr
\nodeconnect{a}{b}
\nodeconnect{a}{c}
\end{avm}
}
}

In the surface form, lists are ordered and sets are permuted. For
example, the sign in (\ex{1}a) has the surface forms listed in
(\ex{1}b).  It is assumed that {\em adam} (man) is substituted for the
subject, {\em \c{c}ocu\u{g}a} (child-{\em Dat}) is substituted for the dative object,
{\em evden} (house-{\em Abl}) is for {\em ablative} argument and {\em kalem}
(pencil) is the nonreferential object.

\myeenumsentence{
\item
{\scriptsize
\begin{avm}
\[ PHON & \<`{\rm getirdi}'\> \; {\rm \% bring-{\it Past-3Sg}}\\
   SYNSEM\|LOCAL\|CAT & \[ HEAD & verb\\
			   SUBJ & \@1 \\
			   SUBCAT & \<\{ \@1 {\rm NP[nom], 
					NP[dat],}
			     {\rm NP[abl]} \}, {\rm NP[nom]} \>
			\]
\]
\end{avm}
}
\item
``Adam \c{c}ocu\u{g}a evden kalem getirdi'' \\
``Adam evden \c{c}ocu\u{g}a kalem getirdi'' \\
``\c{C}ocu\u{g}a adam evden kalem getirdi'' \\
``\c{C}ocu\u{g}a evden adam kalem getirdi'' \\
``Evden \c{c}ocu\u{g}a adam kalem getirdi'' \\
``Evden adam \c{c}ocu\u{g}a kalem getirdi'' 
}

Sentential complements and any kind of argument-head relation can be declared
in this manner. For example the verb ``s\"{o}yledi'' (told) can be
defined as:

\myenumsentence{
{\scriptsize
\begin{avm}
\avml
\[ PHON & \<{\rm `s\"{o}yledi'}\> \; \; {\rm \% tell-{\em Past-3Sg}}\\
   SYNSEM\|LOCAL\|CAT & \[ HEAD & {\aV finite-verb} \\
			   SUBJ & \@1  \\
			   SUBCAT &  \{ {\rm NP[nom], NP[dat], S[
			   			inf,acc]} \} 
			\]
\] \\
{\rm Where:} \\
{\rm S[inf,acc]}:\; \[ SYNSEM\|LOCAL\|CAT & 
				   \[ HEAD & \[ {\aV infinitival} \\
						CASE & acc 
					     \] \\
			     	      SUBCAT & \< \; \>
				   \]
			   \]
\avmr
\end{avm}
}
}

Verbal categories have the most complex complement structures. All
verbal heads (finite verbs, infinitivals, sentential adverbs) require
one or more complements according to their valence. Turkish is a
complement-drop language so complements can be dropped even if they are
obligatory arguments. 

Other typical complement-head relationship is in the possessive noun
group (\ex{1}a).  A possessive marked noun subcategorizes for a
genitive noun and the part of speech of the complement should agree
with the possessive suffix (\ex{1}b).

\myeenumsentence{
\item[] \shortex{2}{saray{\i}n&kap{\i}s{\i}}
		   {palace-{\em Gen} & door-{\em 3SP}}
		   {`door of the palace'}
\item[] 
{\scriptsize \begin{avm}
\[ PHON & \< {\rm `kap{\i}s{\i}'} \> \; \; {\rm \% ``door-{\em 3SP}''}\\
   SYNSEM\|LOCAL\|CAT  & \[ HEAD & \[ common \\
				      POSS & \@1 agr
				   \]\\
			    SUBCAT & \< NP$_{gen}$[AGR \@1] \>
			\]
\]
\end{avm}
}
}

The order of the complements and adjuncts are variable which means
adjuncts specifying the head can be in any position. So, instead of
generating the surface form from the subcat list directly by a phrase
structure rule, we chose to retrieve the complements one at a time.
This allows the adjunct rule which will be described in following
sections to be applied to the head at any position.

\myenumsentence{\label{schema1a}
{\scriptsize
\begin{avm}
\avml
\[ SYNSEM\|LOCAL\|CAT & \[ HEAD & \@1 \\
			   SUBCAT & \@2
			\]
\] \; $\longrightarrow$ \\
\hspace*{2cm}
\[ SYNSEM & \@3 \] , \[ SYNSEM\|LOCAL\|CAT & \[ HEAD & \@1 \\
					SUBCAT & \@4
					\]
			   \] , {\tt selectlast(\@3,\@4,\@2)}\\
\vspace{1em}
{\rm Where {\tt selectlast} selects the last {\em synsem} value (\@3) from the 
{\aF SUBCAT} structure (\@4)}, \\
{\rm and rest is stored in third parameter (\@2).}
\avmr
\end{avm}
}
}

This rule applies to the head-final complements. Handling scrambling of
verbal head to pre-complement position is made possible by another
schema:

\myenumsentence{\label{schema1b}
{\scriptsize
\begin{avm}
\avml
\[ SYNSEM\|LOCAL\|CAT & \[ HEAD & \@1 \\
                           SUBCAT & \@2
                        \]
\] \; $\longrightarrow$ \\
\hspace*{2cm}
\[ SYNSEM\|LOCAL\|CAT & \[ HEAD & \@1 {\aS verb} \\
                           SUBCAT & \@4
                         \]
\] \; \[ SYNSEM & \@3 \] , {\tt selectfirst(\@3,\@4,\@2)}\\
\vspace{1em}
{\rm Where {\tt selectfirst} selects the first {\em synsem} value (\@3)
from the {\aF SUBCAT} structure (\@4)}, \\ {\rm and rest is stored in
third parameter (\@2).} 
\avmr 
\end{avm} 
} 
}

\section{Pronoun Drop}

One of the distinct properties of Turkish is the pronoun drop; pronoun
in the subject position can be omitted since it is marked by agreement
of the head. There are three constructs where pronouns
drop: subject of the verbal heads, substantive predicates and 
possessive noun groups. In both cases, including embedded sentences in
which the subject has genitive case, the dropped pronoun has either
nominal or genitive case.

\myeenumsentence{
\item
\shortex{3}{(biz) & Treni & g\"{o}rd\"{u}k.}
	   {We & train-{\em Acc} & see-{\em Past-1Pl}}
	   {`We saw the train.'}
\item
\shortex{3}{(benim) & G\"{u}zel & bah\c{c}em.}
	   {I-{\em Gen} & nice & garden-{\em 1SP}}
	   {`My beautiful garden.'}
\item
\shortex{5}{(o)& (benim) & Eve & gitti\u{g}imi & g\"{o}rd\"{u}.}
	   {he & I-{\em Gen} & house & go-{\em Part-1Sg-Acc}& 
			see-{\em Past-3Sg}}
	   {`He saw that I went to house.'}
\item
\shortex{5}{(o)& (benim) & En & yak{\i}n & arkada\c{s}{\i}md{\i}r.}
	   {(he) & (I-{\em Gen}) & most & close & friend-{\em 1SP-Cop(3Sg)}}
		      {`He is my best friend.'}
}

A solution to pro-drop is using {\em empty categories}, which have
null surface forms. A possible declaration for dropped pronoun as empty
category is: 

\myenumsentence{
{\scriptsize
\begin{avm}
\[ PHON & \< \; \>\\
   SYNSEM\|LOCAL\|CAT & \[ HEAD & \[ {\aS pronoun} \\
				     CASE & {\aS nominative $\vee$
				       genitive}
				  \] \\
			   SUBCAT & \<\;\>
			\]
\]
\end{avm}
}
}

This declaration will fill the subject position required by any head
feature. However, empty categories usually cause major problems. In most
of the implementations, they are inserted into any position available in
the sentence. This is simply inefficient. More critically, when the
order of the complement filled by the empty category has free order as
it is in Turkish, superfluous parses are generated for each possible
position that the subject can occupy. Therefore more constraints may be
necessary to deal with the empty categories. The same problem also
exists for management of the trace in relative clauses
(Section~\ref{SeRelCl}). For the time being we have chosen the keep
dropped pronouns as empty categories.

\section{Adjuncts}
\label{SeAdjunct}

Adjuncts are optional elements in the phrase structure.  Adjuncts
cannot be modeled in the same way as the complements.  Their most
distinct property is that they do not change the valence of the phrase
they combined with. In other words, a head can be specified/modified by
any number of adjuncts, which may possibly have the same category.

Another problem about adjuncts is whether the heads should select their
adjuncts or adjuncts should select their heads.  One solution proposed
by Pollard and Sag \cite{PolSag87} takes the approach where heads
select their adjuncts. A new set-typed feature called {\em adjuncts} is
added to sort {\em cat}, and adjunct is checked by whether it is
unified with one of the elements of the set. The number of elements
in the set does not change. However, adjuncts may come in many
different varieties and this set may grow to an unmanageable size.

In the other approach, adjuncts select their heads \cite{PolSag94}.
This provides a simpler solution because the heads that an adjunct can
modify are more restricted. {\aF MOD} attribute of type {\aS synsem}
defined in the lexical entry fot the adjunct is used to select the
syntactic category of the head. {\aF MOD} is a head feature containing
the restrictions for the head to be modified, and is unified with the
{\aF SYNSEM} value of the head. In examples (\ex{1}a--b), the adjunct
category of adjective/adverb subcategorizes for the head noun/verb
respectively.

\myeenumsentence{
\item
{\scriptsize
\begin{avm}
\[ PHON & \< {\rm `mavi'} \> \; \; {\rm \% blue} \\
   SYNSEM\|LOCAL\|CAT & \[ HEAD & \[ {\aS qualitative-adj} \\
					  MOD & \[ LOCAL\|CAT\|HEAD &
							{\aV noun}
						\]
				  \]
			\]
\]
\end{avm}
}
\item
{\scriptsize
\begin{avm}
\[ PHON & \< {\rm `\c{c}abuk'} \> \; \; {\rm \% fast}\\
   SYNSEM\|LOCAL\|CAT & \[ HEAD & \[ {\aS adverb} \\
					  MOD & \[ LOCAL\|CAT\|HEAD &
							{\aV verb}
						\]
				 \]
			\]
\]
\end{avm}
}
}

With this model, all adjuncts have similar structure and can be handled
by the same rule. In Turkish, an adjunct with the appropriate {\aF MOD}
attribute can precede the phrase anywhere. So a preliminary version of
adjunct principle can be written as:

\myenumsentence{
{\scriptsize
\begin{avm}
\avml
\[ DTRS \; \[ HD-DTR & \@1 \\
	     ADJ-DTRS & \@2 $\oplus$ \@3 \\
	  \]\\
   SYNSEM\|LOCAL\|CAT\|HEAD \; \@4
\] \; $\longrightarrow$ \; \\
\hspace{1cm}
\@3 \[ SYNSEM\|LOCAL\|CAT\|HEAD\|MOD \@5 \] ,
\@1 \[ SYNSEM & \@5 \[ LOCAL\|CAT\|HEAD & \@4 \] \\
       DTRS & \[ ADJ-DTRS & \@2 \]
    \]
\avmr
\end{avm}
}
}

Although adjuncts can modify a phrase in any preceding position, there
are restrictions on the possible combinations and order of the adjuncts
modifying the same head. Rules defining the grammatical combinations
vary; an adjunct modifying the head may prevent other adjuncts to
modify the same head. In (\ex{1}a) ``g\"{u}zel'' modifies ``bah\c{c}edeki'' and
does not modify ``\c{c}i\c{c}ek''. Similarly, the quantitative adjective
``iki'' cannot modify the noun phrase ``bu \c{c}i\c{c}ek''. However ``iki''
does not prevent ``bu'' from specifying ``\c{c}i\c{c}ek'' (\ex{1}b--c).

\myeenumsentence{
\item \shortex{3}{g\"{u}zel & bah\c{c}edeki & \c{c}i\c{c}ek}
	   {beautiful & garden-{\em Loc-Rlvz} & flower}
	   {`The beautiful flower in the garden'}
\item * \shortex{3}{iki & bu & \c{c}i\c{c}ek}
	     {two & this & flower}
	     {}
\item \shortex{3}{bu & iki & \c{c}i\c{c}ek}
	   {this & two & flower}
           {'These two flowers'}
}

In order to control the combinations of adjuncts, we introduce a new
feature for all categories under the {\aF CAT} feature called {\aF
ADJUNCTS}. This structure consists of a group of boolean attributes
that keep track of the adjuncts that have been applied to the category.
In the adjunct part, the {\aF MOD} attribute is divided into two
attributes: a {\aS synsem} value ({\aF MODSYN}) with the same purpose of {\aF
MOD} in (\ex{-1}), and {\aF MODADJ} defining the resulting {\aF
ADJUNCTS} structure which will be projected to the mother phrase.
Adjunct still selects the head together with the {\aF ADJUNCT} value
included in the {\aF SYNSEM} of the head, and defines which flags will
be set and passed to the mother phrase. For example, assume that {\aF
ADJUNCTS} consist of three flags:  {\aF RLV} indicating that the
relativized noun has been applied, {\aF DEM} indicating the
demonstrative adjective has been applied and {\aF QLT} indicating that
the qualitative adjective is applied. Simplified lexical entries for
each category could be as in the example (\ex{1}).

\myeenumsentence{
\item
{\scriptsize
\begin{avm}
\[ PHON & \< {\aV `g\"{u}zel' }\> \; \; {\em \% beautiful}\\
   SYNSEM\|LOCAL\|CAT\|HEAD\|MOD & \[ MODSYN & 
			{\aF LOCAL\|CAT\|ADJUNCTS} & \[ RLV & $-$ \\
						  DEM & $-$ 
					       \] \\
				     MODADJ &
			\[ RLV & $-$ \\
			   DEM & $-$ \\
			   QLT & $+$ 
			\]
		\]
\]
\end{avm}
}
\item
{\scriptsize
\begin{avm}
\[ PHON & \< `bu' \> \; \; {\em \% this}\\
   SYNSEM\|LOCAL\|CAT\|HEAD\|MOD & \[ MODSYN & 
			{\aF LOCAL\|CAT\|ADJUNCTS} & \[ RLV & $-$ \\
						  DEM & $-$ \\
						  QLT & \@1
					       \] \\
				     MODADJ &
			\[ RLV & $-$ \\
			   DEM & $+$ \\
			   QLT & \@1 
			\]
		\]
\] 
\end{avm}
}
\item
{\scriptsize
\begin{avm}
\[ PHON & \< `bah\c{c}edeki' \> \; \; {\em \% one that is in the garden}\\
   SYNSEM\|LOCAL\|CAT\|HEAD\|MOD & \[ MODSYN & 
			{\aF LOCAL\|CAT\|ADJUNCTS} & \[ RLV & $-$ \\
						  DEM & \@1 \\
						  QLT & \@2
					       \] \\
				     MODADJ &
			\[ RLV & $+$ \\
			   DEM & \@1 \\
			   QLT & \@2
			\]
		\]
\] 
\end{avm}
}
}

With these definitions, a revised adjunct principle can be written as:

\myenumsentence{\label{schema2}
{\scriptsize
\begin{avm}
\avml
\[ DTRS \; \[ HD-DTR & \@1 \\
	     ADJ-DTRS & \@2 $\oplus$ \@3 
	  \] \\
   SYNSEM\|LOCAL\|CAT \; \[ HEAD & \@4 \\
			    ADJUNCTS & \@6
			 \]
\] \; $\longrightarrow$ \; \\
\hspace{2cm}
\@3 \[ SYNSEM\|LOCAL\|CAT\|HEAD\|MOD & \[ MODSYN & \@5 \\
					   MODADJ & \@6 \]
    \],\\
\hspace{2cm}
\@1 \[ SYNSEM & \@5 \[ LOCAL\|CAT\|HEAD & \@4 \] \\
       DTRS & \[ ADJ-DTRS & \@2 \]
    \]
\avmr
\end{avm}
}
}

When relative clauses, quantifiers, article `bir', classifier nouns, and 
quantitative adjectives are defined, all noun phrase combinations can
be covered. On the other hand, genitive noun in possessive noun group
is not a specifier. It is an argument of the possessive noun. Thus it
requires a special interpretation. Specifiers and modifiers can
specify/modify the possessive marked noun as long as they are between
the genitive noun and the possessive noun. Otherwise they
specify/modify the genitive noun. To prevent adjuncts from passing over
the genitive noun, we defined another constraint which can be
informally expressed as: ``a noun modifier/specifier modify/specify a
possessive marked noun if it is not saturated''. This constraint can be
shown as:

\myenumsentence{\label{AdjPoss}
{\scriptsize
\begin{avm}
\avml
\[ SYNSEM\|LOCAL\|CAT\|HEAD\|MOD\|LOCAL\|CAT \; \@1 \[ HEAD & \[ noun\\
							    POSS & $\neg$
							    none\]
						     \]
\] $\Longrightarrow$ \\
\hspace{4cm}
\@1 \[ SUBCAT \; $\neg$ \< \; \> \]
\avmr
\end{avm}
}
}

\section{Relative Clauses}
\label{SeRelCl}

Filler-gap dependencies are the contracts in which elements are
extracted from their positions (leaving gaps) and appear in other
positions (filler).  In Turkish, typical filler-gap construction is the
relative clauses. Two basic strategies exist for relative clauses which 
are called {\em wa} and {\em ga} by Hankamer and
Knecht~\cite{hankamerknecht76} which are realized respectively by {\tt
-(y)An} and {\tt -DHk-{\em Agr\/}} or {\tt -(A)cAk-{\em Agr\/}}
relative participles:

\myenumsentence{\label{rc}
\begin{itemize}
\item[i.] When the gap is the relative clause subject, or a
subconstituent of the relative clause subject, use the {\em wa} strategy.
\item[ii.] When there is no relative clause subject, use the
{\em wa} strategy.
\item[iii.] When th gap is not a part of relative clause subject, use 
the {\em ga} strategy.
\end{itemize}
}

(\ex{1}a--b) are examples of {\em wa}, (\ex{1}c--d) are examples of {\em ga} 
strategy.

\myeenumsentence{
\item
\shortex{6}{\uline{\hspace{0.5cm}}$_1$ & Adama & kalemi& 
					veren & \c{c}ocu\u{g}u$_1$ & g\"{o}rd\"{u}m.}
	   {& man-{\em Dat}& pencil-{\em Acc} &  give-{\em Rel} &child-{\em Gen} & see-{\em Past}-{\em 1Sg}}
	   {`I saw the child who gave man the pencil.'}
\item
\shortex{5}{\uline{\hspace{0.5cm}}$_1$ & yak{\i}n{\i}na & k\"{o}pr\"{u} & yap{\i}lan & ev$_1$}
	   {             &near-{\em 3SP-Dat}& bridge& 
			build-{\em Pass-Rel} & house}
	   {`The house$_1$ to which a bridge is built next
				\uline{\hspace{0.5cm}}$_1$'}
\item
\shortex{6}{\c{C}ocu\u{g}un & adama& \uline{\hspace{0.5cm}}$_1$ &
					verece\u{g}i& kalemi$_1$ & g\"{o}rd\"{u}m.}
	   {child-{\em Gen} & man-{\em Dat}& & give-{\em Rel-3Sg} & 
			pencil-{\em Acc} & see-{\em Past}-{\em 1Sg}}
	   {`I saw the pencil$_1$ that the child will 
				give \uline{\hspace{0.5cm}}$_1$ to the man.'}
\item
\shortex{6}{\c{C}ocu\u{g}un & kalemi& \uline{\hspace{0.5cm}}$_1$ &
					verdi\u{g}i& adam{\i}$_1$ & g\"{o}rd\"{u}m.}
	   {child-{\em Gen} & pencil-{\em Acc}& & give-{\em Rel-3Sg} & 
				man-{\em Acc} & see-{\em Past}-{\em 1Sg}}
	   {`I saw the man$_1$ to whom the child gave 
				\uline{\hspace{0.5cm}}$_1$ the pencil.'}
}

(\ex{-1}ii) introduces a special condition where the relative clause
has no subject. In Turkish there are two cases for clauses with no
subject~\cite{barker90}: impersonal passives and verbs with
incorporated subject. In these cases, the real agent of the verb does
not exist. The noun in the subject position incorporates to the verb.
In the example~(\ex{0}b), `k\"{o}pr\"{u}' is an incorporated subject. Similarly,
examples below show the relativization of an adjunct NP---a locative
adjunct in this case, with subject incorporation (\ex{1}a) and no
incorporation (\ex{1}b).

\myeenumsentence{
\item[a.1]\shortex{4}{ Kedi &  \c{c}ocu\u{g}un &
                   yata\u{g}{\i}nda &  uyudu.}
             { cat & child-{\em Gen} & bed-{\em 3SP-Loc} & 
				sleep-{\em Past-3Sg}}
             {'The cat slept in the child's bed.'}
\item[a.2]\shortex{4}{yata\u{g}{\i}nda & kedi & uyuyan & \c{c}ocuk}
             { bed-{\em 3SP-Loc} & cat & sleep-{\em Rel(wa)} & child}
             {'the child whose bed cat slept in'}
\item[b.1]\shortex{4}{Ay\c{s}e & \c{c}ocu\u{g}un &
              yata\u{g}{\i}nda & uyudu.}
             { & child-{\em Gen} & bed-{\em 3SP-Loc} & sleep-{\em Past-3Sg}}
             {'Ay\c{s}e slept in the child's bed.'}
\item[b.2]\shortex{4}{yata\u{g}-{\i}n-da & Ay\c{s}e'nin & uyu-du\u{g}-u & \c{c}ocuk}
             { bed-{\em 3SP-Loc} & Ay\c{s}e-{\em Gen} & 
			sleep-{\em Rel(ga)-3Sg} & child}
             {'the child whose bed Ay\c{s}e slept in'}
}

Gaps in relative clauses may involve dependencies which exist in nested
constituents (\ex{1}a--b). Infinitival verbs and possessives produce
gaps from missing noun phrase constituents and pass them to the upper
clause. This gap information is nonlocal to phrase, and projected until
a verb with the relative suffix is reached.  The clause headed by the
verb behaves as a modifier and gap is filled (i.e, structure-shared) by
the noun phrase at modified position.

\myeenumsentence{
\item
\shortex{6}{\uline{\hspace{0.5cm}}$_1$ & \c{C}ocu\u{g}u & kaybolan & kad{\i}n$_1$ &
		\c{c}ok & tela\c{s}land{\i}.}
	  { & child-{\em 3SP} & lost-{\em Rel} & woman & very &
				panic-{\em Past}-{\em 3Sg}}
	  {`The woman whose child is lost has panicked.'}
\item
\shortex{6}{Babama & \uline{\hspace{0.5cm}}$_1$ & be\u{g}endi\u{g}imi & 
				s\"{o}yledi\u{g}im& araba$_1$ & sat{\i}lm{\i}\c{s}.} 
	  {father-{\em 1SP}-{\em Dat} & & like-{\em Part-1Sg-Acc} & 
				tell-{\em Rel-1Sg} & car &
						    sell-{\em Pass-Past-3Sg}}
	  {`The car$_1$ that I told my father that I like 
				\uline{\hspace{0.5cm}}$_1$ is sold.'}
}

Such dependencies and information interaction with  the other phrases
over the local phrase boundary are called {\em non-local features} by
HPSG.  These features are ruled by a principle called {\bf Non-local
Feature Principle} \cite{PolSag94} which is adapted from {\bf Foot
Feature Principle} of GPSG.  For filler-gap dependencies, a nonlocal
feature called {\aF SLASH} is introduced.  In English more than one gap
is possible in a clause, so set type is used for {\aF SLASH}
attribute.  However in Turkish, a relative clause can contain only one
trace at any intermediate phrase. In case of nested relative clauses,
the gap is always filled and bound to a sister NP.  Therefore in our
design, {\aF SLASH} attribute can be {\aS null} or of type {\aS local}.
When the trace (empty category) is introduced, non-local feature {\aF
SLASH} is coindexed with the {\aF LOCAL} feature of the gapped argument
position.

\myenumsentence{
{\scriptsize
\begin{avm}
\[ PHON \; \< \; \> \\
   SYNSEM \; \[ LOCAL \; \@1 \\
		NONLOCAL\|INHERITED\|SLASH \; \@1
	     \]
\]
\end{avm}
}
}

\begin{figure}[ht]
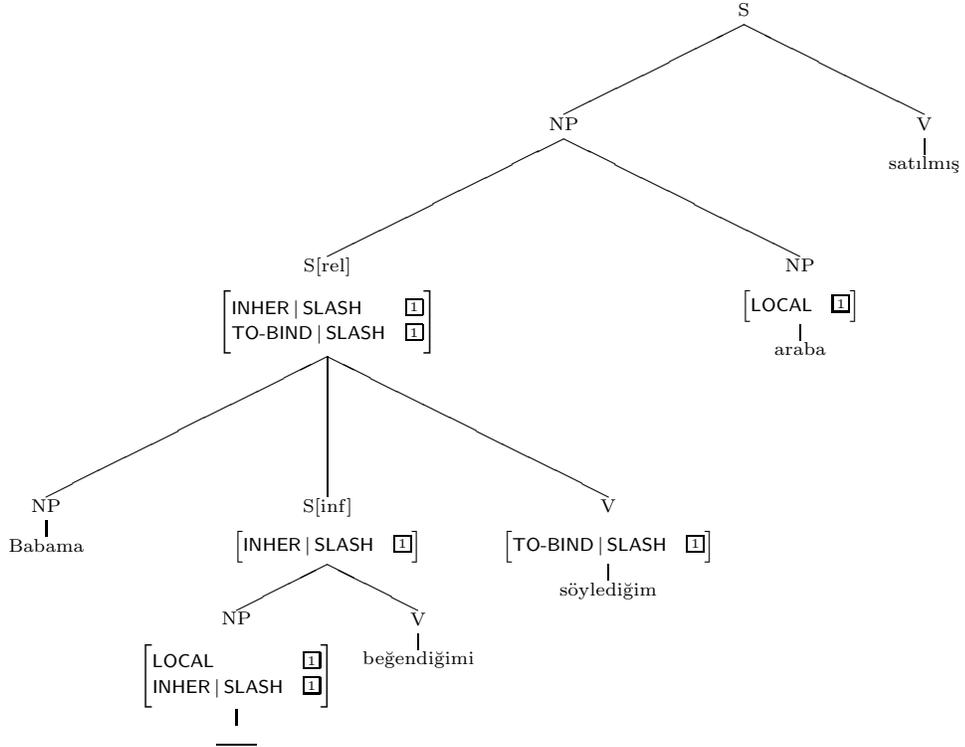
 
{\scriptsize
			\leaf{Babama}
			\branch{1}{ {\rm NP} }
				\leaf{ \uline{\hspace{0.5cm}} }
				\branch{1}{ NP \\
					    \begin{avm}
					    \[ LOCAL & \@1 \\
					       INHER\|SLASH & \@1
					    \]
					    \end{avm}
					   }
				\leaf{ be\u{g}endi\u{g}imi }
				\branch{1}{ V }
			\branch{2}{ S[inf] \\
				    \begin{avm}
				       \[ INHER\|SLASH & \@1 \]
				    \end{avm}
				  }
			\leaf{s\"{o}yledi\u{g}im}
			\branch{1}{V\\
				   \begin{avm}
				   \[ TO-BIND\|SLASH & \@1 \]
				   \end{avm}
				  }
		\branch{3}{ S[rel] \\
			    \begin{avm}
			    \[ INHER\|SLASH & \@1\\
			       TO-BIND\|SLASH & \@1
			    \]
			    \end{avm}
			  }
		\leaf{araba}
		\branch{1}{NP \\
			   \begin{avm}
			   \[ LOCAL & \@1 \]
			   \end{avm}
			   }
	\branch{2}{NP}
	\leaf{sat{\i}lm{\i}\c{s}}
	\branch{1}{V}
\branch{2}{S}
\tree
}

\caption{Projection of the SLASH feature}
\label{RelTree}

\end{figure}

Slash feature introduced by the trace is inherited to upper levels.
However, in some position, inheritance should be broken and filler
should be searched. In case of Turkish, this is the level where a
relative verb is the head of the phrase. HPSG marks these positions by
dividing {\aF NONLOCAL} feature into two attributes {\aF INHERITED} and {\aF
TO-BIND} of the   same structure. {\aF TO-BIND$\mid$SLASH} feature of
relative verbs are marked as {\aS local} (not {\aS null}) and
coindexed with the {\aF INHERITED$\mid$SLASH} feature. The resulting
phrase becomes a modifier of which the {\aF LOCAL} feature of the
modified phrase is also coindexed with the slash, so that the filler and
its trace are combined (Figure~\ref{RelTree}).

A lexical entry for relativized verbs is given in (\ex{1}). However,
some head features such as case, relativization, possession and
subcategorization are not supposed to be the same for the filler and
the trace.  To handle this, selected features {\aF CONT$\mid$INDEX},
{\aF HEAD$\mid$AGR} are passed to modified structure instead of
{\aF LOCAL} feature.

\myenumsentence{
{\scriptsize
\begin{avm}
\[ PHON \; \<{\rm `s\"{o}yledi\u{g}im'}\>  \; {\rm  \% tell-{\em Rel-1Sg}} \\
   SYNSEM \; \[ LOCAL\|CAT &  \[ HEAD & \[ {\aS obj-rel-verb} \\
				       AGR \; \[ PERSON & {\aS first} \\
					        NUMBER & {\aS sing} \]\\
				       MOD\|MODSYN\|LOCAL\|CONT\|INDEX \; \@1
				       \]\\
			     \]  \\
		NONLOCAL & \[ INHERITED\|SLASH & null \\
			      TO-BIND\|SLASH & \[ {\aS local} \\
						  CONT\|INDEX \; \@1 \]
			   \]
	    \] \\
\]
\end{avm}
}
}

The problem with the dropped pronouns also exists in relative clauses.
When trace is realized with empty category, efficiency and superfluous
ambiguity problems may arise. In our design, we used a simple technique
for raising slash feature, relying on two properties: First is valid
for most of the languages.  Every trace should be subcategorized by a
head.  Second is the free constituent order of Turkish. Since
complement order is relatively free in Turkish, we could assume that
the missing item is the last constituent.  Because any constituent may
be in the last (first in the surface but retrieved last) position in
the complement list. So we have introduced the following rule to
introduce trace instead of empty category:

\myenumsentence{\label{schema3}
{\scriptsize
\begin{avm}
\avml
\[ {\aF SYNSEM\|LOCAL\|CAT}  \; \[ HEAD & {\aS head} \\
  			        SUBJ & \@1 \\
				SUBCAT & \< \; \>
			     \]\\
   {\aF DTRS} \; \[ 
		   COMP-DTR & \@2 $\oplus$ \@1
		\]
\] \; $\longrightarrow$ \\
\hspace*{3cm}
\[
{\aF SYNSEM\|LOCAL\|CAT} & \[ HEAD & {\aS head} \\
			      SUBCAT & \@3 
			  \]\\
  {\aF DTRS\|COMP-DTRS} & \@2 
\] \;  , \\ \hspace*{3cm} {\tt select}\(\@1 \[ LOCAL \; \@4 \\
			      NONLOCAL\|INHERITED\|SLASH \; \@4 \],
			\@3, \< \; \>\) 
\avmr
\end{avm}
}
}

When the argument is the last item in the {\aF SUBCAT} list, it is
deleted, and the trace is introduced.  This solves the ambiguity in
subcategorized constituents. However yet another problem exists with the
traces which may occur in adverbs, which are not subcategorized for. In
English, prepositions subcategorize for an NP so that the trace could
be generated from subcategorization. However two case suffixes {\tt
-dA} (locative) and {\tt -(y)lA} (instrumental) in Turkish produce
nominal adjuncts that act as VP modifiers.  When they are missing,
since they have no surface form, it is impossible to introduce them by
the rule above. We introduce the trace as an empty category for these
two cases:

\myenumsentence{\label{nomadj}
{\scriptsize
\begin{avm}
\[ PHON \; \<\; \> \\
   SYNSEM \; \[ LOCAL \; \@1 \[ CAT\|HEAD \; \[ {\aS noun} \\
                                                CASE \; {\aS inst $\vee$
                                                                locative} \\
                                                MOD\|MODSYN\|LOCAL\|CAT \; 
					\[ HEAD \; {\aS verb}\\
				           SUBCAT \; \<\;\> 
					\]
			                          \] \]\\
                NONLOCAL\|INHERITED\|SLASH \; \@1 \]
\]
\end{avm}
}
}

The second problem is the definition of the constraints for the
use of {\em wa} and {\em ga} strategies described in~(\ref{rc}).
When the slash value is introduced in the subject position (subject daughter 
or one of its daughters is missing), {\em wa} strategy is used,
otherwise {\em ga} strategy is used.
These are expressed as:

{\bf Relative Clause Principle}
\myeenumsentence{
\item [a)] {\scriptsize \begin{avm}
\avml
\[ SYNSEM \;  \@1 \[ LOCAL\|CAT \[ HEAD \; {\aS subject-relative} \\
                                   SUBCAT \; \< \; \>
                                 \]
              \]
\] \; $\Longrightarrow$ \\
\hspace{1cm}
\@1 \[  LOCAL\|CAT\|SUBJ \; \[ NONLOCAL\|INHERITED\|SLASH \; \@2\] \\
        NONLOCAL\|TO-BIND \; \@2
\]
\avmr
\end{avm}}\toplabel{rcprinciple}
\item [b)]{\scriptsize \begin{avm}
\avml
\[ SYNSEM \;  \[ \@1 LOCAL\|CAT \[ HEAD \; {\aS object-relative} \\
                             SUBCAT \; \< \; \>
                                 \]
              \]
\] \; $\Longrightarrow$ \\
\hspace{1cm}
\@1 \[  LOCAL\|CAT\|SUBJ \; \[ NONLOCAL\|INHERITED\|SLASH \; \@2\] \\
        NONLOCAL\|TO-BIND \; $\neg$ \@2
\]
\avmr
\end{avm}}
}

In cases where the subject NP is
incorporated, the {\em wa} strategy
can be used even though the gap is not the subconstituent of the subject.
Such verbs are marked with a boolean
head feature called {\aF N-INCORP} standing for the noun incorporation.
The main constraint on this type of relative clause is that the type of
the noun in the subject position should be indefinite (or nonreferential)
because it is incorporated. The following additional constraint
solves the problem:

\myeenumsentence{\label{incrule}
\item[c)] {\scriptsize
\begin{avm}
\avml
\[ SYNSEM \;  \@1 \[ LOCAL\|CAT \[ HEAD \; \[ {\aS subject-relative} \\
                                              N-INCORP \; $+$
                                           \] \\
                             SUBCAT \; \< \; \>
                                 \]
              \]
\] \; $\Longrightarrow$ \\
\hspace{1cm}
\@1 \[  LOCAL\|CAT\|SUBJ \; \[ LOCAL\|CAT\|ADJUNCTS\|DEFINITE \; $-$ \]
\]
\avmr
\end{avm}
}
}

When a relative clause satisfying these constraints is saturated, and
its {\aF TO-BIND} feature is bound to the {\aF INHERITED} feature, it
acts as a noun modifier. The content index of the modified noun is
coindexed with the index of the gap to bind the semantic features of
the relative clause and the filler. The rest is handled by the adjunct
schema.  Nested relative clauses can modify the same noun so that
multiple gaps may be bound to the same filler (\ex{1}).

\myenumsentence{
\shortex{7}{annemin&\uline{\hspace{1cm}}$_1$&yapt{\i}\u{g}{\i}&
			\uline{\hspace{1cm}}$_1$&\c{c}ok&sevdi\u{g}im&kurabiyeler$_1$}
	   {mother-POSS & &cook-{\em Rel}& & much& like-{\em Rel} & cookie-{\em Plu}}
	   {`the cookies that my mother cooked, that I like'}
}

\section{Substantive Predicates}

As mentioned in Chapter~\ref{ChTurkish}, Turkish sentences may have
verbal, existential or substantive heads. Substantive predicates are
formed by substantive heads with auxiliary ({\tt -DH-}{\em Agr}, {\tt
-mH\c{s}-}{\em Agr}) or copula suffixes.  Syntactically, substantive heads
subcategorize for an NP which have the same semantic index. In other
words, substantive head and the subcategorized NP describe the same
nominal object (\ex{1}a--b). Copula and agreement suffixes marks the
agreement of the categorized NP.

\myeenumsentence{
\item \shortex{3}{Ben & \c{c}ok & hastay{\i}m.}
		 {I & much & ill-{\em Cop(1Sg)}}
		 {`I'm too sick.'}
\item \shortex{3}{B\"{u}t\"{u}n & kad{\i}nlar & \c{c}i\c{c}ektir.}
		 {every & woman-{\em Plu} & flower-{\em Cop(3Sg)}}
		 {`Every woman is a flower.'}
}

Another design consideration is to distinguish predicative NP's from the
others. First, this is necessary to determine whether a saturated NP forms
a sentence or not. Second, the same problem with the possessive NP exists
for substantive predicates. A saturated predicative NP should not be
further modified by another adjective. For these two reasons, we have added
a boolean type head feature for substantial types called {\aF PREDICATIVE}.
The following is a sample entry for a predicative noun ``insan{\i}m''.

\myenumsentence{
{\scriptsize
\begin{avm}
\[ PHON \; {\rm ``insan{\i}m''} \; \; {\rm \% Human-{\em Cop(1Sg)},`I am a human'} \\
   SYNSEM\|LOCAL \;  \[ CAT \; \[  HEAD \; \[ {\aS noun}\\
					      PREDICATIVE \; $+$ 
					    \]\\
				   SUBCAT \; \< NP$_{\it 1sg}$$_{\@1}$ \>
				\] \\
			CONT\|INDEX \; \@1
		     \]
\]
\end{avm}
}
}

After this feature is defined, the following constraint is added to 
the constraint~(\ref{AdjPoss}) for the adjunct rule:

\myenumsentence{\label{AdjCop}
{\scriptsize
\begin{avm}
\avml
\[ SYNSEM\|LOCAL\|CAT\|HEAD\|MOD\|LOCAL\|CAT \; \@1 \[ HEAD & \[ noun\\
						    PREDICATIVE & $+$
						              \]
						     \]
\] $\Longrightarrow$ \\
\hspace{4cm}
\@1 \[ SUBCAT \; $\neg$ \< \; \> \]
\avmr
\end{avm}
}
}

In the implementation, substantive predicates are realized by a lexical
rule which maps lexical entry for a non-predicative substantive word to
substantive predicate by an auxiliary or copula suffix:

\myenumsentence{
{\scriptsize
\begin{avm}
\avml
\[ PHON \; \@1 \\
   SYNSEM\|LOCAL \; \[ CAT \; \[ HEAD \; \[ {\aS subst} \\
					    PREDICATIVE \; $-$
					 \]\\
				 SUBCAT \; \@2 
			      \] \\
		       CONT\|INDEX \; \@3
		     \]
\] \; \; $\longmapsto$ \\
\[ PHON \;  {\tt apply-cop(}\@1,\@5{\tt )} \\
   SYNSEM\|LOCAL \; \[ CAT \; \[ HEAD \; \[ PREDICATIVE \; $+$ \] \\
				 SUBCAT \; NP[AGR \@5,INDEX \@3] $\oplus$
				 			\@2
			      \] 
		    \]
\]
\avmr
\end{avm}
}
}

Where {\tt apply-cop} is a general predicate applying the copula suffix
corresponding to agreement feature marked with the second argument to
the first argument and returning the resulting string.

\chapter{ALE IMPLEMENTATION}
\label{ChImp}

In the implementation, we have used ALE (Attribute Logic Engine)\cite{ALE}.
ALE is an integrated system of definite clause logic programming and phrase
structure parsing. All operations and declarations in ALE use {\em typed
feature structures} as terms. ALE is designed and suited for
implementations of unification-based language formalisms.

ALE is a strongly typed language. Every structure must have a declared
type. Types are defined by an inheritance structure and subtype
relation.  Basic representation scheme used is the typed feature
structures. Types are assigned to appropriate feature-value pairs. Type
structure of ALE is very similar to the HPSG including properties like
inheritance, nesting, and well-typedness. However it is most restricted
in favor of efficiency and implementation considerations.

ALE allows definition of general constraints on types. One can put
restrictions on the feature structures of a particular type. Another
feature of ALE is the definite clauses in which all functionality of PROLOG
definite clauses is provided with feature structure unification instead of
simple term unification. Also complex descriptions can be simplified by
the use of macros.

One of the most distinct features of ALE from other tools like TFS and
CUF \cite{Manand} is the support for phrase structure grammars. ALE
provides phrase structure rules to be coded like Definite Clause
Grammars of PROLOG. It has a built-in bottom-up chart
parser in addition to feature structure unification. DCG's are top-down
and depth-first.  However ALE parser works in a combined manner
asserting edges to chart right to left while applying rules left to
right. ALE also allows lexical rules for dealing with lexical
redundancy. Lexical rules can be defined for inflectional or
derivational morphology as well as zero derivations like nominalization
of adjectives.  Morphological constraints (suffixation, affixation
etc.) can be controlled by some built in mechanisms or PROLOG
predicates. It also allows empty categories to be integrated into
grammar.

\section{Grammar Rules and Principles}

We have four phrase structure rules, each corresponding to a schema
that we have introduced in the preceding chapter. First two
(\ref{schema1a}, \ref{schema1b}) handle the complement retrieval and
subcategorization. Applying both rules cause superfluous parses due to
the application order. The third is the adjunct schema (\ref{schema2})
which handles the adjunct-head relation. And the fourth is the rule
introducing the slash (\ref{schema3}).

Rules are coded by standard Immediate Dominance and Linear Precedence
notation. The application of rules are governed and constrained by a set
of ALE definite clauses. These include two simple clauses modifying the 
{\aF DAUGHTER} and {\aF PHON} features of the mother phrase. The others are
constraints and basic principles. 

{\tt head-feature-principle} applies the {\bf Head Feature Principle} of
HPSG; head feature of the mother is structure shared with the head feature
of the head daughter. {\tt selectlast} and {\tt selectfirst} implements the 
{\bf Subcategorization Principle} of HPSG. The surface form of the combined 
{\aF SUBCAT} structure of list and sets with optional arguments is
generated, one item is selected, and rest is returned as the {\aF SUBCAT}
feature of the mother phrase. Figure~\ref{PrincSrc} shows the source for
the simplified versions of these two principles. 

\begin{figure}[hbt]

{\small \alltt \tt
\%
\%  head-feature-principle(MothSign,HeadSign)
\%
head-feature-principle(synsem:local:cat:head:X,synsem:local:cat:head:X) 
\hspace{3cm}if true.

\% 
\%  subject-retrieval-rule 
\%
subcat_retr1 rule
(Mother, synsem:local:cat:subcat:SubcatRest) ===>
\hspace{2cm}cat>  (Complement,synsem:CompSyn),
\hspace{2cm}cat>  (Head,synsem:local:cat:subcat:Subcat),
\hspace{2cm}goal> (head-feature-principle(Mother,Head),
\hspace{2cm}       seleclast(CompSyn,Subcat,SubcatRest)).

}

\label{PrincSrc}
\caption{Sample Source for Head Feature and Subcategorization Principles}
\end{figure}

Linear precedence and word order constraints of the system are realized
by the definite clauses {\tt removeop} and {\tt surface}. The mixed
subcat structure (cf. \ref{SubType}) consisting of nested sets and lists
with optional arguments is converted into permutations of the surface
form by these clauses. {\tt removeop} produces omitted and existed
permutations of optional arguments, and {\tt surface} produces the
flattened lists from the resulting structure.

{\tt nonlocal-principle} combines the {\aF NONLOCAL}  features of the
daughters and modifies the mother. It inherits the gap
information in one of the daughters from {\aF INHERITED} feature to the
mother phrase. If {\aF TO-BIND} feature of the head is not {\aS
null} it binds the gap and applies the constraint for the relative
clauses.  {\tt adjunct-principle} puts the constraints in
schema~(\ref{schema2}) and two constraints introduced in
(\ref{AdjPoss}) and (\ref{AdjCop}). Grammar rules are coded in file {\tt
T.rule} and principles are coded in {\tt T.clause} in
Appendix~\ref{Source}.

\section{Lexicon and Lexical Rules}

Lexical redundancy becomes a crucially important problem in agglutinative
languages where a large number of derivations and inflections of a root
word exists. It is almost impossible to store all derivations and
inflections of Turkish words into a lexical database. Therefore some sort
of morphological analysis and application of lexical rules are essential.
Also application of a lexical inheritance hierarchy could be used to deal
with redundancy.

ALE does not suggest a standart mechanism for implementing lexical type
hierarchies. Two methods seem to be applicable: use of macros, and type
hierarchies with general constraints. Macro definitions in ALE allow for
variable substitution and ---not recursive--- nesting. Each node in the
lexical hierarchy can be defined by a macro which contains the calls for
the parent macros. At the lexical level, items are defined by one or
more macros. Macros are expanded at run time, and ALE operates on the
expanded descriptions. Second solution is to construct a type hieararchy 
under the type {\aS word} which is the type of the lexical elements,
and put general constraints on these types. However, in this approach,
it is impossible to assign a lexical item to several nodes in the type
hierarchy since each item can belong to only one type. Besides, ALE has
a very restricted type mechanism so multiple inheritence is very
limited.  On the other hand, constraints are evaluted at compile time
which is efficient compared to macros.

\begin{figure}[hbt]
\centerline{\psfig{file=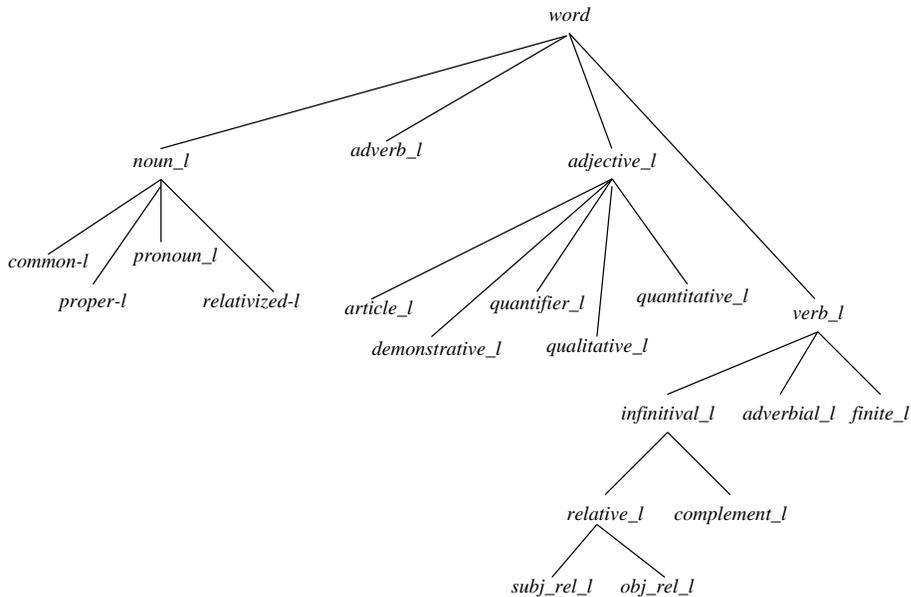}}
\caption{Lexical hierarchy}
\label{LexHier}
\end{figure}

In implementation, we have used a lexical type hierarchy. We defined a
lexical type tree under the type {\aS word} for the lexical entries.
Each common class of lexical entries are defined as a node in the tree
(Figure~\ref{LexHier}). We have used ``{\tt \_l}'' as the last two
characters in the names of these lexical types to distinguish them from
{\aS head} types.

We also defined constraints on these types (see source files {\tt
T.type} and {\tt T.cons} in Appendix~\ref{Source}). After all these
declarations, defining a lexical entry as one of the types above will
apply all constraints associated with the supertypes along the path to
{\aS word}.

ALE supports lexical rules. At the feature structure level, unification
and use of user defined ALE definite clauses are provided. At the surface
level, the user can define string operations by
concatenation and user defined PROLOG predicates. Use of PROLOG
at string level enables user to implement any kind of complex morphological 
phenomena like affixation, vowel harmony, drops etc. Also complex structure
changes in signs can be coded by the help of the ALE's definite clauses.

However, ALE lexical rules are inefficient for an agglutinative language
like Turkish. First, ALE applies all rules at compile time and asserts all
generated combinations as PROLOG predicates which consumes too much
memory and increases compilation time considerably. Second, it
implements all kinds of string-to-string mappings to handle a wide range of
languages. However this makes it very slow for a large number of lexical
rules and lexical entries. Both inefficiencies make the developement
very difficult. So we have tried to edit some portion of ALE source
code and made some changes which will apply lexical rules at runtime
and apply lexical mappings only to closed class of words. This
increased the efficiency to a reasonable level.

We have coded lexical rules for nominal casess, possesive suffixes
(which mark agreement in case of sentential complements and relative
clauses), and noun relativizer suffix. Zero derivations like
adjective-to-noun promotion and production of non-referential object
case of verbs are implemented by lexical rules. These lexical rules are
defined in the file {\tt T.lex\_rule} (Appendix~\ref{Source}). For
reusability considerations, similar feature structure transformations
are grouped into ALE definite clauses which are defined in the file
{\tt T.clause}. For example, rules for application of case suffixes are
implemented in clause {\tt apply-case/3} and lexical rules for all
cases call this clause with case passed as argument.

In the appendix~\ref{Source}, some part of grammar code is given. It
requires Quintus or SICStus Prolog. Full system can be obtained via
anonymous ftp from {\tt ftp.lcsl.metu.edu.tr} in path {\tt
/pub/theses/sehitoglu-ms-96.tar.gz}.


\chapter{CONCLUSION}

In recent years, computational studies on Turkish have proliferated.
These studies are important in two respects: First, building foundations of
linguistic description of Turkish within the light of the contemporary
linguistic theories. Second, providing basic tools for natural
language processing which has applications in computer science ranging from
simple text processing utilities to translation and learning tools.

HPSG is the synthesis of the some of the recent linguistics theories. 
It is a developing theory, and new principles and approaches are being
introduced for expanding the universal coverage. Being one of the most
powerful among the other unification based and phrase structure formalisms,
it models the language in informational perspective and describes the
linguistic events by a set of universal principles and metarules. It is a
general theory trying to be as flexible as possible to cover
principles of all natural languages.

In this study, we have worked on a computational sign-based model of
Turkish, following and adapting the HPSG framework. HPSG uses feature
structures to describe linguistic phenomena. This allows the grammar
designers to concentrate on the {\em constraints} imposed by a 
particular language on a well-defined set of linguistic features. This is
in contrast to earlier context-free grammar rules, where language-specific
rules do not allow generalizations. Postulating principles and writing
constraints on these principles show how different languages model the same
phenomenon in different ways. To this end, we have analyzed and implemented
the general principles such as subcategorization, adjunct-head selection, 
relative extraction. We have also studied the principles such as word-order
variation, pronoun and complement drops and unbounded dependencies, which
are particularly important for Turkish.

For the time being, the parser has not been combined with a 
lexical analyzer and tested on a real corpus. Since most of the syntactic
information is coded in the lexicon, an intelligent mechanism for
gathering all lexical entries for Turkish should be employed. HPSG
proposes solutions like lexical inheritance hierarchy, and lexical rules.
Turkish is an agglutinative language and has many
syntactically effective and productive suffixes. This means that there is 
more interactions between morphology and syntax, compared to a language such
as English.

The computational tool we have used for HPSG, ALE, supports
lexical rules with morphological analysis. However it is inefficient for  
running a grammar with large lexicon and all lexical rules.  As the
main problem about lexicon, ALE does all lexical processing at compile
time and generates all possible results of lexical rules statically,
which is not suitable for agglutinative languages.  Also, since
morphological rules are defined in PROLOG, they are very inefficient.
We made some changes to apply lexical rules at run time and
make morphology a little bit faster. However, for an efficient lexical 
analysis, use of an external lexicon and morphological analyser is
necessary. Necessary interface routines can be coded into PROLOG source
code of ALE as the changes we have already done.

Another approach could be integration of syntactic and 
morphological analysis. This is achived by encoding
morphological analysis combined with syntactic rules in the style of HPSG 
principles. This is also desirable from the linguistic point of view;  
morphological and syntactic phrasing can affect each other in a principled 
way.\footnote{for an integrated analysis of morphology and syntax 
cf.~\cite{BozGoc95}}

ALE has some drawbacks as well as powerful features. The strong typing
cause description domain to be restricted. Type hierarchies requiring
latice-like multiple inheritances cannot be coded efficiently. Also it
disallows the usage of atomic types without type declaration. It has a 
unification based description language and type inferencing mechanism 
provided with definite clauses with all functionality of PROLOG. However 
ALE lacks some sort of {\em overwrite} operation especially in lexical 
rules which are procedural in nature. Overwrite operation changes some 
part of a feature rather than unifying it. Such an operator may ease the
formulations and descriptions of lexical rules.

We have limited semantic analysis of signs to minimum. Since HPSG is a
complete linguistic theory for both syntax and semantics, for a complete
HPSG analysis of Turkish, semantic principles and model should be
analyzed.

\nocite{*}

\appendix
\chapter{PARSER SOURCE}
\label{Source}

{\scriptsize \alltt \tt

\section{Type Definitions}
%
%

bot sub [bool,sign,null_synsem,cat,head,case,null_agr,per,num,posses,
                list,char,set,tense,aux_tense,nonloc,null_adjstr,null_mod,
                list_or_set_subcat,subcat_or_ne_set,subcat_or_ne_list,
                psoa_arg,qfpsoa,sem_det,sem_obj,null_local,arg_type,
                subcat,conx,null_local,nonlocal,daughters].

sign sub [lexical,phrase]
   intro [phon:list_string,
          synsem:synsem,
          qstore:set_quant,
          qretr:list_quant].
  lexical sub [word].
      word sub [noun_l,adj_l,adv_l,verb_l].
        noun_l sub [common_l,proper_l,pronoun_l,relativized_l].
           common_l sub [].
           proper_l sub [].
           pronoun_l sub [].
           relativized_l sub [].
        adj_l sub [quantif_l,article_l,demonstra_l,quantitive_l,
                                                qualitative_l].
           quantif_l sub [].
           article_l sub [].
           demonstra_l sub [].
           quantitive_l sub [].
           qualitative_l sub [].
        adv_l sub [].
        verb_l sub [finite_l,sadv_l,inf_l].
           finite_l sub [].
           sadv_l sub [].
           inf_l sub [relcl_l,complement_l].
             relcl_l sub [subj_rel_l,obj_rel_l].
               subj_rel_l sub [].
               obj_rel_l sub [].
             complement_l sub [].
  phrase sub []
     intro [ dtrs: daughters ].

null_synsem sub [null,synsem].
  null sub [].
  synsem sub []
     intro [local:local,
              nonlocal:nonlocal].

null_local sub [null,local].
local sub []
   intro [cat:cat,cont:sem_obj,conx:conx].

conx sub [].

nonlocal sub []
   intro [inherited:nonloc,tobind:nonloc].

nonloc sub []
   intro [slash:null_local].

daughters sub [hd_subj_st,hd_st]
    intro [hd_dtr: sign,
           comp_dtrs: list_sign,
           spec_dtrs: list_sign].
 hd_subj_st sub []
    intro [subj_dtr:sign].
 hd_st sub [].

bool sub [plus,minus].
  plus sub [].
  minus sub [].

cat sub []
   intro [head:head,
          subj:null_synsem,
          adjuncts:null_adjstr,
          subcat:list_or_set_subcat].

head sub [subst,prep,adverb,verb]
    intro [mod:null_mod].
  subst sub [adj,noun]
    intro [pred:bool].
  adj sub [determiner,quantitative_adj,qualitative_adj,questional_adj]
      intro [countable:bool,gradable:bool].
    determiner                sub [article,demonstrative_adj,quantifier].
      article                sub [].
      demonstrative_adj        sub [].
      quantifier        sub [].
    quantitative_adj        sub [number,distributive_adj,grouping_adj].
      number                 sub [cardinal,fractional].
        cardinal                sub [].
        fractional        sub [].
      distributive_adj        sub [].
      grouping_adj        sub [].
    qualitative_adj        sub [].
    questional_adj        sub [].

  noun sub [common,proper_noun,pronoun]
     intro [case:case,
            agr:agr,
            n_ind:agr,
            rel:bool,
            poss:posses].
    common sub [].
    proper_noun sub [].
    pronoun sub [personal_pr,demonstrative_pr,reflexive_pr,indefinite_pr
                        ,quantificational_pr,questional_pr].
        personal_pr                 sub [].        
        demonstrative_pr         sub [].        
        reflexive_pr                 sub [].        
        indefinite_pr                 sub [].        
        quantificational_pr         sub [].        
        questional_pr                 sub [].        
  prep sub [].
  adverb sub [dir_adv,dir_adv,temp_adv,manr_adv,quant_adv,sent_adv,quest_adv].
    dir_adv sub []
       intro  [dir:direction].
    temp_adv sub [t_unit_adv,pot_adv,t_per_adv].
      t_unit_adv sub [].
      pot_adv sub [].
      t_per_adv sub [dayt,dayw,seas].
        dayt sub [].
        dayw sub [].
        seas sub [].
    manr_adv sub [qual_adv,rep_adv].
      qual_adv sub [].
      rep_adv sub [].
    quant_adv sub [approx,comp,superl,excess]. 
      approx sub [].
      comp sub [].
      superl sub [].
      excess sub [].
    sent_adv sub [].
    quest_adv sub [].

  verb sub [infinitival,adverbial,finite]
      intro [tense:tense,neg:bool,vagr:null_agr,n_inc:bool].
    infinitival sub [relative,complementary]
        intro [vcase:case].
      relative sub [subj_rel,obj_rel].
        subj_rel sub [].
        obj_rel sub [].
      complementary sub [mak,{\i}\c{s},complemented].
        mak  sub [].
        {\i}\c{s} sub [].
        complemented sub [].
    adverbial sub [].
    finite sub []
        intro [aux_tense:aux_tense].

case sub [nom,obj,gen,loc,direction,ins].
  nom sub [].
  obj sub [].
  gen sub [].
  loc sub [].
  direction sub [dat,abl].
    dat sub [].
    abl sub [].
  ins sub [].

null_agr sub [null,agr].
  agr sub []
    intro [per:per,
           num:num].
        
per sub [first,second,third].
  first sub [].
  second sub [].
  third sub [].

num sub [sing,plur].
  sing sub [].
  plur sub [].

posses sub [none,poss].
  none sub [].
  poss sub []
     intro [by:agr].

null_adjstr sub [null,adjstr].
  adjstr sub []
     intro [qtfd:bool,dmstrtd:bool,rltvzd:bool,rltclsd:bool,qntfcd:bool,
                                qltfd:bool,non_ref:bool].
                
null_mod sub [null,mod].
  mod sub []
     intro [modsyn:synsem,modadj:null_adjstr].

tense sub [base,future,contin,pres,past,rep_past].
  base sub [].
  future sub [].
  contin sub [].
  pres sub [].
  past sub [].
  rep_past sub [].

aux_tense sub [null,hikaye,rivayet,condition].
  hikaye sub [].
  rivayet sub [].
  condition sub [].

psoa_arg sub []
    intro [argname: string,arg: arg_type].

arg_type sub [agr,psoa].

qfpsoa sub [property, relation]
    intro [name:string].
  property sub []
        intro [inst:agr].
  relation sub []
        intro [args:list_psoa_arg].

sem_det sub [forall,exists,the].
  forall sub [].
  exists sub [].
  the sub [].
sem_obj sub [nom_obj, psoa, quant].
  nom_obj sub [npro, pron]
          intro [index:agr,
                 restr:set_psoa].
    npro sub [].
    pron sub [ana, ppro].
      ana sub [recp, refl].
        recp sub [].
        refl sub [].
      ppro sub [].
  quant sub []
        intro [det:sem_det,
               restind:npro].
  psoa sub []
      intro [quants:list_quant,nucleus:qfpsoa].

subcat sub [optionalcat,subcat_type].
  optionalcat sub [opt,obl]
    intro [s_arg:subcat_type].
    opt sub [].
    obl sub [].
  
subcat_type sub [char,synsem,sign].

list_or_set_subcat sub [set_subcat,list_subcat,list_xxx].
subcat_or_ne_set sub [subcat,ne_set_subcat].
subcat_or_ne_list sub [subcat,ne_list_subcat].

list sub [e_list,ne_list,list_cat,string,list_string,list_sign,
          list_quant,list_xxx,list_psoa_arg].
  e_list sub [].
  ne_list sub [ne_list_cat,ne_string,ne_list_string,
                ne_list_xxx,ne_list_sign,ne_list_quant,ne_list_psoa_arg]
     intro [hd:bot,
            tl:list].
  ne_list_xxx sub [ne_list_subcat,ne_list_synsem].
  list_cat sub [e_list,ne_list_cat].
     ne_list_cat sub []
        intro [hd:cat,
               tl:list_cat].
  string sub [e_list,ne_string].
     ne_string sub []
        intro [hd:char,
               tl:string].
  list_xxx sub [list_subcat,list_synsem,ne_list_xxx].
  list_subcat sub [e_list,ne_list_subcat].
     ne_list_subcat sub []
         intro [hd: subcat_or_ne_set,
                tl: list_subcat].
  list_synsem sub [e_list,ne_list_synsem].
     ne_list_synsem sub []
        intro [hd:synsem,
               tl:list_synsem].
  list_string sub [e_list,ne_list_string].
     ne_list_string sub []
        intro [hd:string,
               tl:list_string].
  list_sign sub [e_list,ne_list_sign].
     ne_list_sign sub []
        intro [hd:sign,
               tl:list_sign].
  list_quant sub [e_list,ne_list_quant].
     ne_list_quant sub []
        intro [hd:quant,
               tl:list_quant].
  list_psoa_arg sub [e_list,ne_list_psoa_arg].
     ne_list_psoa_arg sub []
        intro [hd:psoa_arg,
               tl:list_psoa_arg].

set sub [e_list,ne_set,set_char,set_subcat,set_psoa,set_quant].
  ne_set sub [ne_set_char,ne_set_subcat,ne_set_psoa,ne_set_quant]
     intro [el:bot,
            els:set].
  set_char sub [e_list,ne_set_char].
     ne_set_char sub []
        intro [el:char,
               els:set_char].
  set_subcat sub [e_list,ne_set_subcat].
     ne_set_subcat sub []
         intro [el: subcat_or_ne_list,
                els: set_subcat].
  set_psoa sub [e_list,ne_set_psoa].
     ne_set_psoa sub []
         intro [el: psoa,
                els: set_psoa].
  set_quant sub [e_list,ne_set_quant].
     ne_set_quant sub []
         intro [el:quant,
                els: set_quant].

char sub [a,b,c,\c{c},d,e,f,g,\u{g},h,{\i},i,j,k,l,m,n,o,\"{o},p,q,r,s,\c{s},t,u,\"{u},v,w,x,y,z,ø].
   a sub [].
   ........
   .......

\section{Phrase Structure Rules}
%
%

subcat_retr1 rule
(Mother,phrase,phon:PhonMot,
               synsem:local:cat:(subcat:SubMot,adjuncts:Adjs,subj:Subj),
               dtrs:DtrsMot)
===>
cat> (Arg,phon:PhonArg,synsem:SynArg),
cat> (Head,phon:PhonHead,synsem:local:cat:(subcat:SubHead,adjuncts:Adjs,
                                           subj:Subj)),
goal> (append(PhonArg,PhonHead,PhonMot),
       head_feature_principle(Mother,Head),
       sselectlast(SynArg,SubHead,SubMot),
       combine_semantics(Head,Arg,Mother),
       append_comp(DtrsMot,Head,Arg),
       nonlocal_principle(Arg,Head,Mother)).

subcat_retr2 rule
(Mother,phrase,phon:PhonMot,
               synsem:local:cat:(subcat:SubMot,adjuncts:Adjs,subj:Subj),
               dtrs:DtrsMot)
===>
cat> (Head,phon:PhonHead,synsem:local:cat:(head:verb,
                                        subcat:SubHead,adjuncts:Adjs,
                                        subj:Subj)),
cat> (Arg,phon:PhonArg,synsem:SynArg),
goal> (append(PhonArg,PhonHead,PhonMot),
       head_feature_principle(Mother,Head),
       sselectlast(SynArg,SubHead,SubMot),
       combine_semantics(Head,Arg,Mother),
       append_comp(DtrsMot,Head,Arg),
       nonlocal_principle(Arg,Head,Mother)).

adj_head rule
(Mother,phrase,phon:PhonMot,dtrs:DtrsMot)
===>
cat> (Adjunct,phon:PhonAdj),
cat> (Head,phon:PhonHead),
goal> (append(PhonAdj,PhonHead,PhonMot),
       combine_semantics(Head,Adjunct,Mother),
       head_feature_principle(Mother,Head),
       adjunct_principle(Mother,Adjunct,Head),
       append_spec(DtrsMot,Head,Adjunct),
       nonlocal_principle(Adjunct,Head,Mother)).

slash rule
(Mother,phrase,phon:PhonMot,
               synsem:(local:cat:(subcat:SubMot,subj:Subj),
                       nonlocal:(inherited:slash:Local,
                                 tobind:slash:HT)),
               dtrs:DtrsMot)
===>
cat>(Head,phon:PhonHead,synsem:(local:cat:(head:(noun;infinitival),
                                           subcat:SubHead,
                                           subj:Subj),
                                nonlocal:(inherited:slash:null,
                                          tobind:slash:HT))),
goal> (append((PhonSl,[e_list]),PhonHead,PhonMot),
       head_feature_principle(Mother,Head),
       nonlocal_principle(synsem:Slsynsem,Head,Mother),
       sselectlast((Slsynsem,local:(Local,
                                cat:head:(agr:per:third,n_ind:SlInd),
                                cont:index:SlInd),
                   nonlocal:(inherited:slash:Local,
                             tobind:slash:null)),SubHead,(SubMot,e_list)),
       append_comp(DtrsMot,Head,(Slash,phon:PhonSl,qretr:e_list,qstore:e_list,
                                    synsem:Slsynsem)),
       combine_semantics(Head,synsem:local:cont:(index:SlInd,restr:e_list)
                                                                ,Mother)).

\section{Constraints and Macros}

%
%

determiner cons (gradable: minus).
article    cons (countable: plus).
quantitative_adj  cons (gradable: minus,countable: plus).
word cons (qretr:e_list,synsem:nonlocal:inherited:slash:null).
subj_rel cons (tense:base,vagr:null).
mak cons (tense:base).
{\i}\c{s} cons (tense:base,vagr:agr).
complemented cons (tense:(future;past),vagr:agr).
noun_l cons (synsem:local:cat:head:(n_ind:I,[agr] == [n_ind])).
common_l cons (synsem:local:(cat:(head:n_ind:I,
                                  adjuncts:(qtfd:minus,
                                     dmstrtd:minus,
                                     rltvzd:minus,
                                     rltclsd:minus,
                                     qntfcd:minus,
                                     qltfd:minus,
                                     non_ref:plus)),
                              cont:index:I)).

pronoun_l cons (synsem:local:(cat:(head:n_ind:I,
                                  adjuncts:(
                                     qtfd:minus,
                                     dmstrtd:minus,
                                     rltvzd:minus,
                                     rltclsd:minus,
                                     qntfcd:minus,
                                     qltfd:minus,
                                     non_ref:minus)),
                              cont:index:I)).

quantif_l cons 
  (synsem:local:cat:head:(quantifier,
                            mod:(modsyn:(local:cat:(head:(common),
                                                   adjuncts:(qtfd:minus,
                                                     dmstrtd:minus,
                                                     rltvzd:minus,
                                                rltclsd:A,
                                                qntfcd:B,
                                                qltfd:C))),
                                modadj:(qtfd:plus,
                                     dmstrtd:minus,
                                     rltvzd:minus,
                                     rltclsd:A,
                                     qntfcd:B,
                                     qltfd:C,non_ref:minus)))).

demonstra_l cons 
  (synsem:local:cat:head:(demonstrative_adj,
                            mod:(modsyn:(local:cat:(head:(common),
                                                   adjuncts:(qtfd:minus,
                                                     dmstrtd:minus,
                                                     rltvzd:minus,
                                                rltclsd:A,
                                                qntfcd:B,
                                                qltfd:C))),
                                modadj:(qtfd:minus,
                                     dmstrtd:plus,
                                     rltvzd:minus,
                                     rltclsd:A,
                                     qntfcd:B,
                                     qltfd:C,
                                     non_ref:minus)))).

qualitative_l cons 
  (synsem:local:cat:head:(qualitative_adj,
                            mod:(modsyn:(local:cat:(head:(common),
                                                   adjuncts:(qtfd:A,
                                                     dmstrtd:minus,
                                                     rltvzd:minus,
                                                rltclsd:minus,
                                                qntfcd:B,
                                                non_ref:C))),
                                modadj:(qtfd:A,
                                     dmstrtd:minus,
                                     rltvzd:minus,
                                     rltclsd:minus,
                                     qntfcd:B,
                                     qltfd:plus,
                                     non_ref:C)))).

relativized_l cons 
  (synsem:local:cat:head:( mod:(modsyn:(local:cat:(head:(common),
                                                   adjuncts:(qtfd:A,
                                                     dmstrtd:B,
                                                rltclsd:minus,
                                                qntfcd:D,
                                                qltfd:E)
                                        )),
                                modadj:(qtfd:A,
                                     dmstrtd:B,
                                     rltvzd:plus,
                                     rltclsd:minus,
                                     qntfcd:D,
                                     qltfd:E,
                                     non_ref:minus)))).

subj_rel_l cons 
  (synsem:(local:(cat:(head:(subj_rel,
                                mod:(modsyn:(local:(cat:(head:(common,
                                                            n_ind:NInd),
                                                   adjuncts:(
                                                qtfd:A,
                                                     dmstrtd:B,
                                                rltvzd:minus,
                                                qntfcd:D,
                                                qltfd:E)
                                                ),
                                                cont:(Cont,index:Ind))),
                                modadj:(qtfd:A,
                                     dmstrtd:B,
                                     rltvzd:minus,
                                     rltclsd:plus,
                                     qntfcd:D,
                                     qltfd:E,
                                     non_ref:minus)))),
                  cont:_),
           nonlocal:tobind:slash:(cat:head:(common,n_ind:NInd
                                                )
                                       ))).
obj_rel_l cons 
  (synsem:(local:(cat:(head:(obj_rel,
                             mod:(modsyn:(local:(cat:(head:(common,
                                                            n_ind:NInd),
                                                   adjuncts:(
                                                qtfd:A,
                                                     dmstrtd:B,
                                                rltvzd:minus,
                                                qntfcd:D,
                                                qltfd:E)),
                                                cont:Cont)),
                                modadj:(qtfd:A,
                                     dmstrtd:B,
                                     rltvzd:minus,
                                     rltclsd:plus,
                                     qntfcd:D,
                                     qltfd:E,
                                     non_ref:minus)))),
                   cont:index:NInd),
           nonlocal:tobind:slash:(cat:head:(common,
                                                n_ind:NInd),
                                  cont:Cont))).                

finite_l cons
  (synsem:(local:(cat:head:finite),
           nonlocal:tobind:slash:null)).

%
%

common_noun macro
     (common_l,
      synsem:(local:(cat:(head:(common,
                              case:nom,
                              agr:(num:sing,
                                   per:third),
                              mod:null,
                              n_ind:NInd,
                              pred:minus,
                              rel:minus,
                              poss:none),
                         subcat:e_list,
                         subj:null),
                     cont:(Cont,index:NInd)),
              nonlocal:(inherited:slash:null,
                        tobind:slash:null)
             ),
       qstore:e_list
      ).
        
opt(X) macro
   (opt,s_arg:X).
obl(X) macro
   (obl,s_arg:X).

np(Head,Ind) macro
   (local:(cat:(head:(Head,noun,mod:null,rel:minus,pred:minus),
                subcat:e_list),
           cont:index:Ind),
    nonlocal:( tobind:slash:null)
   ).        

vp(Head,Cont) macro
     (local:(cat:(head:(Head,mod:null),
                  subcat:e_list),
             cont:Cont),
      nonlocal:(tobind:slash:null)
     ).

slashinh(X) macro
 (synsem:nonlocal:inherited:slash:X).
slashtob(X) macro
 (synsem:nonlocal:tobind:slash:X).

f_phrase macro
  (phrase,
  synsem:local:cat:subcat:e_list,synsem:nonlocal:inherited:slash:null,
      synsem:nonlocal:tobind:slash:null).

f_sent macro
  (@f_phrase,synsem:local:cat:head:(finite;pred:plus)).

\section{Definite Clauses}
%
%

head_feature_principle(synsem:local:cat:head:X,synsem:local:cat:head:X) if
        true.

%
combine_semantics( synsem:local:cont:(index:HInd,restr:Hrest),
                    synsem:local:cont:(index:DInd,restr:Drest),
                    synsem:local:cont:(index:HInd,restr:MRest)) if
        appendset(Hrest,Drest,MRest).

combine_semantics( synsem:local:cont:(index:HInd,restr:Hrest),
                    synsem:local:cont:(Drest,psoa),
                    synsem:local:cont:(index:HInd,restr:MRest)) if
        appendset(Hrest,(el:Drest,els:[]),MRest).

combine_semantics( synsem:local:cont:(nucleus:HNuc,quants:HQ),
                   synsem:local:cont:(DCont),
                   synsem:local:cont:(nucleus:HNuc,
                        quants:([(det:the,restind:DCont) |HQ]     ))) if
                true.

combine_semantics( synsem:local:cont:(nucleus:HNuc,quants:HQ),
                   synsem:local:cont:(psoa),
                   synsem:local:cont:(nucleus:HNuc,quants:HQ)) if true.

%
adjunct_principle((synsem:local:cat:(subj:Subj,subcat:Subcat,adjuncts:MAdjs)),
                  (synsem:local:cat:(head:mod:(modsyn:(Mod),
                                                modadj:MAdjs),
                                     subcat:[])),
                  synsem:(Mod,local:cat:(CatH,subj:Subj,subcat:Subcat))) if 
        checkposs(Mod),
        checksubst(CatH).

checkposs(local:cat:head: =\(\backslash\)= noun) if true,!.
checkposs(local:cat:head:poss:none) if true.
checkposs(local:cat:(head:(poss:poss,pred:minus),subcat:ne_set)) if true.
checkposs(local:cat:(head:(poss:poss,pred:minus),subcat:ne_list)) if true.
checkposs(local:cat:(head:(poss:poss,pred:plus),subcat:tl:ne_set)) if true.
checkposs(local:cat:(head:(poss:poss,pred:plus),subcat:tl:ne_list)) if true.

checksubst(head: =\(\backslash\)= noun) if true,!.
checksubst(head:(pred:minus)) if true.
checksubst((head:(pred:plus),subcat:ne_set)) if true.
checksubst((head:(pred:plus),subcat:ne_list)) if true.

%
nonlocal_principle((@slashtob((local,LAdj)),@slashinh((LAdj))),
                   (@slashtob((HT,null)),@slashinh(null)),
                   (@slashtob(HT),@slashinh(null))) if true.

nonlocal_principle((@slashinh(null),@slashtob(null)),
                   (@slashinh(HI),@slashtob(HT)),
                   (@slashinh(HI),@slashtob(HT))) if true,!.

nonlocal_principle((@slashinh(null),@slashtob(null)),
                   (@slashtob((HT)),@slashinh(null)),
                   (@slashinh(null),@slashtob(HT))) if true.

nonlocal_principle((@slashinh(AI),@slashtob(null)),
                   (@slashtob((null)),@slashinh(null)),
                   (@slashinh(AI),@slashtob(HT))) if true.

nonlocal_principle((@slashinh(HT),Arg),
                   (@slashtob((HT,local)),Head),
                   (Mother,@slashtob(HT),@slashinh(HT))) if 
                                check_rel(Arg,Mother).

check_rel((@slashinh(S)),
          (synsem:(local:cat:(head:subj_rel,
                              subj:nonlocal:inherited:slash:S),
                   nonlocal:tobind:slash:S))) if true.

check_rel((@slashinh(S), synsem:local: =\(\backslash\)= S),
          (synsem:(local:(cat:(head:subj_rel,n_inc:plus,
                               subj:local:cat:adjuncts:non_ref:plus)),
                   nonlocal:tobind:slash:S))) if true.

check_rel((@slashinh(S)),
          (synsem:(local:cat:(head:obj_rel,
                              subj:nonlocal:inherited:slash: =\(\backslash\)= S),
                   nonlocal:tobind:slash:S))) if true.


append([],Xs,Xs) if
  true.
append([H|T1],L2,[H|T2]) if
  append(T1,L2,T2).

appends(e_list,X,X) if true,!.
appends([A],(X,ne_set),[A|[X]]) if true,!.
appends(X,[],X) if true,!.
appends((X,set),(L,list),[X|L]) if true.
appends([X|Rx],Y,[X|Res]) if
                appends(Rx,Y,Res).
appends((el:X,els:Rx),(Y,set),(el:X,els:Res)) if
                appends(Rx,Y,Res).

listlast(X,(hd:X,tl:e_list),e_list) if true,!.
listlast(X,[H|T],[H|R]) if
        listlast(X,T,R).

permut(e_list,e_list) if true , !.
permut((X,set),(X,set)) if true.
permut((el:X,els:R),(el:Y,els:(el:X,els:R2))) if
                permut(R,(el:Y,els:R2)).

selectlast(Arg,(ne_list_synsem,Sursub),Reslist) if
        listlast(Arg,Sursub,Reslist),!.

selectlast(Arg,Sub,Reslist) if
        removeop(Sub,SubRem),
        surface(SubRem,Sursub),
        listlast(Arg,Sursub,Reslist).

selectfirst(Arg,(ne_list_synsem,[Arg|Rest]),Rest) if !,true.

selectfirst(Arg,Sub,Reslist) if
        removeop(Sub,SubRem),
        surface(SubRem,[Arg|Reslist]).

select(Arg,(s_arg:Arg),[]) if true.
select(Arg,(Arg,subcat_type),[]) if true.
select(T,(X,set),Z) if
                permut(X,(el:Ct,els:Cr)),
                select(T,Ct,Res),
                appends(Res,Cr,Z).
select(T,[Xt|R],Z) if
                select(T,Xt,Res),
                appends(Res,R,Z).

appendset(e_list,(set,X),X) if true.
appendset((el:El,els:Rels),(set,S2),(el:El,els:Res)) if
        appendset(Rels,S2,Res).

removeop([],[]) if true.
removeop(s_arg:X,@obl(X)) if true.
removeop([(opt)|R],R2) if removeop(R,R2).
removeop([H|R],[H2|R2]) if removeop(H,H2),removeop(R,R2).
removeop((el:(opt),els:R),R2) if removeop(R,R2).
removeop((el:H,els:R),(el:H2,els:R2)) if removeop(H,H2),removeop(R,R2).

surface(e_list,e_list) if true.
surface(X,[T|Rs]) if
                select(T,X,R),surface(R,Rs).

append_comp((hd_dtr:Head,comp_dtrs:[],spec_dtrs:[],subj_dtr:Comp),
            (Head,word,(synsem:local:cat:subj:Subj)),
            (Comp,synsem:Subj)) if true,!.
append_comp((subj_dtr:Comp,comp_dtrs:Cdtrs,spec_dtrs:Sdtrs,hd_dtr:Hdtr),
            (synsem:local:cat:subj:Subj,
               dtrs:(comp_dtrs:Cdtrs,spec_dtrs:Sdtrs,hd_dtr:Hdtr)),
            (Comp,synsem:Subj)) if true,!.
append_comp((hd_dtr:Head,comp_dtrs:[Comp],spec_dtrs:[]),
                (Head,word), Comp) if true,!.
append_comp((hd_dtr:Hdtr,comp_dtrs:ResComp,spec_dtrs:Sdtrs),
            (dtrs:(comp_dtrs:Comp1,hd_dtr:Hdtr,spec_dtrs:Sdtrs)),
            Comp) if
                append([Comp],Comp1,ResComp).

append_spec((hd_dtr:Head,comp_dtrs:[],spec_dtrs:[Adjunct]),
                (Head,word), Adjunct) if true,!.
append_spec((hd_dtr:Hdtr,comp_dtrs:Cdtrs,subj_dtr:SUdtr,spec_dtrs:ResSpec),
            dtrs:(hd_dtr:Hdtr,comp_dtrs:Cdtrs,subj_dtr:SUdtr,spec_dtrs:Spec1),
            Spec) if
                append([Spec],Spec1,ResSpec).


apply_case(
(word,
 synsem:(local:(cat:(head:(common,
                           case:nom,
                           agr:(num:Num,per:third),
                           mod:Mod,
                           rel:(Rel,minus),
                           pred:(Pred,minus),
                           n_ind:NInd,
                           poss:Poss),
                     subcat:Subcat,
                     adjuncts:Adjuncts,
                     subj:Subj),
                cont:Cont,
                conx:Conx),
          nonlocal:Nonlocal),
    qstore:Qs) ,
(word,
 synsem:(local:(cat:(head:(common,
                           case:Case,
                           agr:(num:Num,per:third),
                           mod:Mod,
                           rel:Rel,
                           pred:Pred,
                           n_ind:NInd,
                           poss:Poss),
                     subj:Subj,
                     adjuncts:Adjuncts,
                     subcat:Subcat),
                cont:Cont,
                conx:Conx),
          nonlocal:Nonlocal),
    qstore:Qs) , Case , CMod ) if check_case_mod(Poss,CMod).

check_case_mod(none,a) if true.
check_case_mod(by:per:first,a) if true.
check_case_mod(by:per:second,a) if true.
check_case_mod(by:per:third,b) if true.

apply_poss(
(word,
 synsem:(local:(cat:(head:(common,
                           case:nom,
                           agr:(Agr,per:third),
                           mod:Mod,
                           rel:(Rel,minus),
                           pred:(Pred,minus),
                           poss:none),
                     subcat:Subcat),
                cont:(index:Ind,restr:(el:EL)),
                conx:Conx),
          nonlocal:Nonlocal),
  qstore:Qs) ,
(word,
 synsem:(local:(cat:(head:(common,
                           case:nom,
                           agr:Agr,
                           mod:Mod,
                           rel:Rel,
                           pred:Pred,
                           n_ind:Ind,
                           poss:by:By),
                     subj:Subj,
                     subcat:Subcat2),
                cont:(index:Ind,restr:(el:EL,els:(el:(nucleus:(relation,
                                                     name:[p,o,s,s,e,s,s],
                                                     args:
                                        [(argname:[o,w,n,e,r],arg:Ind2),
                                         (argname:[o,w,n,e,d],arg:Ind)]),
                                         quants:e_list),els:e_list))),
                conx:Conx),
          nonlocal:Nonlocal),
  qstore:Qs) , By ) if 
           appends([@obl((Subj,@np((case:gen,agr:By),Ind2)))],Subcat,Subcat2).


apply_copula(
(word,
 synsem:(local:(cat:(head:(common,
                           case:nom,
                           rel:(Rel,minus),
                           pred:minus,
                           poss:Poss),
                     subj:Subj,
                     subcat:Subcat),
                cont:(C1,restr:Restr),
                conx:Conx),
          nonlocal:Nonlocal),
  qstore:Qs) ,
(noun_l,
 synsem:(local:(cat:(head:(common,
                           case:nom,
                           agr:(Agr),
                           mod:null,
                           rel:Rel,
                           pred:plus,
                           n_ind:Ind,
                           poss:Poss),
                     subj:Subj2,
                     subcat:Subcat2),
                cont:C2,
                conx:Conx),
          nonlocal:Nonlocal),
  qstore:Qs), Agr) if 
                contentcop(C1,C2,Ind),
                appends([@obl((Subj2,@np((case:nom,agr:Agr),_)))],Subcat,Subcat2).


apply_adj2noun(
(word,
 synsem:(local:(cat:(head:((qualitative_adj;rel:plus),
                           mod:(modsyn:Mod,modadj:Modadj)),
                     subj:Subj,
                     subcat:Subcat),
                cont:(index:Ind,Cont),
                conx:Conx),
         nonlocal:Nonlocal),
  qstore:Qs),
(noun_l,
 synsem:(Mod,local:(cat:(head:(mod:null,
                               case:nom,
                               agr:(per:third,num:sing),
                               rel:minus,
                               pred:minus,
                               n_ind:Ind,
                               poss:none),
                          subj:Subj,
                          subcat:Subcat),
                    cont:Cont,
                    conx:Conx),
              nonlocal:Nonlocal),
   qstore:Qs)) if true.

\section{Lexicon}
%
%

k{\i}rm{\i}z{\i} ---> 
    (qualitative_l,phon:[[k,{\i},r,m,{\i},z,{\i}]],
     synsem:(local:(cat:(head:(countable:plus,gradable:plus,
                             mod:modsyn:(local:(cat:head:n_ind:NInd,
                                          cont:(index:Ind)))),
                         subcat:[],
                        subj:null),
                   cont:(index:Ind,
                         restr:(el:(quants:e_list,
                                nucleus:(name:[r,e,d],inst:NInd)),
                                els:e_list)
                        )
                   ),
            nonlocal:tobind:slash:null)
     ).

ben --->
    (word,phon:[[b,e,n]],
     synsem:(local:(cat:(head:(personal_pr,
                         case:nom,
                         rel:minus,
                            agr:(num:sing,
                              per:first),
                         mod:null,
                         poss:none),
                            subcat:e_list,
                       subj:null),
                   cont:(npro,
                         index:(Ind,per:first,num:sing),
                         restr:e_list),
                   conx:conx
                   ),
             nonlocal:tobind:slash:null)
                 ).

kap{\i} --->
     ( @common_noun,
      phon:[[k,a,p,{\i}]],
      synsem:local:cont:(npro,
                         index:(agr,Ind,per:third,num:sing),
                         restr:(el:(nucleus:(name:[d,o,o,r],inst:Ind),
                                 quants:[]),els:[]))
     ).

ev --->
    (@common_noun,
     phon:[[e,v]],
     synsem:local:cont:(npro,
                         index:(agr,Ind,per:third,num:sing),
                         restr:(el:(nucleus:(name:[h,o,u,s,e],inst:Ind),
                                 quants:[]),els:[])
                       )
     ).

gitti --->
    (finite_l,
     phon:[[g,i,t,t,i]],
     synsem:local:(cat:(head:(finite,mod:null,
           tense:past,
           vagr:(Agr,(per:third,num:sing))),
      subcat: { @obl((Subj,@np((agr:Agr,case:nom),SInd))),
                            @opt(@np(case:abl,FInd)),
                            @opt(@np(case:dat,TInd))},
      subj:Subj
            ),
      cont:(quants:[],nucleus:(name:[g,o],args:[(argname:[g,o,e,r],arg:SInd),
                                                (argname:[t,o],arg:TInd),
                                                (argname:[f,r,o,m],arg:FInd)])
           ))).

giden --->
    (subj_rel_l,
     phon:[[g,i,d,e,n]],
     synsem:(local:(cat:(head:(subj_rel,
           vcase:nom),
     subcat: { @obl((Subj,@np(case:nom,SInd))),
                            @opt(@np(case:abl,FInd)),
                            @opt(@np(case:dat,TInd))},
      subj:Subj
            ),
      cont:restr:(el:(quants:[],nucleus:(name:[g,o],
                                      args:[(argname:[g,o,e,r],arg:SInd),
                                                (argname:[t,o],arg:TInd),
                                                (argname:[f,r,o,m],arg:FInd)])
           ),els:[]))
      )). 

geldi\u{g}i --->
    (obj_rel_l,
     phon:[[g,e,l,d,i,\u{g},i]],
     synsem:(local:(cat:(head:(obj_rel,
           tense:past,
           vagr:(Agr,(per:third,num:sing))),
     subcat: { @obl((Subj,@np((agr:Agr,case:gen),SInd))),
                                 @opt(@np(case:abl,FInd)),
                                 @opt(@np(case:dat,TInd))},
      subj:Subj
            ),
      cont:restr:(el:(quants:[],nucleus:(name:[g,o],
                                      args:[(argname:[c,o,m,e,r],arg:SInd),
                                                (argname:[t,o],arg:TInd),
                                                (argname:[f,r,o,m],arg:FInd)])
        ),els:[]))
    )). 

s\"{o}yl\"{u}yor --->
    (finite_l,
     phon:[[s,\"{o},y,l,\"{u},y,o,r]],
     synsem:(local:(cat:(head:(finite,
           tense:past,
           mod:null,
           vagr:(Agr,(per:third,num:sing))),
     subcat: { @obl((Subj,@np((agr:Agr,case:nom),SInd))),
                                 @obl(@vp((vcase:obj,complemented),Spsoa)),
                                 @opt(@np(case:dat,TInd))},
      subj:Subj
            ),
      cont:(quants:[],nucleus:(name:[t,e,l,l],
                                      args:[(argname:[t,e,l,l,e,r],arg:SInd),
                                                (argname:[t,o],arg:TInd),
                                              (argname:[w,h,a,t],arg:Spsoa)])
        ))
    )).

empty
    (word,phon:[[p,r,o]],
     synsem:(local:(cat:(head:(pronoun,
                               case:(nom;gen),
                                  agr:Agr,
                               mod:null,
                               n_ind:Agr,
                               pred:minus,
                               rel:minus,
                               poss:none),
                         subcat:e_list,
                         subj:null),
                    cont:(npro,
                          index:(Agr),
                          restr:[])),
              nonlocal:(inherited:slash:null,
                        tobind:slash:null)
                  )).

\section{Lexical Rules}
%
%

back(a).
back({\i}).
back(o).
back(u).
kalin_hece([X]) :- back(X),!.
kalin_hece([X,_]) :- back(X).

front(e).
front(i).
front(\"{o}).
front(\"{u}).
ince_hece([X]) :- front(X),!.
ince_hece([X,_]) :- front(X).
wovel(X) :- front(X),!.
wovel(X) :- back(X).

backrounded(o).
backrounded(u).
b_r_hece([X]) :- backrounded(X),!.
b_r_hece([X,_]) :- backrounded(X).

frontrounded(\"{o}).
frontrounded(\"{u}).
f_r_hece([X]) :- frontrounded(X),!.
f_r_hece([X,_]) :- frontrounded(X).

backunrounded(a).
backunrounded({\i}).
b_u_hece([X]) :- backunrounded(X),!.
b_u_hece([X,_]) :- backunrounded(X).

frontunrounded(e).
frontunrounded(i).
f_u_hece([X]) :- frontunrounded(X),!.
f_u_hece([X,_]) :- frontunrounded(X).

yumusa(p,b).
yumusa(\c{c},c).
yumusa(t,d).
yumusa(k,\u{g}).

kal_yum([X,Y],Yum) :- back(X),yumusa(Y,Yum).
ince_yum([X,Y],Yum) :- front(X),yumusa(Y,Yum).

f_u_yum([X,Y],Yum) :- frontunrounded(X),yumusa(Y,Yum).
b_u_yum([X,Y],Yum) :- backunrounded(X),yumusa(Y,Yum).
f_r_yum([X,Y],Yum) :- frontrounded(X),yumusa(Y,Yum).
b_r_yum([X,Y],Yum) :- backrounded(X),yumusa(Y,Yum).

:-lex_rule_depth(4).

plural lex_rule
  Cat1 **> Cat2
if apply_plural((Cat1,phon:[Phon]),(Cat2,phon:[Phon2])),
   append(Phon,[ø,p,l,u],Phon2)
morphs
  (X,L2) becomes (X,L2,lar) when kalin_hece(L2),
  (X,L2) becomes (X,L2,ler) when ince_hece(L2).


accusative_a lex_rule
  Cat1 **> Cat2
if apply_case((Cat1,phon:[Phon]),(Cat2,phon:[Phon2]),obj,a),
   append(Phon,[ø,o,b,j],Phon2)
morphs
  (X,[L]) becomes (X,[L],y{\i}) when backunrounded(L),
  (X,[L]) becomes (X,[L],yi) when frontunrounded(L),
  (X,[L]) becomes (X,[L],yu) when backrounded(L),
  (X,[L]) becomes (X,[L],y\"{u}) when frontrounded(L),
  (X,[L1,L2]) becomes (X,L1,[Y],[{\i}]) when b_u_yum([L1,L2],Y),
  (X,[L1,L2]) becomes (X,L1,[Y],[i]) when f_u_yum([L1,L2],Y),
  (X,[L1,L2]) becomes (X,L1,[Y],[u]) when b_r_yum([L1,L2],Y),
  (X,[L1,L2]) becomes (X,L1,[Y],[\"{u}]) when f_r_yum([L1,L2],Y),
  (X,L2) becomes (X,L2,[{\i}]) when b_u_hece(L2),
  (X,L2) becomes (X,L2,i) when f_u_hece(L2),
  (X,L2) becomes (X,L2,u) when b_r_hece(L2),
  (X,L2) becomes (X,L2,[\"{u}]) when f_r_hece(L2).

accusative_b lex_rule
  Cat1 **> Cat2
if apply_case((Cat1,phon:[Phon]),(Cat2,phon:[Phon2]),obj,b),
   append(Phon,[ø,o,b,j],Phon2)
morphs
  (X,[L]) becomes (X,[L],n{\i}) when backunrounded(L),
  (X,[L]) becomes (X,[L],ni) when frontunrounded(L),
  (X,[L]) becomes (X,[L],nu) when backrounded(L),
  (X,[L]) becomes (X,[L],n\"{u}) when frontrounded(L).


possessive_3_s lex_rule
  Cat1 **> Cat2
if
apply_poss((Cat1,phon:[Phon]),(Cat2,phon:[Phon2]),(num:sing,per:third)),
  append(Phon,[ø,t,s,g],Phon2)
morphs
  (X,[L]) becomes (X,[L],s{\i}) when backunrounded(L),
  (X,[L]) becomes (X,[L],si) when frontunrounded(L),
  (X,[L]) becomes (X,[L],su) when backrounded(L),
  (X,[L]) becomes (X,[L],s\"{u}) when frontrounded(L),
  (X,[L1,L2]) becomes (X,L1,[Y],[{\i}]) when b_u_yum([L1,L2],Y),
  (X,[L1,L2]) becomes (X,L1,[Y],[i]) when f_u_yum([L1,L2],Y),
  (X,[L1,L2]) becomes (X,L1,[Y],[u]) when b_r_yum([L1,L2],Y),
  (X,[L1,L2]) becomes (X,L1,[Y],[\"{u}]) when f_r_yum([L1,L2],Y),
  (X,L2) becomes (X,L2,[{\i}]) when b_u_hece(L2),
  (X,L2) becomes (X,L2,i) when f_u_hece(L2),
  (X,L2) becomes (X,L2,u) when b_r_hece(L2),
  (X,L2) becomes (X,L2,[\"{u}]) when f_r_hece(L2).


obj_rel_to_compl lex_rule
(obj_rel_l,
 phon:[Phon],
 synsem:(local:(cat:(head:(obj_rel,
                           vcase:nom,
                           neg:Neg,
                           vagr:Agr,
                           n_inc:N_Inc,
                           tense:Tense
                           ),
                     subj:Subj,
                     adjuncts:Adj,
                     subcat:Subcat),
                cont:restr:el:Rest,
                conx:Conx))
) **>
(complement_l,
 phon:[Phon],
 synsem:(local:(cat:(head:(complemented,
                           vcase:nom,
                           neg:Neg,
                           mod:null,
                           vagr:Agr,
                           n_inc:N_Inc,
                           tense:Tense
                           ),
                     subj:Subj,
                     adjuncts:Adj,
                     subcat:Subcat),
                cont:Rest,
                conx:Conx),
         nonlocal:tobind:slash:null)
)
morphs
   X becomes X.


relativizer lex_rule
  Cat1 **> Cat2
if apply_reltvzr((Cat1,phon:[Phon]),(Cat2,phon:[Phon2])),
   append(Phon,[ø,r,l,t],Phon2)
morphs
   (X) becomes (X,ki).


copula1_s lex_rule
  Cat1 **> Cat2
if apply_copula((Cat1,phon:[Phon]),(Cat2,phon:[Phon2]),(num:sing,per:first)),
   append(Phon,[ø,c,o,p],Phon2)
morphs
  (X,[L2]) becomes (X,[L2],y{\i}m) when backunrounded(L2),
  (X,[L2]) becomes (X,[L2],yim) when frontunrounded(L2),
  (X,[L2]) becomes (X,[L2],yum) when backrounded(L2),
  (X,[L2]) becomes (X,[L2],y\"{u}m) when frontrounded(L2),
  (X,[L1,L2]) becomes (X,L1,[Y],[{\i},m]) when b_u_yum([L1,L2],Y),
  (X,[L1,L2]) becomes (X,L1,[Y],[i,m]) when f_u_yum([L1,L2],Y),
  (X,[L1,L2]) becomes (X,L1,[Y],um) when b_r_yum([L1,L2],Y),
  (X,[L1,L2]) becomes (X,L1,[Y],[\"{u},m]) when f_r_yum([L1,L2],Y),
  (X,L2) becomes (X,L2,[{\i}],m) when b_u_hece(L2),
  (X,L2) becomes (X,L2,im) when f_u_hece(L2),
  (X,L2) becomes (X,L2,um) when b_r_hece(L2),
  (X,L2) becomes (X,L2,[\"{u}],m) when f_r_hece(L2).


copula3_s lex_rule
  Cat1 **> Cat2
if apply_copula((Cat1,phon:[Phon]),(Cat2,phon:[Phon2]),(num:sing,per:third)),
   append(Phon,[ø,c,o,p],Phon2)
morphs
  (X,[L1,L2]) becomes (X,L1,[L2],t{\i}r) when b_u_yum([L1,L2],Y),
  (X,[L1,L2]) becomes (X,L1,[L2],tir) when f_u_yum([L1,L2],Y),
  (X,[L1,L2]) becomes (X,L1,[L2],tur) when b_r_yum([L1,L2],Y),
  (X,[L1,L2]) becomes (X,L1,[L2],t\"{u}r) when f_r_yum([L1,L2],Y),
  (X,L2) becomes (X,L2,d{\i}r) when b_u_hece(L2),
  (X,L2) becomes (X,L2,dir) when f_u_hece(L2),
  (X,L2) becomes (X,L2,dur) when b_r_hece(L2),
  (X,L2) becomes (X,L2,d\"{u}r) when f_r_hece(L2).

adj_promotion lex_rule
  Cat1 **> Cat2
if apply_adj2noun((Cat1,phon:[Phon]),(Cat2,phon:[Phon2])),
   append(Phon,[ø,a],Phon2)
morphs
   X becomes X.


non_ref_object lex_rule
(verb_l,
 phon:[PhonV],
 synsem:(local:(cat:(head:HeadV,
                     subj:SubjV,
                     adjuncts:AdjV,
                     subcat:Subcat1V),
                cont:ContV,
                conx:ConxV),
         nonlocal:NonlocalV)
) **>
(verb_l,
 phon:[PhonV],
 synsem:(local:(cat:(head:HeadV,
                     subj:SubjV,
                     adjuncts:AdjV,
                     subcat:[ SubcatRV, 
                             @obl((local:(cat:(head:(common,
                                                     case:nom,
                                                     agr:AgrN,
                                                     mod:ModN,
                                                     rel:(RelN),
                                                     pred:(PredN),
                                                     n_ind:NIndN,
                                                     poss:PossN),
                                                subcat:SubcatN,
                                                adjuncts:(AdjunctsN,
                                                          non_ref:plus),
                                                subj:SubjN),
                                          cont:ContN,
                                          conx:ConxN),
                                   nonlocal:NonlocalN))]
                     ),
                cont:ContV,
                conx:ConxV),
         nonlocal:NonlocalV)
)
if selectlast( 
 (local:(cat:(head:(common,
                     case:obj,
                     agr:AgrN,
                     mod:ModN,
                     rel:(RelN),
                     pred:(PredN),
                     n_ind:NIndN,
                     poss:PossN),
                subcat:SubcatN,
                adjuncts:AdjunctsN,
                subj:SubjN),
          cont:ContN,
          conx:ConxN),
  nonlocal:NonlocalN), Subcat1V, SubcatRV) 
morphs
   X becomes X.

}
\label{sonsayfa}
\end{document}